\documentclass[]{lumen}
\usepackage{graphicx} 
\usepackage{xcolor}
\usepackage[table]{xcolor}
\usepackage{xparse}
\usepackage{tikz}
\usepackage{pifont}
\usepackage{wrapfig}
\usepackage{longtable}
\usepackage[most]{tcolorbox}
\usepackage{fontawesome5}
\usepackage{varwidth}
\usepackage{needspace}

\definecolor{table_creamyellow}{HTML}{FFFBEE}
\definecolor{table_backbone_category}{HTML}{9ECAD6}
\definecolor{box_midyellow}{HTML}{F4D35E}
\definecolor{crTask}{HTML}{DDBEA8}
\definecolor{crData}{HTML}{D7A449}
\definecolor{crArchitecture}{HTML}{E65E93}
\definecolor{crPerspective}{HTML}{F4D35E}
\definecolor{crRepresentation}{HTML}{21B6A2}
\definecolor{crPerception}{HTML}{0EA5E9}
\definecolor{crGeneration}{HTML}{FA8107}
\definecolor{crEnactment}{HTML}{D691CE}
\definecolor{crReal}{HTML}{154b10}
\definecolor{crSynthetic}{HTML}{5fadec}
\definecolor{crEstimated}{HTML}{F62F63}
\definecolor{crTrainScratch}{HTML}{F0D3F7}
\definecolor{crTrainFull}{HTML}{819A91}
\definecolor{crTrainParameter}{HTML}{A7C1A8}
\definecolor{crTrainInstruction}{HTML}{744577}
\definecolor{crTrainReward}{HTML}{C9996B}
\definecolor{crTrainTask}{HTML}{1F6F5F}
\definecolor{crTrainDistillation}{HTML}{AC58F5}
\definecolor{crInferenceSemantic}{HTML}{03AED2}
\definecolor{crInferenceGuided}{HTML}{748DAE}
\definecolor{crInferenceIterative}{HTML}{FF3737}
\definecolor{crInferencePreference}{HTML}{FF85BB}
\definecolor{crInferenceRetrieve}{HTML}{7148FC}
\definecolor{crDataSelfCaptured}{HTML}{91D06C}
\definecolor{crDataPublicCurated}{HTML}{F77F00}
\definecolor{crDataWebCollected}{HTML}{FE9EC7}
\definecolor{crDataSynthesized}{HTML}{81A6C6}
\definecolor{crEgo}{HTML}{F1D8B9}
\definecolor{crExo}{HTML}{DDECDD}
\definecolor{crSMA}{HTML}{DEB841}
\definecolor{crSFA}{HTML}{FC9047}
\definecolor{crIGA}{HTML}{386F58}
\definecolor{crHA}{HTML}{B6CCD1}

\NewDocumentCommand{\crbx}{ O{0pt} m m }{\raisebox{#1}{\scalebox{0.75}{\fcolorbox{#2}{#2}{\makebox[1.5ex][c]{\rule{0pt}{1.5ex}#3}}}}}
\NewDocumentCommand{\crcc}{ O{0pt} m m }{\raisebox{#1}{\tikz[baseline=(char.base)]{\node[shape=circle,fill=#2,inner sep=0.5pt,minimum width=2.5ex,text=white] (char) {\scalebox{0.85}{#3}};}}}
\NewDocumentCommand{\crbxrect}{ O{0pt} m m }{\raisebox{#1}{\scalebox{0.85}{\fcolorbox{#2}{#2}{\makebox[3.5ex][c]{\rule{0pt}{1.5ex}#3}}}}}

\NewDocumentCommand{\Traditional}{ O{-0.5ex} O{0.026\linewidth}}{\raisebox{#1}{\includegraphics[width=#2]{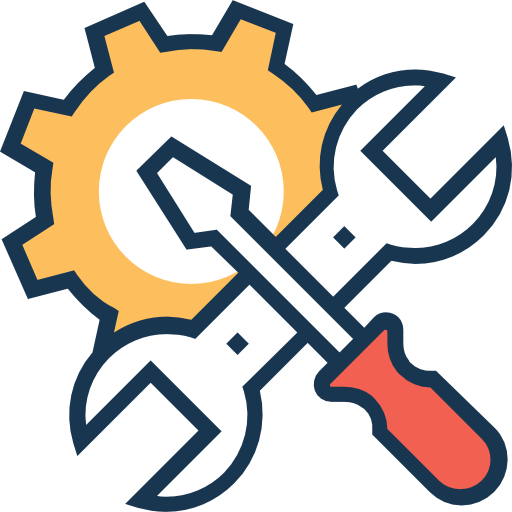}}}
\NewDocumentCommand{\DeepLearning}{ O{-0.5ex} O{0.026\linewidth}}{\raisebox{#1}{\includegraphics[width=#2]{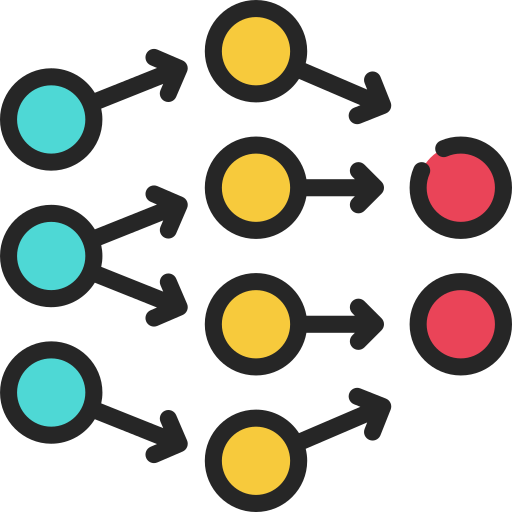}}}
\NewDocumentCommand{\LargeModels}{ O{-0.5ex} O{0.026\linewidth}}{\raisebox{#1}{\includegraphics[width=#2]{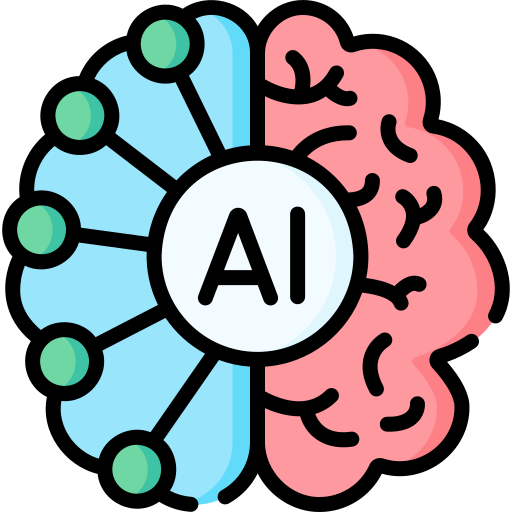}}}

\NewDocumentCommand{\Image}{ O{-0.5ex} O{0.026\linewidth}}{\raisebox{#1}{\includegraphics[width=#2]{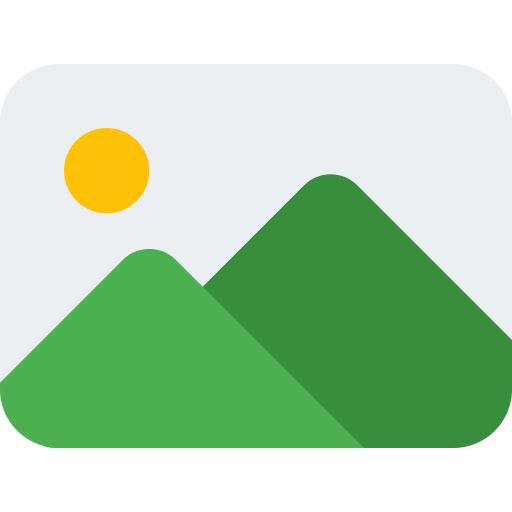}}}
\NewDocumentCommand{\HumanImage}{ O{-0.5ex} O{0.026\linewidth}}{\raisebox{#1}{\includegraphics[width=#2]{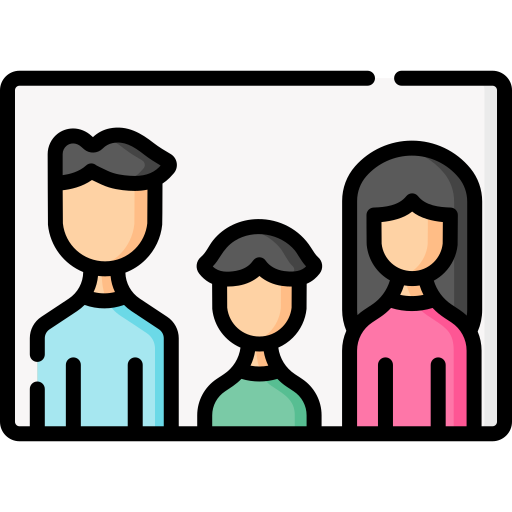}}}
\NewDocumentCommand{\Video}{ O{-0.5ex} O{0.026\linewidth}}{\raisebox{#1}{\includegraphics[width=#2]{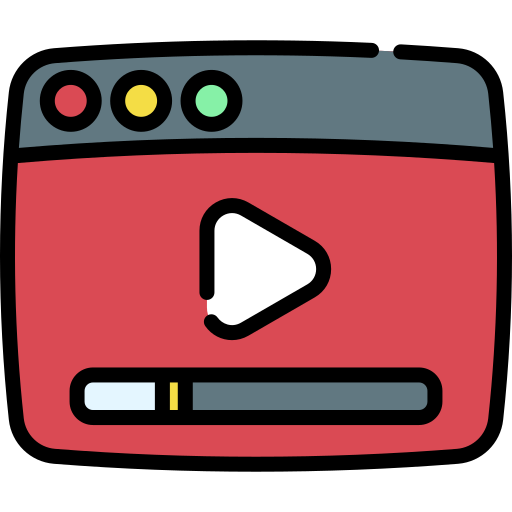}}}
\NewDocumentCommand{\HumanVideo}{ O{-0.5ex} O{0.026\linewidth}}{\raisebox{#1}{\includegraphics[width=#2]{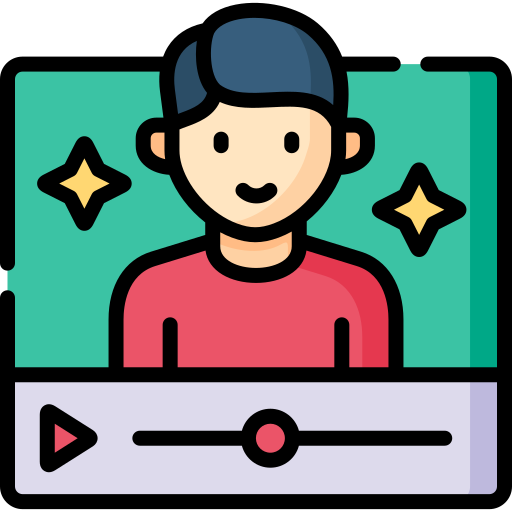}}}
\NewDocumentCommand{\EgoImage}{ O{-0.5ex} O{0.026\linewidth}}{\raisebox{#1}{\includegraphics[width=#2]{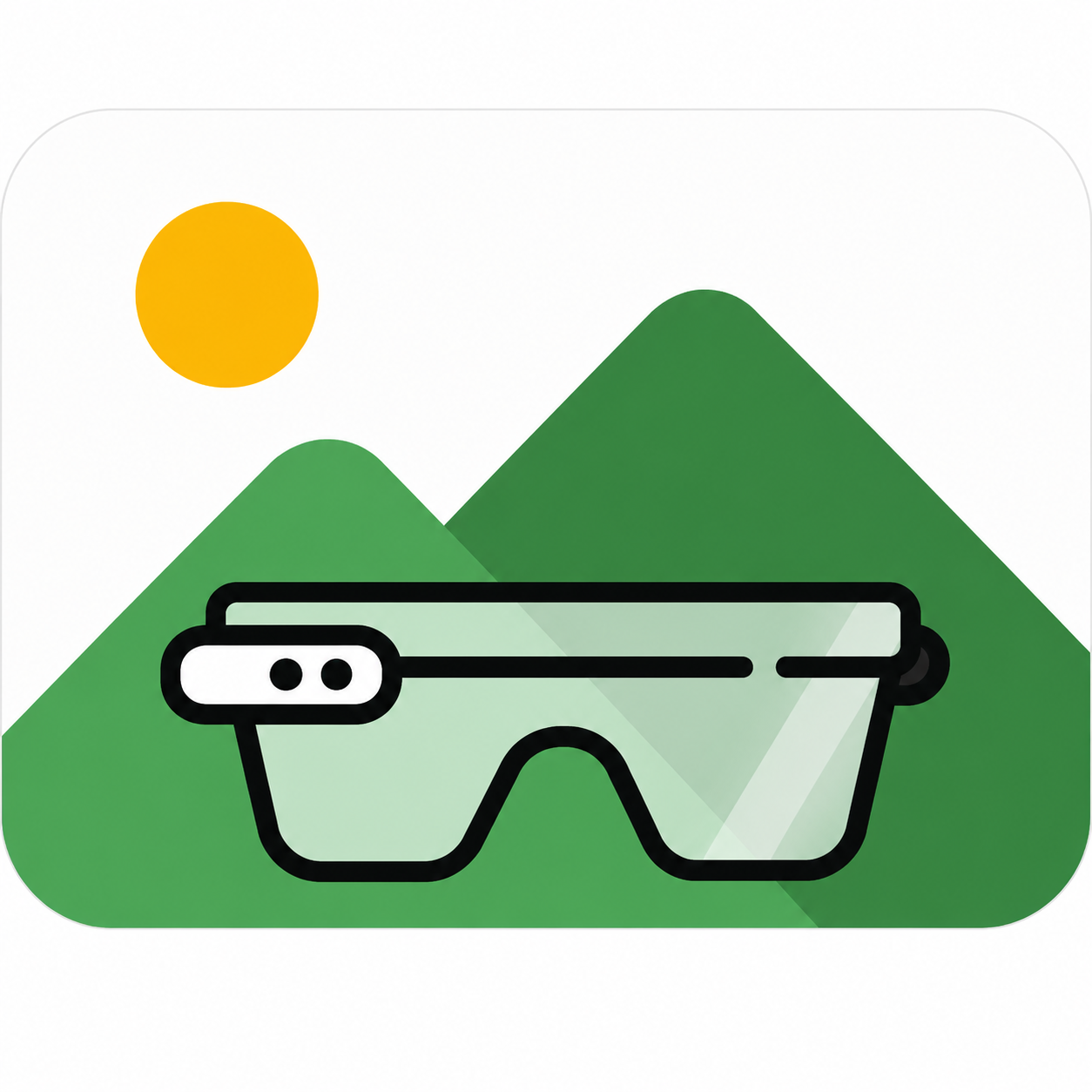}}}
\NewDocumentCommand{\EgoVideo}{ O{-0.5ex} O{0.026\linewidth}}{\raisebox{#1}{\includegraphics[width=#2]{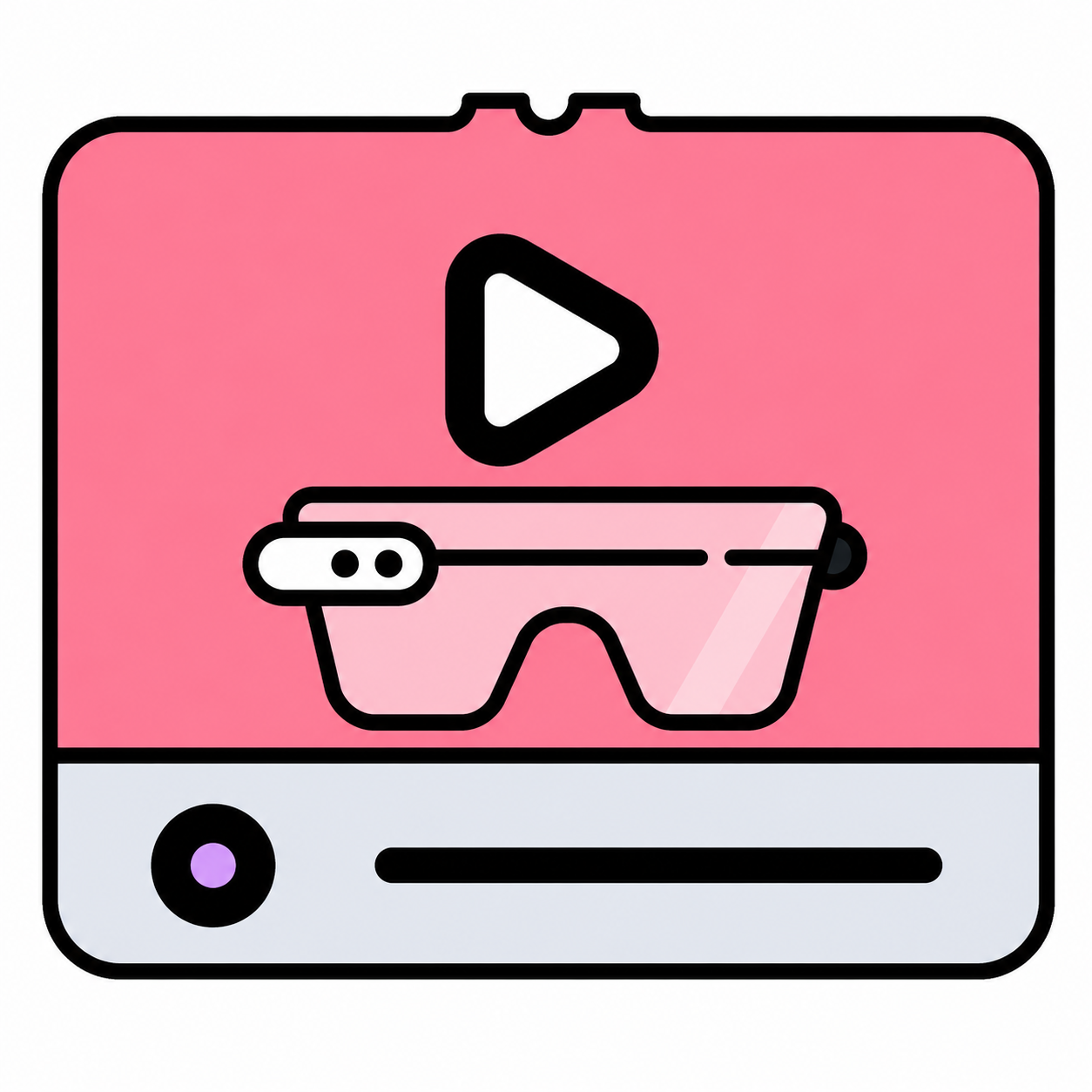}}}
\NewDocumentCommand{\HumanSkeleton}{ O{-0.5ex} O{0.026\linewidth}}{\raisebox{#1}{\includegraphics[width=#2]{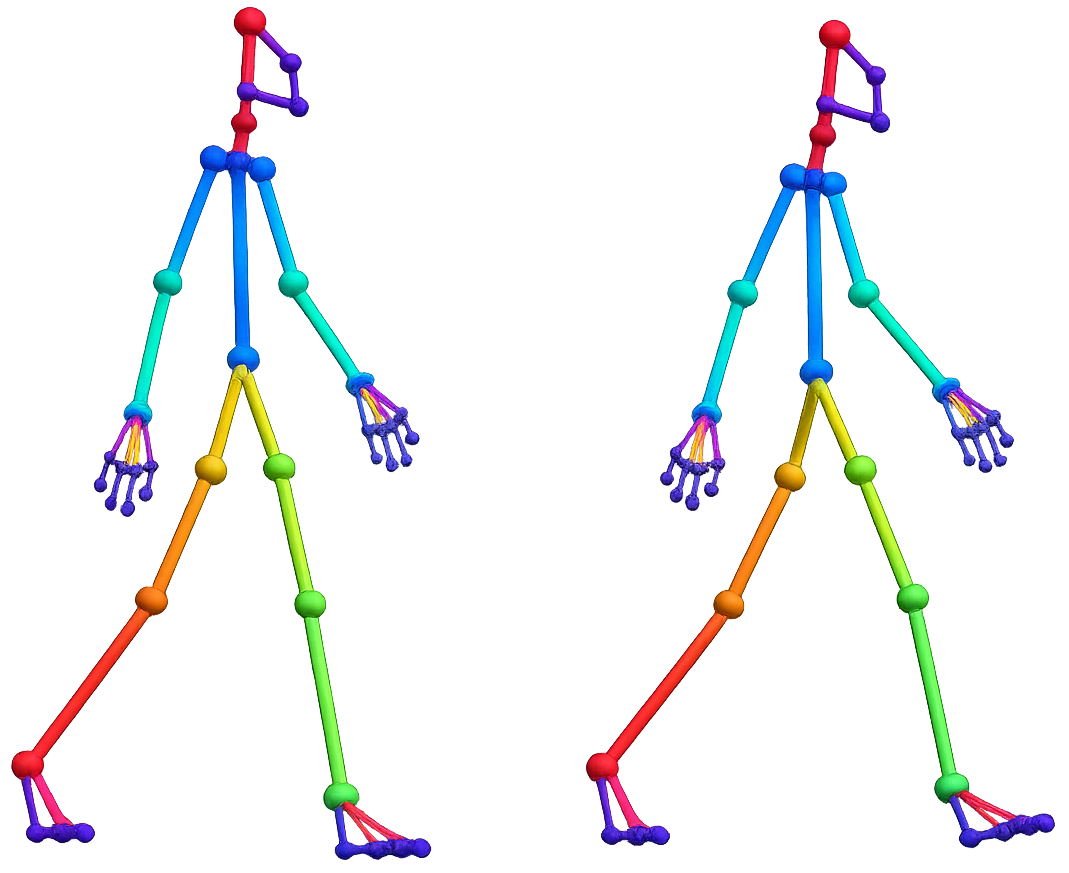}}}
\NewDocumentCommand{\HumanMesh}{ O{-0.5ex} O{0.026\linewidth}}{\raisebox{#1}{\includegraphics[width=#2]{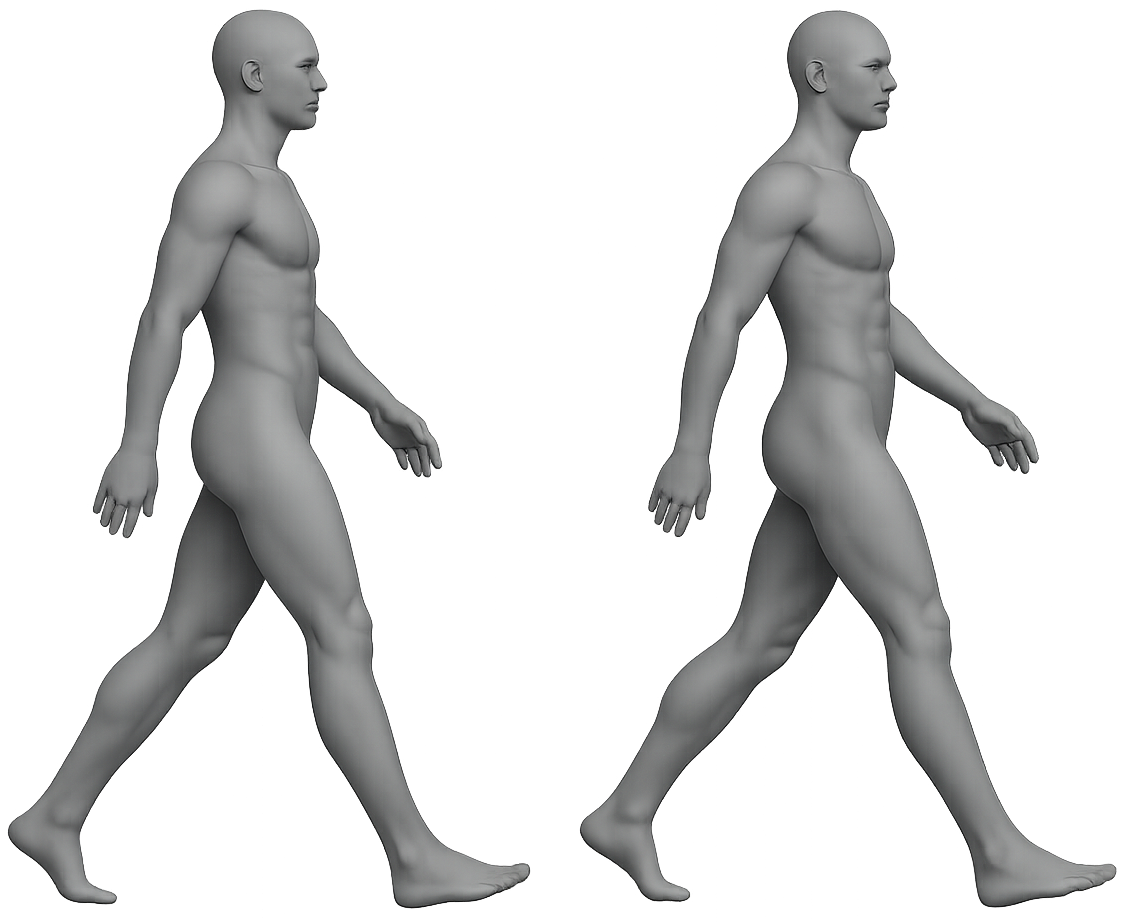}}}
\NewDocumentCommand{\HumanPointCloud}{ O{-0.5ex} O{0.026\linewidth}}{\raisebox{#1}{\includegraphics[width=#2]{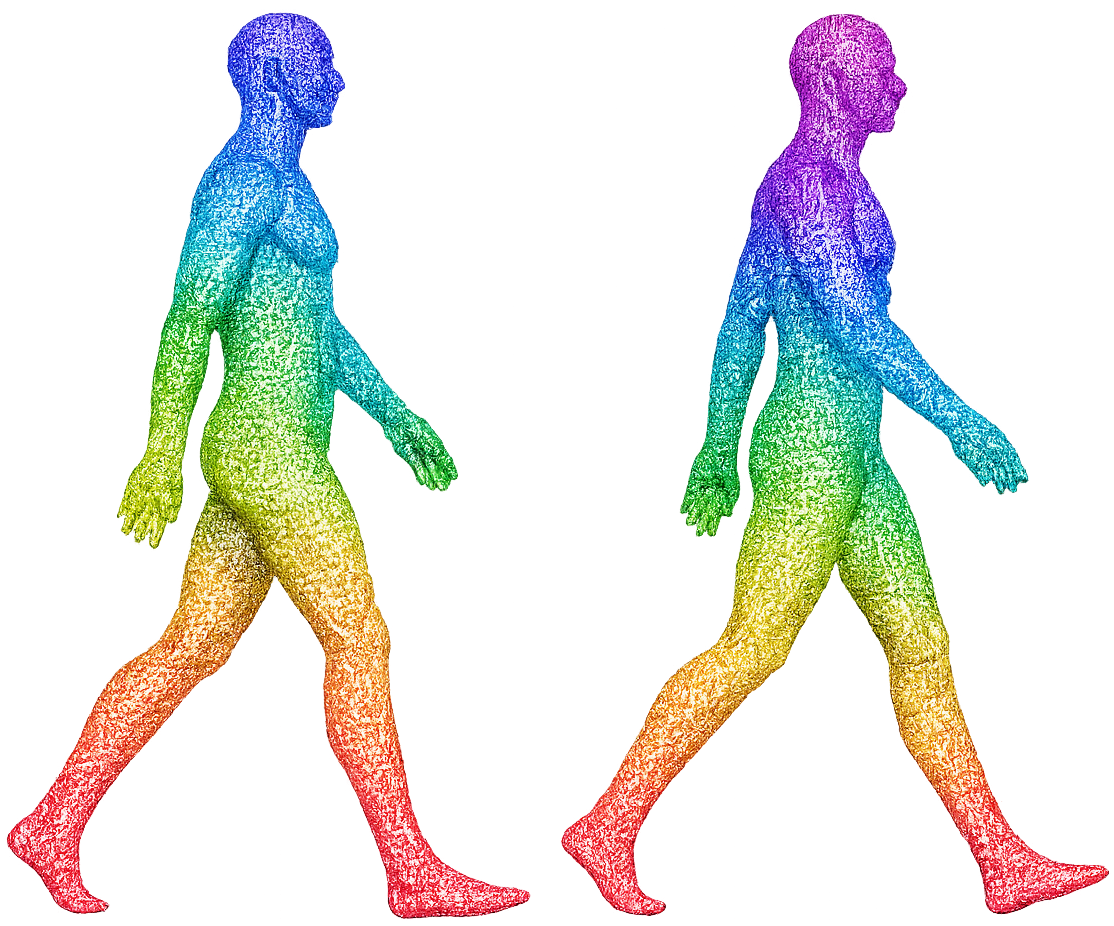}}}
\NewDocumentCommand{\HumanGS}{ O{-0.5ex} O{0.026\linewidth}}{\raisebox{#1}{\includegraphics[width=#2]{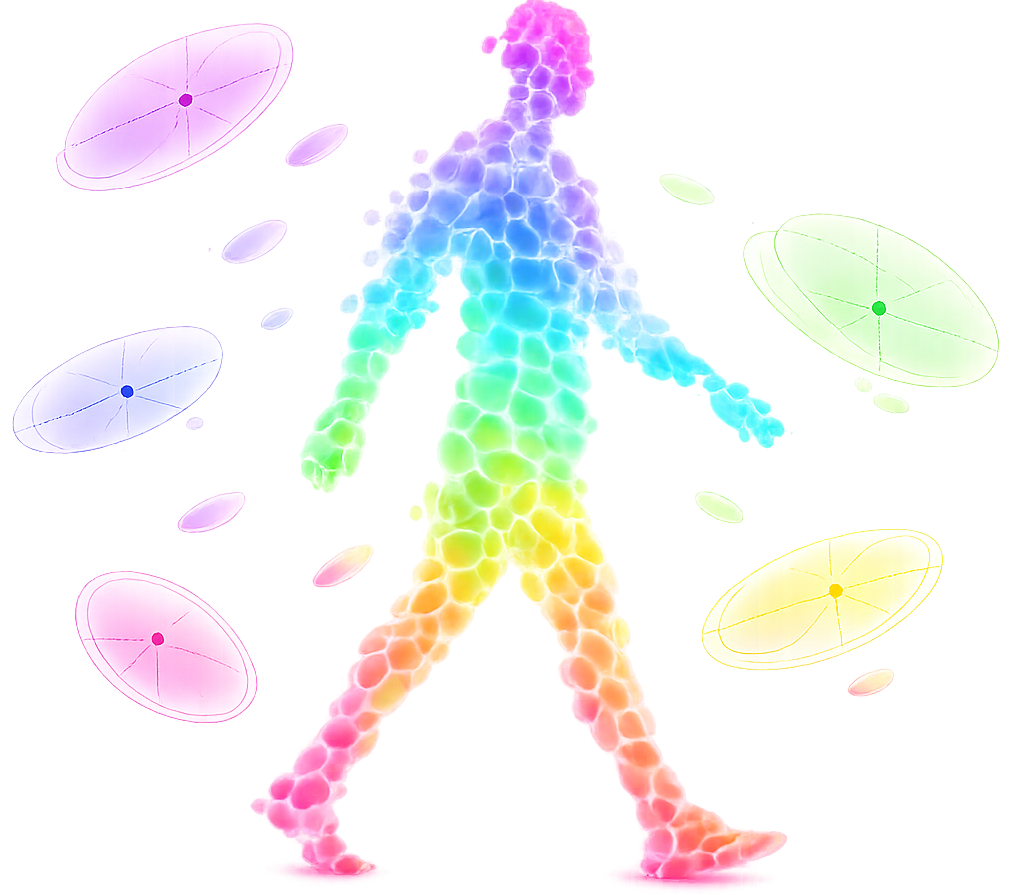}}}
\NewDocumentCommand{\Text}{ O{-0.5ex} O{0.026\linewidth}}{\raisebox{#1}{\includegraphics[width=#2]{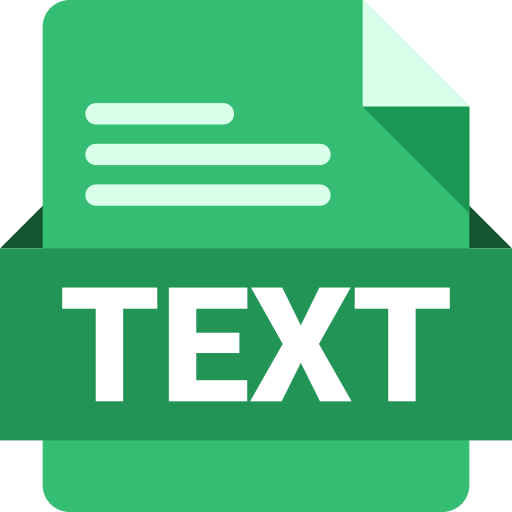}}}
\NewDocumentCommand{\Audio}{ O{-0.5ex} O{0.026\linewidth}}{\raisebox{#1}{\includegraphics[width=#2]{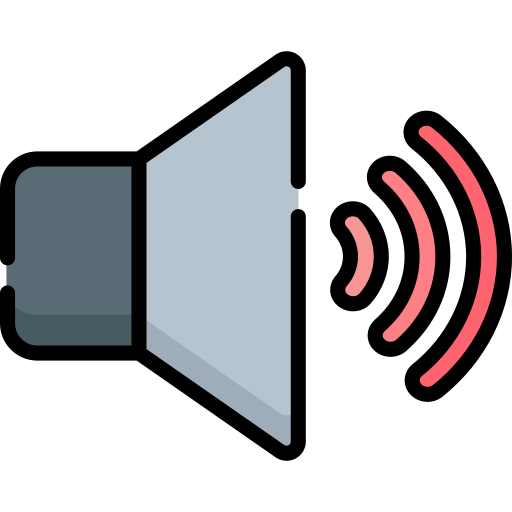}}}
\NewDocumentCommand{\WiFi}{ O{-0.5ex} O{0.026\linewidth}}{\raisebox{#1}{\includegraphics[width=#2]{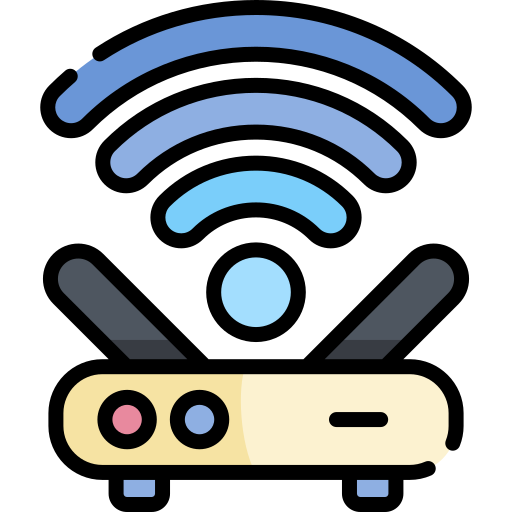}}}
\NewDocumentCommand{\Gaze}{ O{-0.5ex} O{0.026\linewidth}}{\raisebox{#1}{\includegraphics[width=#2]{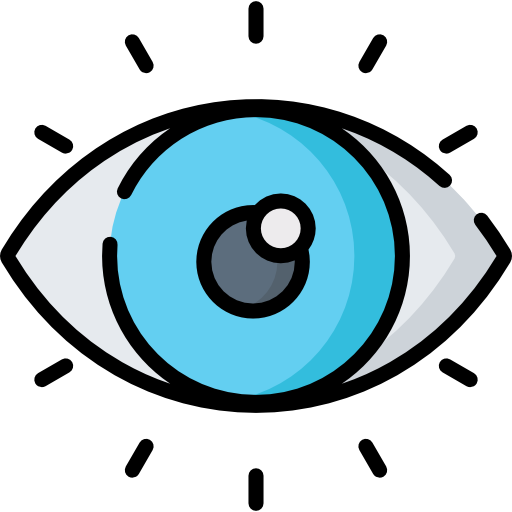}}}
\NewDocumentCommand{\Infrared}{ O{-0.5ex} O{0.026\linewidth}}{\raisebox{#1}{\includegraphics[width=#2]{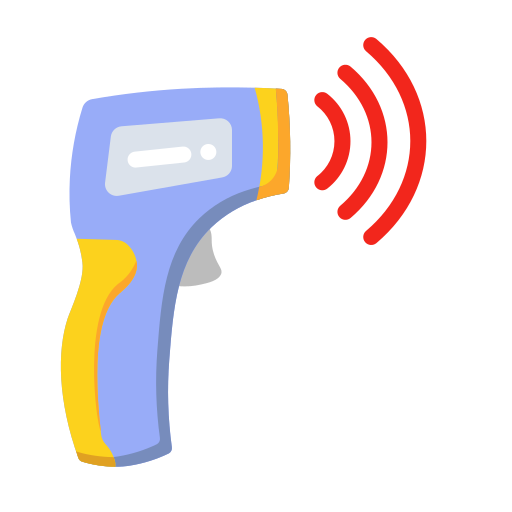}}}
\NewDocumentCommand{\Radar}{ O{-0.5ex} O{0.026\linewidth}}{\raisebox{#1}{\includegraphics[width=#2]{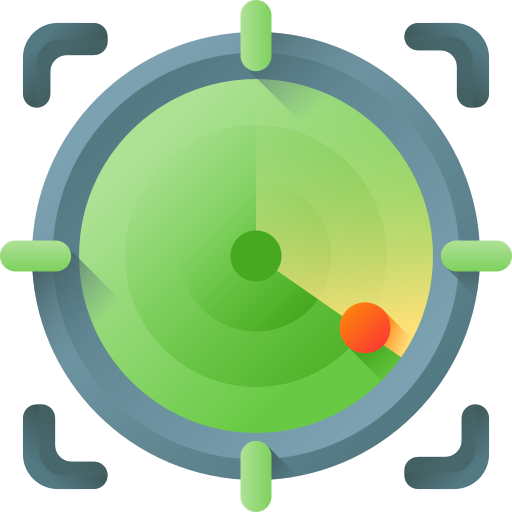}}}
\NewDocumentCommand{\LiDAR}{ O{-0.5ex} O{0.026\linewidth}}{\raisebox{#1}{\includegraphics[width=#2]{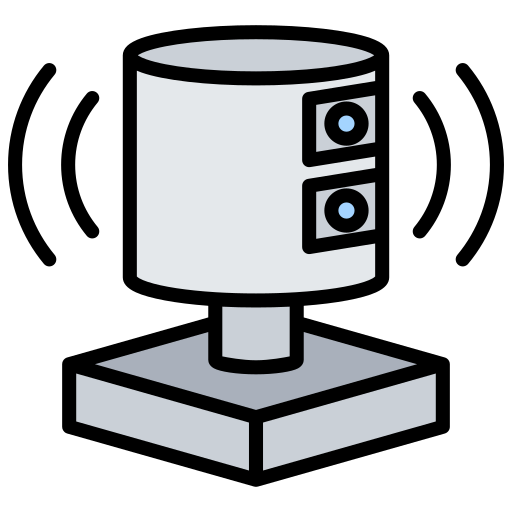}}}
\NewDocumentCommand{\Acceleration}{ O{-0.5ex} O{0.026\linewidth}}{\raisebox{#1}{\includegraphics[width=#2]{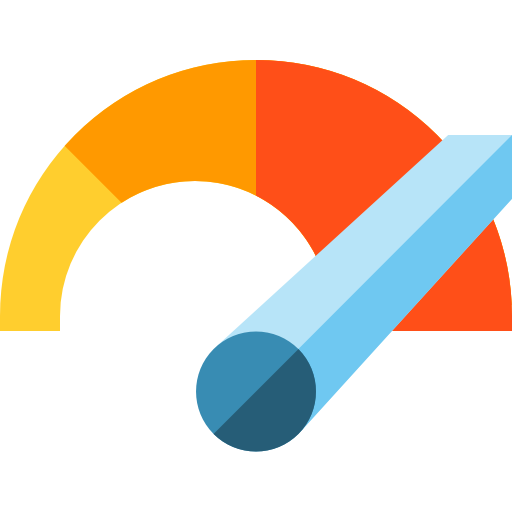}}}
\NewDocumentCommand{\Wearable}{O{-0.5ex} O{0.026\linewidth}}{\raisebox{#1}{\includegraphics[width=#2]{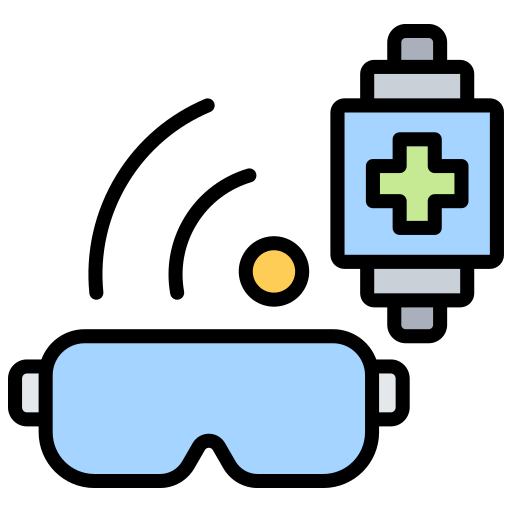}}}
\NewDocumentCommand{\Event}{ O{-0.5ex} O{0.026\linewidth}}{\raisebox{#1}{\includegraphics[width=#2]{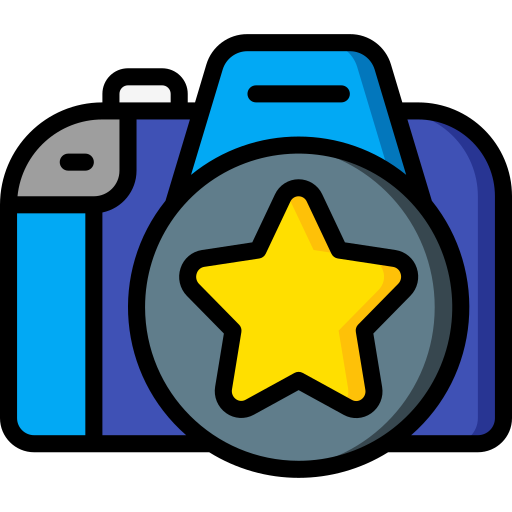}}}
\NewDocumentCommand{\Scene}{ O{-0.5ex} O{0.026\linewidth}}{\raisebox{#1}{\includegraphics[width=#2]{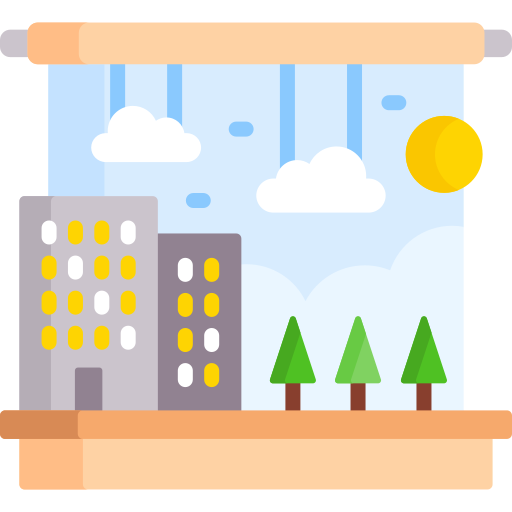}}}
\NewDocumentCommand{\Garment}{ O{-0.5ex} O{0.026\linewidth}}{\raisebox{#1}{\includegraphics[width=#2]{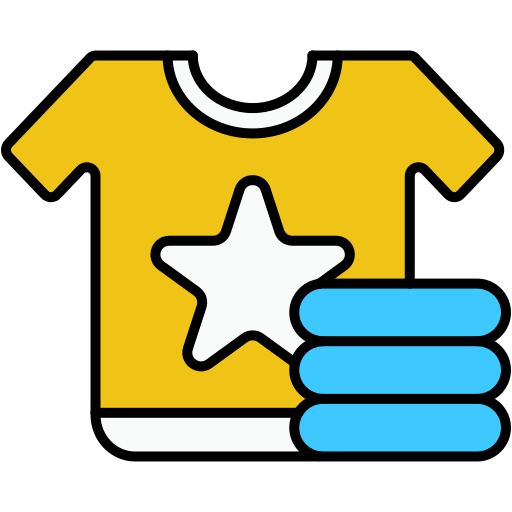}}}
\NewDocumentCommand{\HumanDepth}{ O{-0.5ex} O{0.026\linewidth}}{\raisebox{#1}{\includegraphics[width=#2]{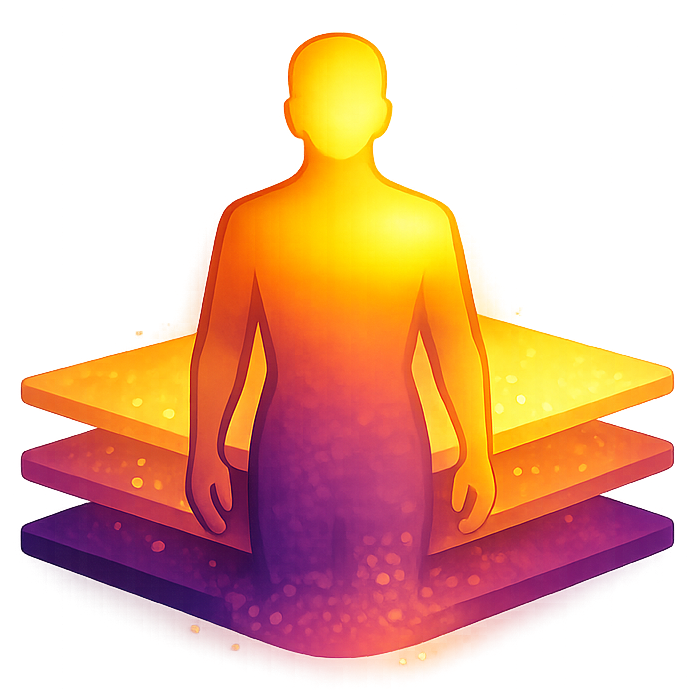}}}
\NewDocumentCommand{\HumanNormal}{ O{-0.5ex} O{0.026\linewidth}}{\raisebox{#1}{\includegraphics[width=#2]{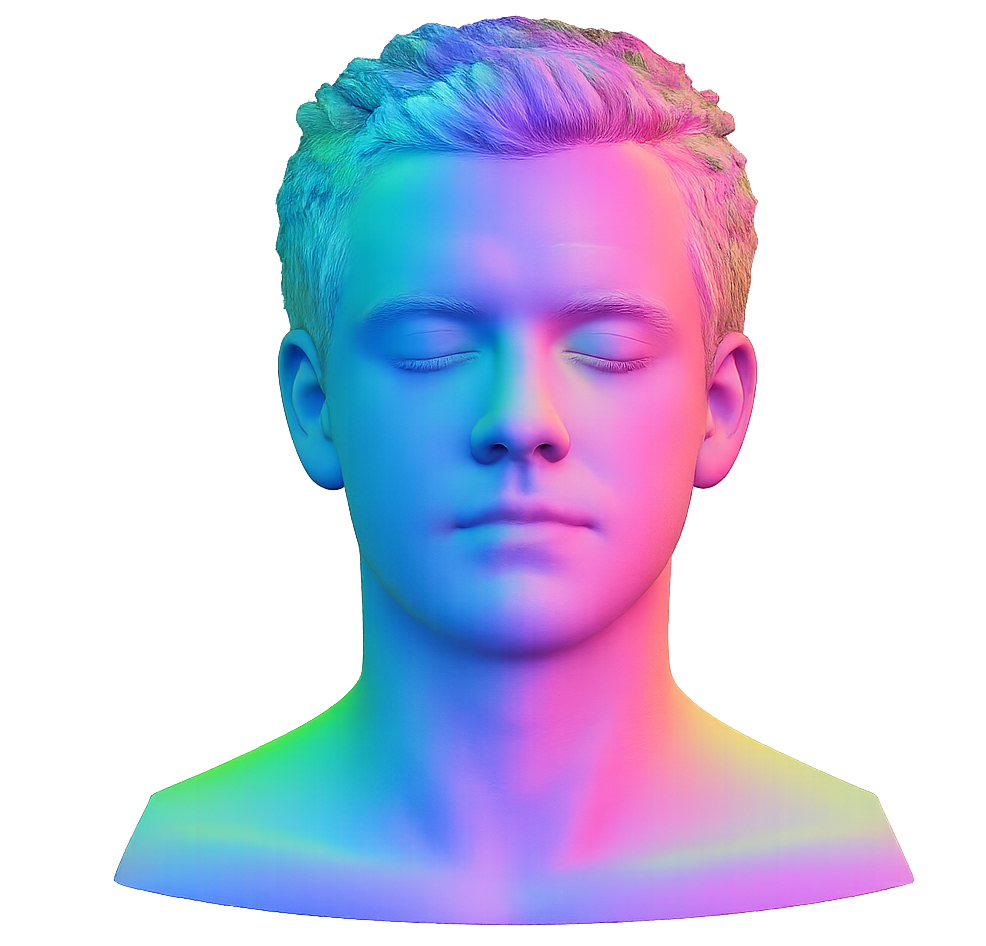}}}
\NewDocumentCommand{\HumanPhysiology}{ O{-0.5ex} O{0.026\linewidth}}{\raisebox{#1}{\includegraphics[width=#2]{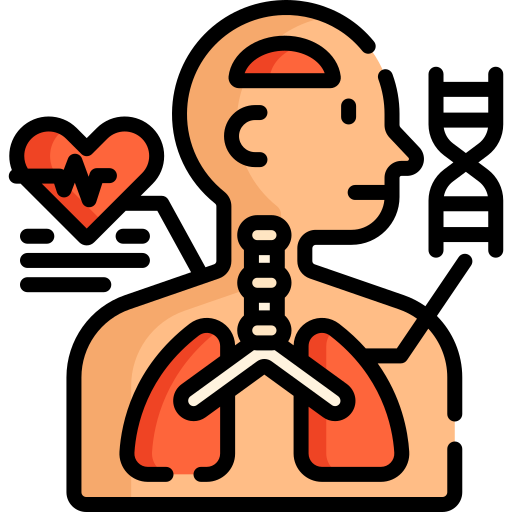}}}
\NewDocumentCommand{\HumanTactile}{ O{-0.5ex} O{0.026\linewidth}}{\raisebox{#1}{\includegraphics[width=#2]{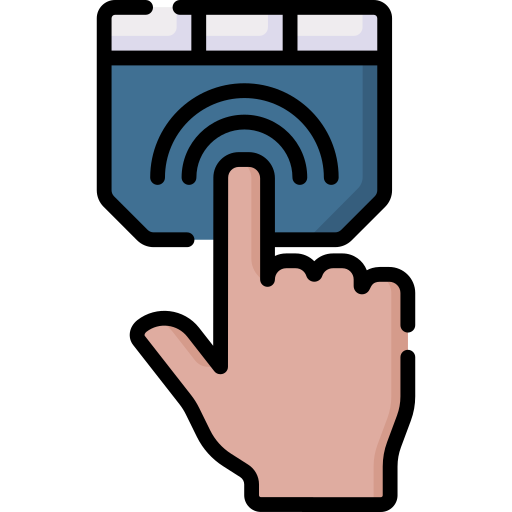}}}
\NewDocumentCommand{\ActionTrajectory}{ O{-0.5ex} O{0.026\linewidth}}{\raisebox{#1}{\includegraphics[width=#2]{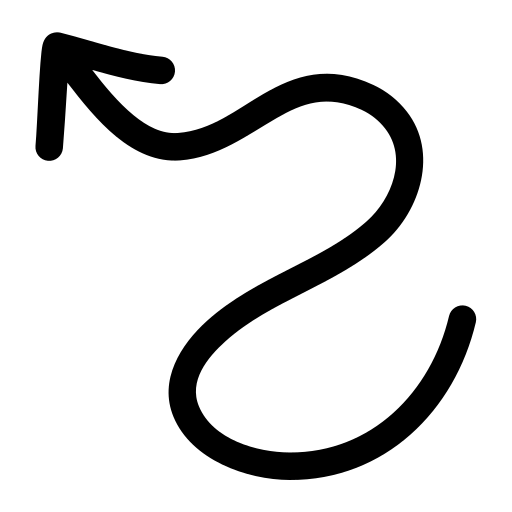}}}
\NewDocumentCommand{\ActionControl}{ O{-0.5ex} O{0.026\linewidth}}{\raisebox{#1}{\includegraphics[width=#2]{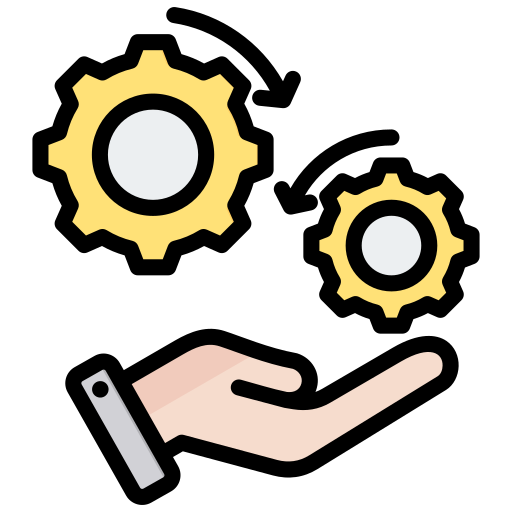}}}

\NewDocumentCommand{\HumanThemselves}{ O{-0.5ex} O{0.026\linewidth}}{\raisebox{#1}{\includegraphics[width=#2]{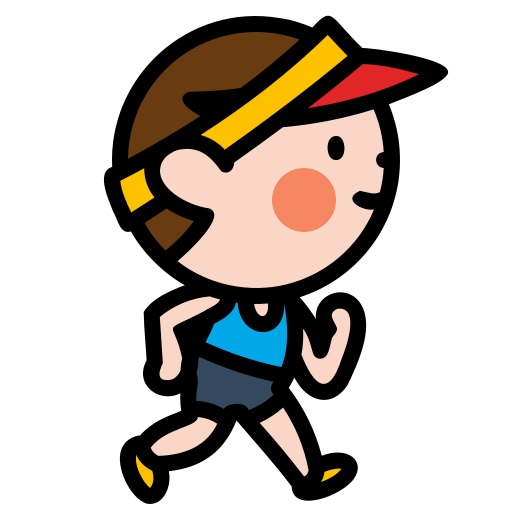}}}
\NewDocumentCommand{\VisualAppearance}{ O{-0.5ex} O{0.026\linewidth}}{\raisebox{#1}{\includegraphics[width=#2]{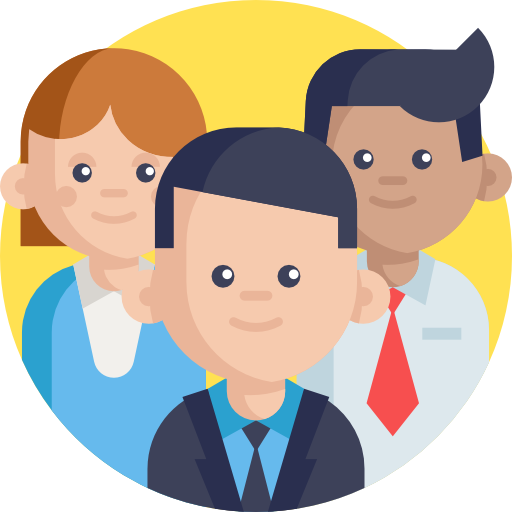}}}
\NewDocumentCommand{\StructuralGeometry}{ O{-0.5ex} O{0.026\linewidth}}{\raisebox{#1}{\includegraphics[width=#2]{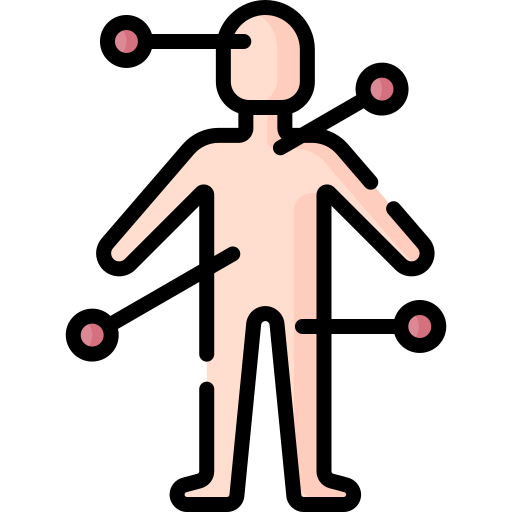}}}
\NewDocumentCommand{\KinematicDynamics}{ O{-0.5ex} O{0.026\linewidth}}{\raisebox{#1}{\includegraphics[width=#2]{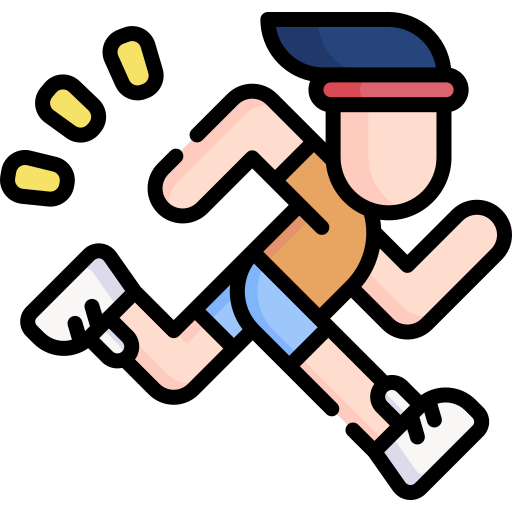}}}
\NewDocumentCommand{\InteractionModeling}{ O{-0.5ex} O{0.026\linewidth}}{\raisebox{#1}{\includegraphics[width=#2]{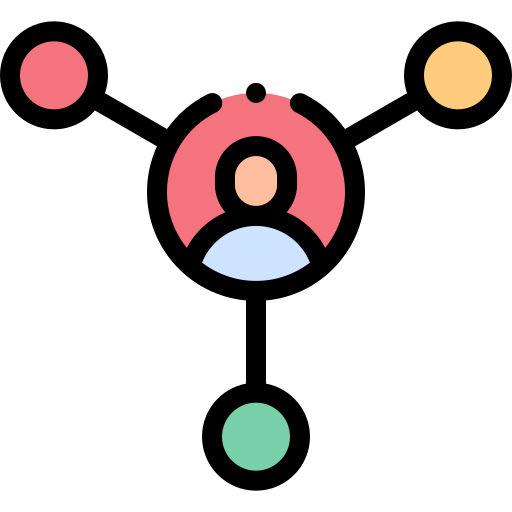}}}
\NewDocumentCommand{\WorldSimulation}{ O{-0.5ex} O{0.026\linewidth}}{\raisebox{#1}{\includegraphics[width=#2]{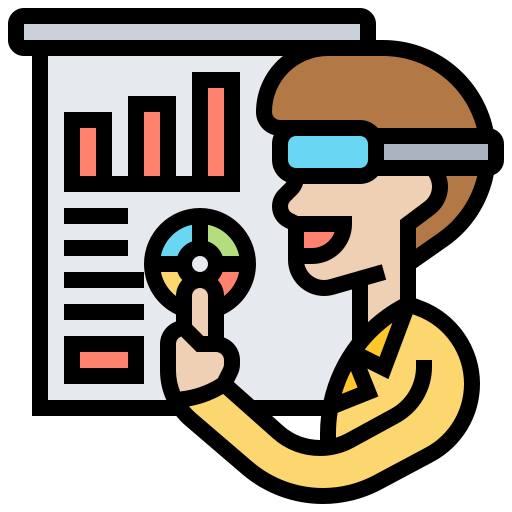}}}
\NewDocumentCommand{\EmbodiedAgent}{ O{-0.5ex} O{0.026\linewidth}}{\raisebox{#1}{\includegraphics[width=#2]{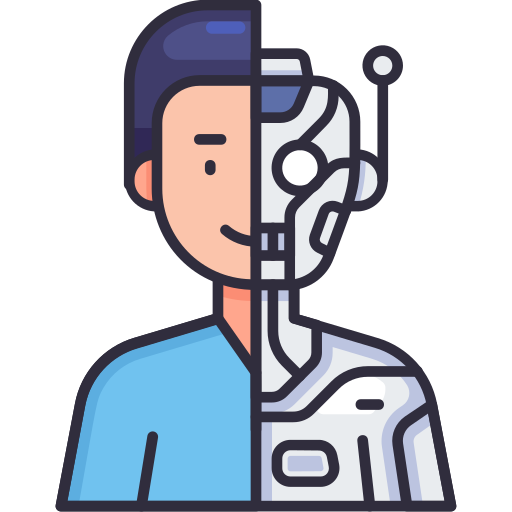}}}

\NewDocumentCommand{\FMnone}{ O{-0.2ex} }{\raisebox{#1}{\begin{tikzpicture}[scale=0.12, baseline=(current bounding box.center)]\draw[gray] (0,0) rectangle (1,1);\end{tikzpicture}}}
\NewDocumentCommand{\FMlimited}{ O{-0.2ex} }{\raisebox{#1}{\begin{tikzpicture}[scale=0.12, baseline=(current bounding box.center)]\fill (0,0) rectangle (1,1);\end{tikzpicture}}}
\NewDocumentCommand{\FMmoderate}{ O{-0.2ex} }{\raisebox{#1}{\begin{tikzpicture}[scale=0.12, baseline=(current bounding box.center)]\fill (0,0) rectangle (1,1);\fill (1.3,0) rectangle (2.3,1.4);\end{tikzpicture}}}
\NewDocumentCommand{\FMextensive}{ O{-0.2ex} }{\raisebox{#1}{\begin{tikzpicture}[scale=0.12, baseline=(current bounding box.center)]\fill (0,0) rectangle (1,1);\fill (1.3,0) rectangle (2.3,1.4);\fill (2.6,0) rectangle (3.6,1.8);\end{tikzpicture}}}

\NewDocumentCommand{\NumberOne}{ O{-0.5ex} O{0.026\linewidth}}{\raisebox{#1}{\includegraphics[width=#2]{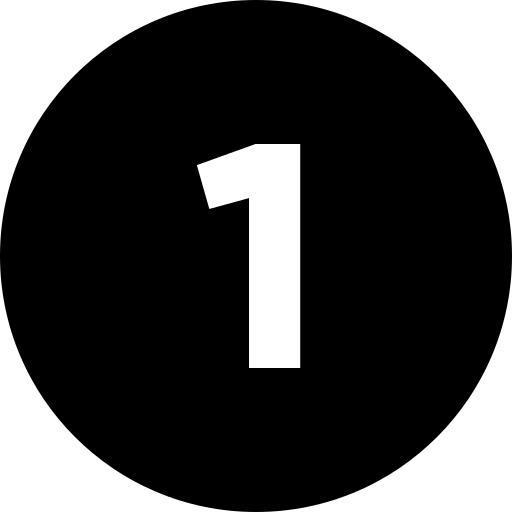}}}
\NewDocumentCommand{\NumberTwo}{ O{-0.5ex} O{0.026\linewidth}}{\raisebox{#1}{\includegraphics[width=#2]{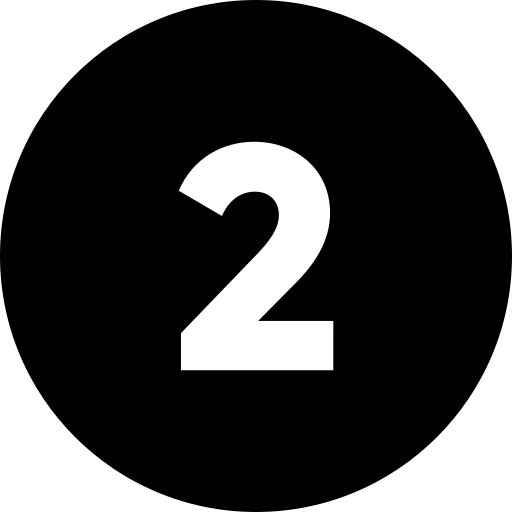}}}
\NewDocumentCommand{\NumberThree}{ O{-0.5ex} O{0.026\linewidth}}{\raisebox{#1}{\includegraphics[width=#2]{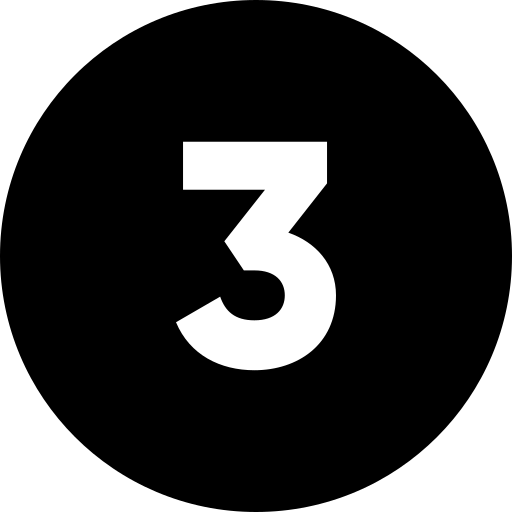}}}
\NewDocumentCommand{\NumberFour}{ O{-0.5ex} O{0.026\linewidth}}{\raisebox{#1}{\includegraphics[width=#2]{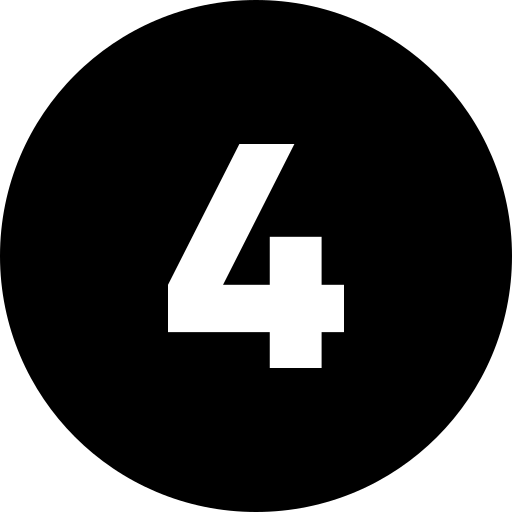}}}
\NewDocumentCommand{\NumberFive}{ O{-0.5ex} O{0.026\linewidth}}{\raisebox{#1}{\includegraphics[width=#2]{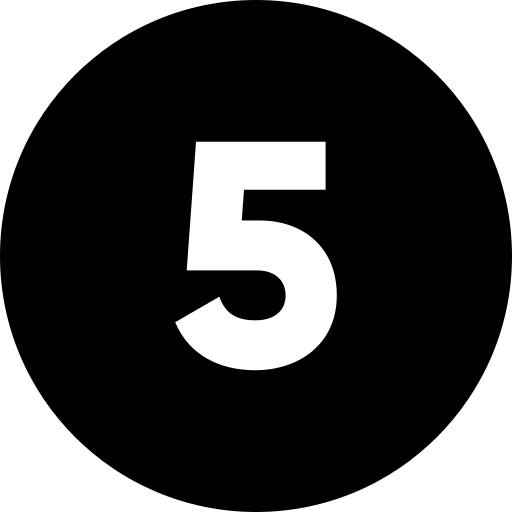}}}

\newcommand{\Yes}{\ding{51}}
\newcommand{\No}{\ding{55}}
\newcommand{\NA}{\textbf{--}}
\newcommand{\Estimated}{\textcolor{crEstimated}{$\dag$}}

\newcommand{\Real}{\textcolor{crReal}{$\S$}}
\newcommand{\Synthetic}{\textcolor{crSynthetic}{$\P$}}

\NewDocumentCommand{\TrainScratch}{ O{0pt} }{\crbx[#1]{crTrainScratch}{\textcolor{white}{\textbf{\textsf{S}}}}}
\NewDocumentCommand{\TrainFull}{ O{0pt} }{\crbx[#1]{crTrainFull}{\textcolor{white}{\textbf{\textsf{F}}}}}
\NewDocumentCommand{\TrainParameter}{ O{0pt} }{\crbx[#1]{crTrainParameter}{\textcolor{white}{\textbf{\textsf{P}}}}}
\NewDocumentCommand{\TrainInstruction}{ O{0pt} }{\crbx[#1]{crTrainInstruction}{\textcolor{white}{\textbf{\textsf{I}}}}}
\NewDocumentCommand{\TrainReward}{ O{0pt} }{\crbx[#1]{crTrainReward}{\textcolor{white}{\textbf{\textsf{R}}}}}
\NewDocumentCommand{\TrainTask}{ O{0pt} }{\crbx[#1]{crTrainTask}{\textcolor{white}{\textbf{\textsf{T}}}}}
\NewDocumentCommand{\TrainDistillation}{ O{0pt} }{\crbx[#1]{crTrainDistillation}{\textcolor{white}{\textbf{\textsf{D}}}}}

\NewDocumentCommand{\InferenceSemantic}{ O{0pt} }{\crbx[#1]{crInferenceSemantic}{\textcolor{white}{\textbf{\textsf{S}}}}}
\NewDocumentCommand{\InferenceGuided}{ O{0pt} }{\crbx[#1]{crInferenceGuided}{\textcolor{white}{\textbf{\textsf{G}}}}}
\NewDocumentCommand{\InferenceIterative}{ O{0pt} }{\crbx[#1]{crInferenceIterative}{\textcolor{white}{\textbf{\textsf{I}}}}}
\NewDocumentCommand{\InferencePreference}{ O{0pt} }{\crbx[#1]{crInferencePreference}{\textcolor{white}{\textbf{\textsf{P}}}}}
\NewDocumentCommand{\InferenceRetrieve}{ O{0pt} }{\crbx[#1]{crInferenceRetrieve}{\textcolor{white}{\textbf{\textsf{R}}}}}

\NewDocumentCommand{\DataSelfCaptured}{ O{0pt} }{\crcc[#1]{crDataSelfCaptured}{\textcolor{black}{\textbf{\textsf{C}}}}}
\NewDocumentCommand{\DataPublicCurated}{ O{0pt} }{\crcc[#1]{crDataPublicCurated}{\textcolor{black}{\textbf{\textsf{P}}}}}
\NewDocumentCommand{\DataWebCollected}{ O{0pt} }{\crcc[#1]{crDataWebCollected}{\textcolor{black}{\textbf{\textsf{W}}}}}
\NewDocumentCommand{\DataSynthesized}{ O{0pt} }{\crcc[#1]{crDataSynthesized}{\textcolor{black}{\textbf{\textsf{S}}}}}

\NewDocumentCommand{\Ego}{ O{0pt} }{\crbxrect[#1]{crEgo}{\textcolor{black}{\textbf{\textsf{Ego}}}}}
\NewDocumentCommand{\Exo}{ O{0pt} }{\crbxrect[#1]{crExo}{\textcolor{black}{\textbf{\textsf{Exo}}}}}

\NewDocumentCommand{\SMA}{ O{0pt} }{\crbxrect[#1]{crSMA}{\textcolor{white}{\textbf{\textsf{SMA}}}}}
\NewDocumentCommand{\SFA}{ O{0pt} }{\crbxrect[#1]{crSFA}{\textcolor{white}{\textbf{\textsf{SFA}}}}}
\NewDocumentCommand{\IGA}{ O{0pt} }{\crbxrect[#1]{crIGA}{\textcolor{white}{\textbf{\textsf{IGA}}}}}
\NewDocumentCommand{\HA}{ O{0pt} }{\crbxrect[#1]{crHA}{\textcolor{white}{\textbf{\textsf{HA}}}}}

\tcbset{
    insight common/.style={
        enhanced,
        boxrule=1.5pt,
        arc=3pt,
        left=10pt,
        right=10pt,
        top=12pt,
        bottom=6pt,
        boxsep=6pt,
        fonttitle=\bfseries,
        coltitle=black,
        attach boxed title to top left={xshift=10pt,yshift=-2mm},
        varwidth boxed title=0.92\linewidth,
        before={\par\Needspace{0.26\textheight}},
        after={\par\medskip}
    }
}
\newtcolorbox{insightbox}[1][]{
    insight common,
    colback=box_midyellow!5,
    colframe=box_midyellow!50,
    colbacktitle=box_midyellow!50,
    title={\small #1},
    boxed title style={
        colback=box_midyellow!50,
        colframe=box_midyellow!50,
        boxrule=0pt,
        arc=2pt,
        outer arc=2pt,
        left=6pt,
        right=6pt,
        top=2pt,
        bottom=2pt,
    }
}

\title{Human-Centric Intelligence in the Era of Foundation Models: A Survey}

\author[1,\dagger]{Yang Chen}
\author[1]{Tianqi Wang}
\author[1]{Xiaorui Jiang}
\author[1]{Yilei Man}
\author[1]{Yihua Shao}
\author[2]{Mengyuan Liu}
\author[3]{Zhi Chen}
\author[4]{Xiaofeng Cao}
\author[5]{Qibin Zhao}
\author[6]{Chi Harold Liu}
\author[7]{Albert Y. Zomaya}
\author[8]{Nicu Sebe}
\author[9]{Jingren Zhou}
\author[10]{Dacheng Tao}
\author[11]{Song Guo}
\author[1,\dagger,\ddagger]{Jingcai Guo}

\affiliation[1]{The Hong Kong Polytechnic University}
\affiliation[2]{Peking University}
\affiliation[3]{University of Southern Queensland}
\affiliation[4]{Tongji University}
\affiliation[5]{RIKEN Center for Advanced Intelligence Project}
\affiliation[6]{Beijing Institute of Technology}
\affiliation[7]{The University of Sydney}
\affiliation[8]{University of Trento}
\affiliation[9]{Alibaba Group}
\affiliation[10]{Nanyang Technological University}
\affiliation[11]{Hong Kong University of Science and Technology}

\contribution[\dagger]{\textbf{Project Lead}}
\contribution[\ddagger]{\textbf{Corresponding Author}}

\abstract{
Human-centric intelligence is evolving in the foundation-model era, with growing emphasis on scale, transferability, and general-purpose modeling. Yet it has not fully integrated with foundation models to achieve the comparable progress seen in them. More importantly, recent advances across this broad landscape remain fragmented across tasks, modalities, and research communities, leaving their intrinsic conceptual and methodological connections unclear. To bridge these divides and rethink human-centric intelligence in the foundation-model era, we introduce a \textbf{full-spectrum human context taxonomy} that integrates six interconnected levels by viewing humans as \textbf{observable subjects} through visual appearance and spatial geometry, as \textbf{dynamic actors} through kinematic dynamics and interaction modeling, and as \textbf{situated agents} through world simulation and embodied agency. We next present the methodological foundations of the field, covering human-centric data families, computational architecture paradigms, and representative training and inference optimization strategies. We then systematically review representative methods across these levels and organize the associated datasets, benchmarks, and evaluation metrics. We further discuss open challenges and promising research directions toward human-centric intelligence that is scalable, trustworthy, physically grounded, and deployable, aiming to provide a coherent framework and practical reference for advancing the field. Finally, we provide a systematically organized and continuously updated collection of human-centric AI literature and resources on our project page.
}

\date{\today}
\metadata[\faEnvelope ~~Main Contact]{\email{cs-yang.chen@connect.polyu.hk},~\email{jc-jingcai.guo@polyu.edu.hk}}
\metadata[\faGithub ~~GitHub Repo]{\href{https://github.com/cseeyangchen/Human-Centric-AI}{Human-Centric AI Resources}}
\metadata[\faHome ~~Homepage]{\href{https://cseeyangchen.github.io/Human-Centric-AI/homepage/}{Project Website}}

\begin{document}

\maketitle

\begin{figure}[h]
    \centering
    \includegraphics[width=\linewidth]{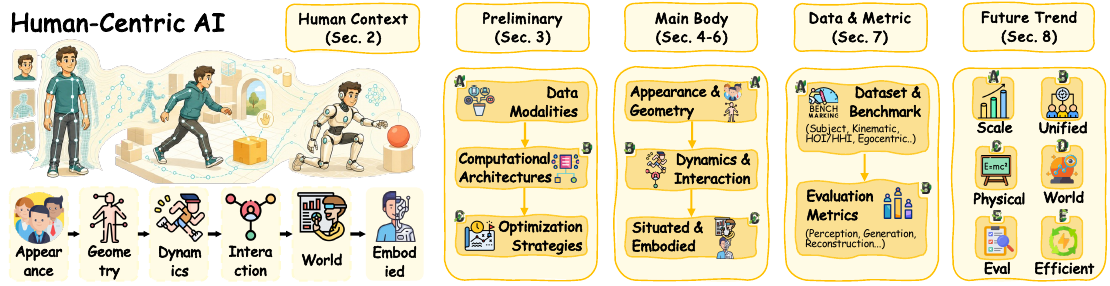}
    \caption{\textbf{Outline of the survey.} This work focuses on \textbf{Human-Centric Intelligence} in the foundation-model era by characterizing humans as \textit{observable subjects}, \textit{dynamic actors}, and \textit{situated agents} across six human context levels (Sec.~\ref{sec::taxonomy}). Built on this taxonomy, we summarize data modalities, computational architectures, and training/inference optimization strategies (Sec.~\ref{sec::preliminary}). We then review main methodologies across six levels (Sec.~\ref{sec::appearance_geometry}--\ref{sec::situated_embodied}). Finally, we summarize datasets, benchmarks, and metrics (Sec.~\ref{sec::datasets_benchmarks_metrics}), followed by open challenges and future directions (Sec.~\ref{sec::future}).}
\label{fig::teaser} 
\end{figure}



\section{Introduction} 
\label{sec::introduction}

\textbf{Full-Spectrum Human-Centric Intelligence.}
Human-centric intelligence encompasses computational capabilities centered on humans and the diverse contexts in which they are observed, involved, and situated in action. Since humans are multifaceted, no single computational perspective is sufficient to capture the full scope of these capabilities. Depending on the question being addressed, a system may emphasize what is observable about a person, characterize how human states evolve in relation to their surroundings, or model how human experience participates in world dynamics and executable behavior. These contexts require distinct but closely related forms of human modeling. Human representations provide the computational foundation for this modeling, while human-centric intelligence concerns how these representations are learned, connected, generalized, and transformed into useful capabilities. Viewed collectively, these research directions define a full spectrum of human-centric intelligence and support applications including autonomous driving\cite{caesar2020nuscenes,grigorescu2020survey}, assistive systems\cite{yang2025socialmind,qiu2023investigating}, and digital humans\cite{cai2025towards,jin2026sentiavatar}. Although this breadth has long characterized human-centric research, the paradigms through which these capabilities are developed have continually evolved.

\textbf{The Foundation-Model Era Shift.}
This evolution can be understood through three broad stages. Early work encoded human characteristics through hand-crafted descriptors~\cite{sargano2017comprehensive}, producing interpretable features whose applicability was largely limited by predefined assumptions. Deep learning shifted human modeling from manually designed features toward data-driven learning, substantially increasing modeling capacity while still being constrained by particular datasets and objectives~\cite{khan2020hand,pham2022video}. More recently, the foundation-model era has shifted the emphasis toward large-scale pretraining, reusable model backbones, multimodal interfaces, and adaptation across tasks and domains~\cite{bommasani2021opportunities,awais2025foundation}. This shift creates new opportunities for human-centric intelligence by enabling broader transfer and more general-purpose modeling, while also exposing human-specific challenges. This ongoing transformation motivates this survey to systematically review the concepts, methods, and emerging trends of human-centric intelligence in the era of foundation models.

\textbf{Comparison with Other Human-Centric Surveys.} 
Recent surveys have provided detailed accounts of focused areas within human-centric research, such as pose estimation~\cite{zheng2023deep,tian2023recovering}, action recognition~\cite{zhang2026self,miao2025wi,gu2021survey,kong2022human,vahdani2022deep}, human motion and video generation~\cite{sui2026survey,fan20253d,zhu2023human,xue2025human}, and human-object interaction~\cite{lin2026hand}. By organizing the literature around individual tasks, these surveys offer in-depth analyses of specific research problems but leave the connections among different human contexts and capabilities largely unexplored. More importantly, except for the HOI survey~\cite{lin2026hand}, the foundation-model era does not constitute their central focus, as their discussions primarily reflect advances within conventional deep-learning paradigms. A recent effort~\cite{tang2025human} has begun to review human-centric foundation models, but its coverage remains largely centered on humans themselves and encompasses a relatively limited body of work. Consequently, to the best of our knowledge, no existing survey has systematically connected the full spectrum of human-centric intelligence and examined its methodological evolution in the foundation-model era.

\textbf{Scope and Contributions.}
This survey examines the evolving landscape of human-centric intelligence in the foundation-model era. Our scope covers methods that characterize humans as observable subjects, dynamic actors, and situated agents. It includes both human-specialized foundation models and recent methods that reflect the broader foundation-era shift through scalable learning, reusable pretraining, multimodal interfaces, cross-task generalization, or transferable human priors. We exclude broader topics such as human-computer interaction, human-oriented language modeling, and policy studies of human-centered AI. To move beyond an organization based solely on architectures or applications, we introduce a human context taxonomy that integrates six interconnected levels through three complementary perspectives. The main contributions of this survey are summarized as follows:
\begin{itemize}
    \item \textbf{First Full-Spectrum Survey.} We develop a human context taxonomy that connects six levels of human-centric intelligence: visual appearance, spatial geometry, kinematic dynamics, interaction modeling, world simulation, and embodied agency. This taxonomy provides a unified perspective for systematically organizing the literature.
    \item \textbf{Methodological Foundations and Advances.} We present the methodological foundations of human-centric intelligence by organizing its data families, computational architecture paradigms, and training and inference optimization strategies. Guided by the taxonomy, we further examine how representative methods develop human-centric capabilities in the foundation-modal era.
    \item \textbf{Evaluation Protocols.} We organize representative datasets and benchmarks and consolidate commonly used evaluation metrics, clarifying the empirical foundations on which different human-centric capabilities are developed and evaluated.
    \item \textbf{Frontiers and Resources.} We discuss open challenges and promising directions for future research, and publicly release systematically organized resources to support continued progress in human-centric AI.
\end{itemize}

\textbf{Roadmap of this Survey.}
The remainder of this survey is organized as follows. Sec.~\ref{sec::taxonomy} introduces the human context taxonomy and defines six interconnected levels through three perspectives on humans. Sec.~\ref{sec::preliminary} presents the methodological foundations of the field by summarizing human-centric data families, computational architecture paradigms, and optimization strategies. Sec.~\ref{sec::appearance_geometry} reviews methods concerned with visible human properties and spatial body structure. Sec.~\ref{sec::dynamics_interaction} examines temporal human behavior and the relations among humans, objects, scenes, and other people. Sec.~\ref{sec::situated_embodied} focuses on human-centered world simulation and the transformation of human knowledge and experience into embodied capabilities.  Sec.~\ref{sec::datasets_benchmarks_metrics} consolidates the datasets, benchmarks, and metrics used to develop and evaluate these methods. Finally, Sec.~\ref{sec::future} discusses open challenges and future directions, and Sec.~\ref{sec::conclusion} concludes the survey.
\section{Human Context Taxonomy} 
\label{sec::taxonomy}

\begin{figure}[t]
    \centering
    \includegraphics[width=\linewidth]{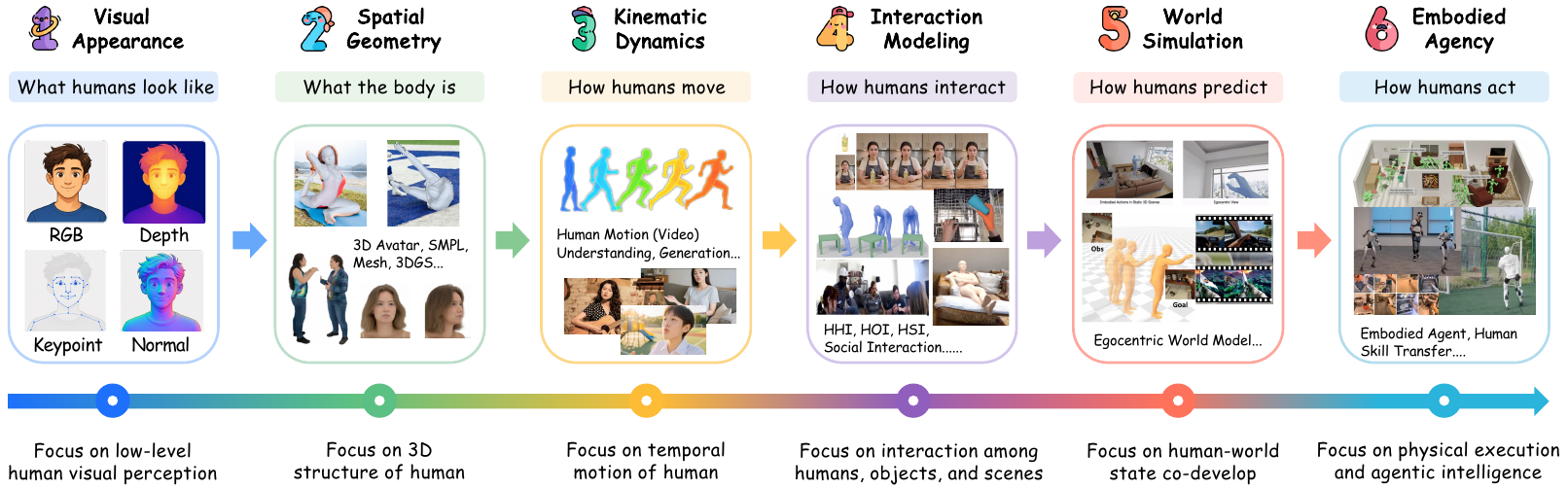}
    \caption{\textbf{Full-spectrum human context taxonomy for human-centric intelligence.} We organize human-centric intelligence along a progressive expansion of representational scope: from modeling humans as visually observable subjects, to recovering their spatial body structure, capturing their motion over time, reasoning about their interactions with surrounding entities, predicting human-centered world dynamics, and finally transferring human experience into physical execution and agentic intelligence. The images are originally shown in~\cite{yang2026sam,li2026large,gan2025omniavatar,cai2026vihoi,li2026omni,liu2026open,bansal2026towards,kim2026dexterous,wang2025dreamactor,xie2026generated,tran2026egoexo,deng2025human,qi2026humanoid,gao2026dreamdojo}.}
\label{fig::taxonomy}
\end{figure}

Human-centric intelligence spans a broad range of tasks, modalities, and model designs, making it difficult to organize the literature solely according to individual tasks, applications, or architecture families~\cite{tang2025human}. To provide a coherent organization, we introduce a human context taxonomy that characterizes each line of work according to its \textit{\textbf{primary human context}}, as shown in Fig.~\ref{fig::taxonomy}. Here, ``context'' refers to the modeling scope within which human-centric capabilities are developed. We view humans from three perspectives: as \textit{observable subjects} defined by visible and structural properties, as \textit{dynamic actors} characterized by motion and relational behavior, and as \textit{situated agents} whose experience is connected to evolving worlds and executable action. The six levels introduced below are interconnected rather than mutually exclusive. For methods spanning multiple contexts, we assign the primary level according to the principal modeling target and evaluation objective.

\subsection{Observable Subjects}
\label{sec::observable_subjects}

The perspective of observable subjects concerns human properties that can be directly perceived from visual evidence or recovered as explicit body structure. We distinguish \textit{visual appearance}, which remains primarily grounded in image-space characteristics, from \textit{spatial geometry}, which makes the organization of the human body explicit in two-dimensional or three-dimensional space. Together, these levels provide the perceptual and structural foundations on which broader human-centric capabilities are built.

\subsubsection{Visual Appearance} 
\label{sec::level1}

\textbf{Conceptual Scope.}
At this level, human-centric intelligence focuses on the visible characteristics of humans in images and videos. These characteristics range from low-level body and surface cues to identity-bearing attributes and controllable appearance. Visual appearance therefore concerns what humans look like and how their visible properties can be perceived, distinguished, interpreted, and synthesized.

\textbf{Representative Tasks.}
Representative tasks can be organized into \textit{generalist human perception}, \textit{identity recognition and retrieval}, and \textit{controllable human generation}. Generalist human perception covers pose estimation~\cite{khirodkar2024sapiens}, human parsing~\cite{khirodkarsapiens2}, depth prediction and surface-normal estimation~\cite{wang2026thfm}, and facial analysis~\cite{narayan2025facexformer} through shared visual features. Identity recognition and retrieval includes person re-identification~\cite{he2025instruct}, face recognition~\cite{you2025lvface}, multimodal person retrieval~\cite{zuo2026reid5o}, and language-guided person search~\cite{niu2025chatreid}. Controllable human generation encompasses text-to-image synthesis~\cite{li2024cosmicman}, identity-preserving generation~\cite{nam2025visual}, human editing~\cite{cheong2023upgpt}, virtual try-on~\cite{lee2025voost}, and multimodal digital-human creation~\cite{bao2026archon}. In the foundation-model era, these previously task-specific capabilities are increasingly connected through human-specialized large-scale pretraining, multimodal query interfaces, and unified multi-task backbones.

\begin{figure}[!t]
    \centering
    \includegraphics[width=\linewidth]{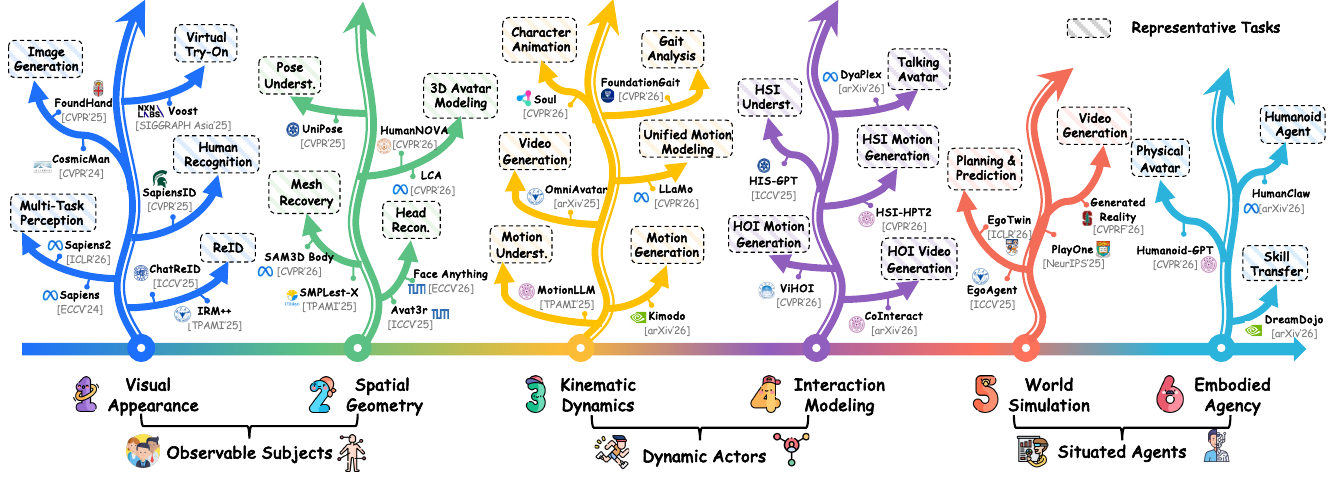}
    \caption{\textbf{Representative task families and selected works across the six human context levels.} For the complete list of related methods, kindly refer to Sec.~\ref{sec::appearance_geometry}, Sec.~\ref{sec::dynamics_interaction}, and Sec.~\ref{sec::situated_embodied}.}
\label{fig::overview_tasks}
\end{figure}

\subsubsection{Spatial Geometry} 
\label{sec::level2}

\textbf{Conceptual Scope.}
Spatial geometry models humans through their articulated organization and geometric form. Compared with visual appearance, it moves beyond image-space characteristics to capture body properties that remain meaningful across viewpoints, including pose, shape, surface geometry, and spatial correspondence. This level provides an explicit account of how the human body occupies and is organized in 3D space, thereby connecting visual perception with physical human modeling.

\textbf{Representative Tasks.}
Representative tasks can be grouped into \textit{pose and mesh recovery} and \textit{renderable avatar modeling}. Pose and mesh recovery covers 2D and 3D pose estimation~\cite{li2025unipose}, parametric body recovery~\cite{fiche2026multi}, expressive whole-body reconstruction~\cite{wu2026pear}, multi-person recovery~\cite{yang2026sam}, and language-guided pose reasoning~\cite{feng2024chatpose}, generation, and editing~\cite{li2025unipose}. Renderable avatar modeling includes single-view human reconstruction~\cite{qiu2025lhm}, few-view human reconstruction~\cite{kirschstein2025avat3r}, animatable avatar creation~\cite{hu2026humannova}, head and facial modeling~\cite{kocasari2026face}, multi-view synthesis~\cite{kant2025pippo}, and novel-view rendering~\cite{yang2025sigman}. In the foundation-model era, scalable human priors, feed-forward reconstruction models, and multimodal semantic interfaces are expanding geometry modeling from specialized estimators toward reusable systems that integrate multiple functions.

\subsection{Dynamic Actors} 
\label{sec::dynamic_actors}
The perspective of dynamic actors extends human-centric intelligence from observable human states to behavior that unfolds over time and is shaped by external relations. We distinguish \textit{kinematic dynamics}, which focuses on the intrinsic temporal evolution of the human body, from \textit{interaction modeling}, which treats objects, scenes, and other people as integral components of human behavior. This distinction separates individual motion from behavior grounded in external relations.

\subsubsection{Kinematic Dynamics} 
\label{sec::level3}

\textbf{Conceptual Scope.}
Kinematic dynamics models humans as evolving subjects whose body states change and coordinate over time. Unlike spatial geometry, which describes body organization at a particular state, this level captures the continuity and variability of human movement. Human dynamics may be expressed through structured motion sequences, sensory measurements, or photorealistic videos. Its defining target is the temporal evolution of the human body rather than the external relations through which movement occurs.

\textbf{Representative Tasks.}
Representative tasks can be organized into \textit{motion understanding and generation} and \textit{human video animation}. Motion understanding and generation covers motion captioning~\cite{li2026llamo}, retrieval~\cite{wu2025mg}, reasoning~\cite{wang2026motiongpt}, prediction~\cite{bae2026ego}, text-to-motion synthesis~\cite{rempe2026kimodo}, motion editing~\cite{jiang2026motionmaster}, and streaming or multimodal generation~\cite{li2026llamo}. Human video animation realizes these dynamics in pixel space through audio-driven animation~\cite{jiang2025omnihuman,gan2025omniavatar,chen2025hunyuanvideo} and reference- and motion-guided animation~\cite{cheng2025wan}, while preserving identity and temporal coherence. In the foundation-model era, large-scale motion and video corpora, language-aligned motion interfaces, and pretrained generative backbones are connecting previously separate understanding, generation, editing, and animation tasks within increasingly general-purpose models.

\subsubsection{Interaction Modeling} 
\label{sec::level4}

\textbf{Conceptual Scope.}
Interaction modeling represents humans as relational participants whose behavior is jointly shaped by external entities. Unlike kinematic dynamics, it treats the relational counterpart as an essential part of the modeling problem. These relations may involve physical contact, object affordances, scene constraints, or communicative intent. Consequently, interaction modeling describes not only how humans behave, but also how that behavior remains compatible with objects, environments, and other people.

\textbf{Representative Tasks.}
Representative tasks are organized according to the primary relational counterpart, including \textit{human-object interaction}, \textit{human-scene interaction}, and \textit{social interaction}. Human-object interaction covers contact-aware motion generation~\cite{cai2026vihoi}, video generation~\cite{luo2026cointeract}, hand-object trajectory prediction~\cite{chen2025flowing}, interaction reconstruction~\cite{xu2026regenhoi}, and affordance-guided synthesis~\cite{li2025hoi}. Human-scene interaction includes scene-conditioned human placement~\cite{kister2026inhabit}, motion generation~\cite{liu2026open}, joint human-scene reconstruction~\cite{chen2025human3r}, and scene-motion-language reasoning~\cite{wang2025hsi}. Social interaction encompasses multi-person behavior understanding~\cite{bai2026humanomni}, human-human motion generation~\cite{wang2026social}, conversational response generation~\cite{luo2026omniresponse}, and interactive avatar animation~\cite{song2026interactiveavatar}. In the foundation-model era, large multimodal and generative priors are enabling more unified and open-ended interaction interfaces, while contact, affordance, scene compatibility, and social coordination remain essential sources of human-specific grounding.

\subsection{Situated Agents} 
\label{sec::situated_agents}
The perspective of situated agents connects human behavior with changing environments and executable capabilities. We distinguish \textit{world simulation}, which models how human actions and world states evolve together, from \textit{embodied agency}, which converts human-centered knowledge and experience into physically or operationally executable behavior. Together, these levels extend human-centric intelligence from describing humans and their relations to predicting consequences and enabling action.

\subsubsection{World Simulation} 
\label{sec::level5}

\textbf{Conceptual Scope.}
World simulation models humans as situated agents within evolving environments, focusing on how human actions, observations, and world states co-develop. It goes beyond interaction modeling by treating the environment not as a source of relational constraints, but as a dynamic system whose future states can be predicted, generated, or simulated around human activity. Egocentric observations provide an interface for this level because they connect human experience with changes in the surrounding world.

\textbf{Representative Tasks.}
Representative tasks comprise \textit{human-centered world generation} and \textit{actionable world modeling and planning}. Human-centered world generation synthesizes egocentric and human-conditioned futures under body motion~\cite{tu2025playerone}, hand interaction~\cite{wang2026hand2world}, viewpoint transformation~\cite{kang2026egox}, scene constraints~\cite{li2026anchorworld}, or goal specifications~\cite{shen2026egoforge}. Actionable world modeling and planning represent action-conditioned state transitions~\cite{chen2025egoagent}, persistent latent or geometric world states~\cite{hao2026egosim}, and action consequences~\cite{tran2026egoexo} supporting reasoning, trajectory evaluation, and planning. In the foundation-model era, pretrained video generators and predictive world models are moving this area beyond passive future synthesis toward scalable, action-conditioned simulators that connect human experience with consequence prediction and decision making.

\subsubsection{Embodied Agency} 
\label{sec::level6}

\textbf{Conceptual Scope.}
Embodied agency connects human-centric intelligence with executable behavior. It differs from world simulation because its primary objective is not to predict how an environment evolves, but to acquire physically or operationally executable capabilities. Human behavior and interaction experience provide priors for learning these capabilities. Embodied agency therefore forms the actionable end of the taxonomy, where human-centered knowledge supports control, adaptation, and skill transfer across embodiments.

\textbf{Representative Tasks.}
Representative tasks can be organized into \textit{humanoid control} and \textit{human-to-agent skill transfer}. Humanoid control covers physics-based character control~\cite{tessler2024maskedmimic}, whole-body motion tracking~\cite{qi2026humanoid}, language- or vision-guided interaction~\cite{deng2025human}, and generalist humanoid policies~\cite{bjorck2025gr00t} that convert semantic or motion specifications into executable behavior. Human-to-agent skill transfer includes learning from human demonstrations and egocentric videos~\cite{ma2026humanscale}, latent-action pretraining~\cite{chen2025moto}, behavior retargeting~\cite{li2025maniptrans}, embodiment alignment~\cite{zhou2025mitigating}, and policy adaptation for robots and other physical agents~\cite{kareer2025egomimic}. In the foundation-model era, large-scale human experience, pretrained multimodal models, and increasingly generalist control policies are creating new pathways for transferring human knowledge across tasks and embodiments, while physical executability and embodiment mismatch remain central challenges.
\section{Preliminary} 
\label{sec::preliminary}
Human-centric intelligence in the foundation-model era depends on three methodological foundations: heterogeneous human-centered data define what information can be learned, computational architectures determine how this information is transformed into capabilities, and optimization strategies govern how these capabilities are acquired, adapted, and elicited. These foundations apply not only to human-specialized foundation models, but also to methods that adapt general-purpose pretrained models or reflect broader foundation-era trends in scaling, transfer, multimodal integration, and general-purpose modeling.

\subsection{Human-Centric Data and Signal Families} \label{sec::data_families}
\begin{wrapfigure}{r}{0.6\textwidth}
    \begin{minipage}{\linewidth}
        \centering
        \includegraphics[width=\linewidth]{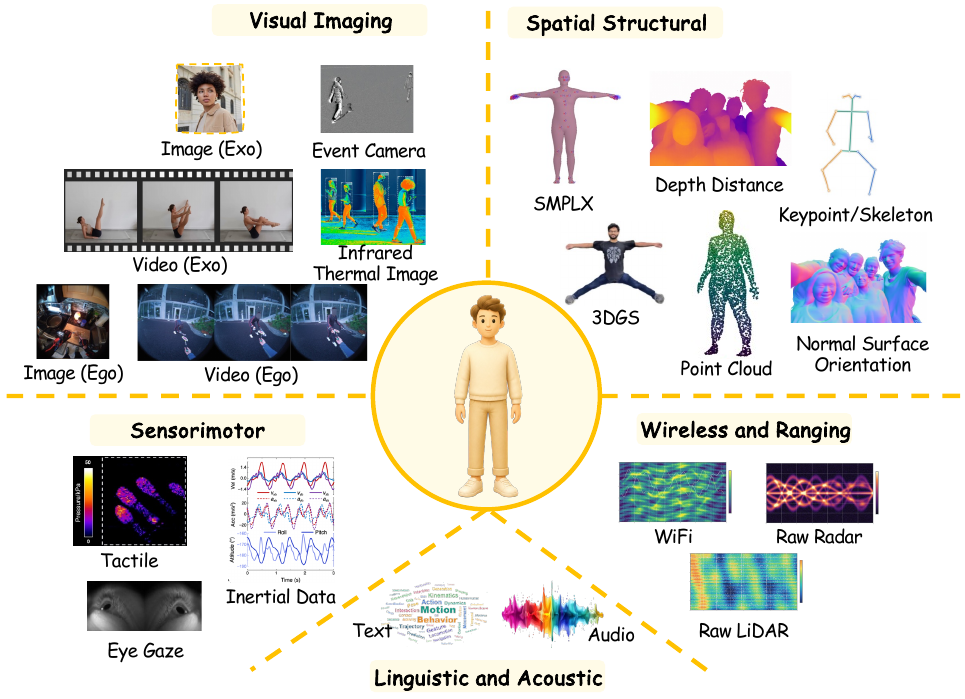}
        \caption{\textbf{Human-Centric Data and Signal Families.} The images are originally shown in~\cite{liu2020wearable,jdisplayTactileSensors,vecteezyDownloadColorful,rong2021frankmocap,joo2020eft,kocabas2024hugs,berlincioni2023upsampling,khirodkar2024sapiens,brady2025chum,chen2025motionllm,ma2024nymeria,visionifyInfraredThermal,braun2025egoppg}.}
        \vspace{-15pt}
        \label{fig::data_family}
    \end{minipage}
\end{wrapfigure}

Human-centric intelligence draws on heterogeneous data that characterize humans at different context levels and from different sensing perspectives. This section reviews the principal signals and structured representation formats used throughout the field. We organize them according to their acquisition mechanisms and the information they preserve into five families, as illustrated in Fig.~\ref{fig::data_family}.

\subsubsection{Visual Imaging Signals}
\label{sec::visual_imaging_signals}
Visual imaging signals provide image-plane observations of humans and their surrounding contexts. They primarily include exocentric RGB images (\HumanImage[-0.5ex][1.3em]) and videos (\HumanVideo[-0.5ex][1.3em])~\cite{khirodkar2024sapiens,li2025openhumanvid}, which observe humans from external viewpoints, and egocentric RGB images (\EgoImage[-0.5ex][1.3em]) and videos (\EgoVideo[-0.5ex][1.3em])~\cite{grauman2022ego4d,ma2024nymeria}, which record human experience from wearable or first-person cameras. Infrared and thermal images (\Infrared[-0.5ex][1.3em]) extend visual observation to low-light conditions~\cite{zhang2023diverse,jiang2025laboratory}, while event-camera signals (\Event[-0.5ex][1.3em]) asynchronously record intensity changes and preserve fast human motion at high temporal resolution~\cite{yan2024reli11d,wu2025cm3ae}. These signals are semantically rich, widely available, and readily scalable, making them the primary data source for developing human-centric capabilities. Nevertheless, their reliability can be affected by occlusion, viewpoint changes, illumination conditions, and motion blur, while the direct recording of human appearance and activity also introduces privacy concerns.

\subsubsection{Spatial-Structural Signals}
\label{sec::spatial_structural_signals}
Spatial-structural signals describe the geometric organization of humans rather than their radiometric appearance. Surface-level geometry can be represented by depth maps (\HumanDepth[-0.5ex][1.3em]) and normal maps (\HumanNormal[-0.5ex][1.3em])~\cite{khirodkar2024sapiens,saleh2025david,wang2026thfm}, 
whereas 2D poses and 3D skeletons (\HumanSkeleton[-0.5ex][1.3em]) provide sparse articulated body structures~\cite{lin2023motion,zhang2025motionxplusplus}. Denser representations include human point clouds (\HumanPointCloud[-0.5ex][1.3em])~\cite{shen2023lidargait,wang2026mmgait}, meshes and parametric body models (\HumanMesh[-0.5ex][1.3em])~\cite{cai2022humman,yang2026sam}, and neural renderable formats such as 3D Gaussian Splatting (\HumanGS[-0.5ex][1.3em])~\cite{qiu2025lhm,yang2025sigman}. These data can be captured using depth cameras and 3D scanners, reconstructed from multiple views, or estimated from visual observations. By making body structure and spatial configuration explicit, they support geometry-aware reconstruction, generation, motion modeling, and physical interaction. However, accurate geometric data remain costly to acquire, and estimated representations may inherit errors from sensors or reconstruction pipelines.

\subsubsection{Sensorimotor Signals}
\label{sec::sensorimotor_signals}
Sensorimotor signals characterize human sensing, bodily dynamics, physical feedback, and executable action. Gaze signals (\Gaze[-0.5ex][1.3em]) describe visual attention~\cite{wang2023holoassist,perrett2025hd}, while inertial measurements (\Acceleration[-0.5ex][1.3em]) record body movement through acceleration, angular velocity, and orientation~\cite{an2022mri,yan2024reli11d}. Action trajectories (\ActionTrajectory[-0.5ex][1.3em]) describe the spatial progression of behavior~\cite{yehia2026egotraj}, whereas control signals (\ActionControl[-0.5ex][1.3em]) specify physically executable commands~\cite{shi2026gpc}. Tactile signals (\HumanTactile[-0.5ex][1.3em]) provide direct evidence of physical contact~\cite{engelbracht2026hoi}, and physiological signals (\HumanPhysiology[-0.5ex][1.3em]) reveal internal bodily activity through measurements such as EEG, ECG, EMG, and PPG~\cite{xi2026egoemg,wang2026translating}. These signals often provide temporally precise and privacy-preserving information that is unavailable from visual observation alone. Their use at scale remains constrained by sensor placement, calibration, drift, individual variation, synchronization requirements, and limited data availability.

\subsubsection{Wireless and Ranging Signals}
\label{sec::wireless_and_ranging_signals}
Wireless and ranging signals sense humans by measuring how their presence and motion affect signal propagation or returned measurements. WiFi signals (\WiFi[-0.5ex][1.3em]) characterize human-induced variations in wireless communication channels~\cite{yang2023mm}, while radar signals (\Radar[-0.5ex][1.3em]) use reflected radio waves to estimate motion, range, and velocity~\cite{an2022mri,rahman2024mmvr,fan2026m4human}. Raw LiDAR measurements (\LiDAR[-0.5ex][1.3em]) instead rely on laser-based ranging to capture the spatial distribution of surrounding surfaces~\cite{yan2024reli11d,shen2023lidargait}. Although their physical sensing mechanisms differ, these signals provide non-contact observations that remain effective when conventional RGB imaging is unreliable or undesirable. Their limitations include restricted semantic detail, sensitivity to sensor configuration and environmental conditions, and difficulty transferring learned capabilities across sensing systems.

\subsubsection{Linguistic and Acoustic Signals}
\label{sec::linguistic_and_acoustic_signals}
Linguistic and acoustic signals provide semantic and communicative descriptions of human behavior. Text (\Text[-0.5ex][1.3em]) may take the form of captions, instructions, dialogues, or narrations~\cite{punnakkal2021babel,lin2023motion}, enabling human-centric representations to be aligned with concepts and user intent. Audio (\Audio[-0.5ex][1.3em]) includes speech as well as nonverbal acoustic information that reflects vocal expression, conversational timing, or rhythmic structure~\cite{liu2022beat,li2021ai,yi2023generating}. These signals serve as important interfaces for cross-modal understanding, controllable generation, and instruction-conditioned modeling, particularly when human behavior must be interpreted beyond directly observable body states. Nevertheless, they provide only indirect descriptions of human geometry and motion, and their informativeness depends on semantic specificity, recording quality, transcription accuracy, and temporal alignment with other modalities.

\subsection{Computational Architectures} 
\label{sec::computational_architectures}
\begin{figure}[t]
    \centering
    \includegraphics[width=\linewidth]{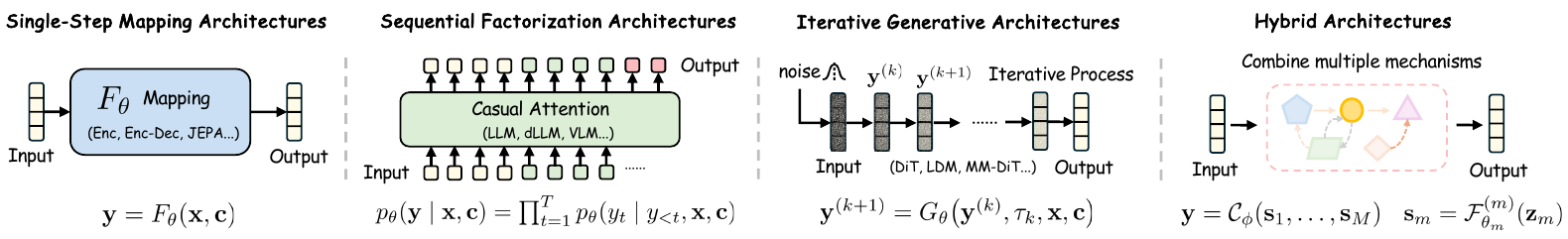}
    \caption{\textbf{Computational architecture paradigms.} These paradigms are shared across human contexts and provide the architectural basis for scaling and integrating human-centric capabilities in the foundation-model era.}
\label{fig::architecture}
\end{figure}

Human-centric models developed in the foundation-model era vary substantially in their internal designs, yet their core computations can be organized according to how encoded inputs are transformed into capability-specific outputs. As illustrated in Fig.~\ref{fig::architecture}, we distinguish them into four paradigms.

\subsubsection{Single-Step Mapping Architectures}
\label{sec::single_step_mapping_architectures}
Single-step mapping architectures (\SMA) produce target representations or outputs through a single feed-forward computation:
\begin{equation}
    \mathbf{y}=F_{\theta}(\mathbf{x},\mathbf{c}),
\end{equation}
where $\mathbf{x}$ denotes the input, $\mathbf{c}$ denotes optional conditioning information, $\mathbf{y}$ is the target representation or output, and $F_{\theta}$ is a parameterized feed-forward mapping. This paradigm includes encoder-only architectures that transform inputs into human representations~\cite{zuo2024plip,kim2025sapiensid}, as well as encoder-decoder architectures that map encoded features to structured predictions~\cite{khirodkar2024sapiens,yang2026sam,qiu2025lhm}. It also encompasses JEPA-style architectures, whose context encoder and predictor estimate target embeddings in representation space rather than reconstructing the input~\cite{chen2025egoagent}. By avoiding sequential or iterative decoding, single-step mapping architectures efficiently support perception, reconstruction, representation alignment, and feed-forward 3D modeling. Their effectiveness, however, depends on whether a single forward mapping can capture ambiguity in the target distribution.

\subsubsection{Sequential Factorization Architectures}
\label{sec::sequential_factorization_architectures}
Sequential factorization architectures (\SFA) represent outputs as ordered sequences and factorize their conditional distribution as
\begin{equation}
    p_{\theta}(\mathbf{y}\mid\mathbf{x},\mathbf{c})
    =
    \prod_{t=1}^{T}
    p_{\theta}(y_t\mid y_{<t},\mathbf{x},\mathbf{c}),
\end{equation}
where $\mathbf{y}=(y_1,\ldots,y_T)$ is an output sequence of length $T$, $y_{<t}$ denotes the preceding output units, and $p_{\theta}$ is the learned conditional distribution given the input $\mathbf{x}$ and optional condition $\mathbf{c}$. Autoregressive LLM and MLLM architectures apply this formulation to linguistic symbols or discretized human representations, allowing text~\cite{wang2026motiongpt}, motion~\cite{wang2025hsi}, and actions~\cite{shi2026gpc} to be modeled through next-token prediction. This formulation is particularly effective at capturing long-range dependencies and composing structured outputs of variable length across different human-centric signals. Nevertheless, sequential decoding introduces inference latency and may propagate early prediction errors to subsequent outputs.

\subsubsection{Iterative Generative Architectures}
\label{sec::iterative_generative_architectures}
Iterative generation architectures (\IGA) produce outputs by progressively transforming an initial state toward the target distribution:
\begin{equation}
    \mathbf{y}^{(0)}\sim p_{0},
    \qquad
    \mathbf{y}^{(k+1)}
    =
    G_{\theta}
    \bigl(\mathbf{y}^{(k)},\tau_k,\mathbf{x},\mathbf{c}\bigr),
    \quad k=0,\ldots,K-1,
    \qquad
    \mathbf{y}=\mathbf{y}^{(K)},
\end{equation}
where $\mathbf{y}^{(0)}$ is sampled from a source distribution $p_0$, $\mathbf{y}^{(k)}$ denotes the state at iteration $k$, $\tau_k$ specifies the corresponding generation stage, and $G_{\theta}$ is the learned iterative transformation. Diffusion models instantiate this process through reverse denoising, whereas flow-based models~\cite{wen2025hy} learn a continuous vector field that transports samples between source and target distributions. Both operate on the output as a globally evolving state, distinguishing them from the token-by-token dependencies of sequential architectures. Representative designs include latent diffusion models (LDMs)~\cite{li2024cosmicman,nam2025visual}, diffusion transformers (DiTs)~\cite{gan2025omniavatar,chen2025hunyuanvideo}, and multimodal diffusion transformers (MM-DiTs)~\cite{lee2025voost,chen2026tstars}. These architectures effectively model complex human-centric distributions, although their iterative sampling generally incurs greater inference cost than single-step mapping.

\subsubsection{Hybrid Architectures}
\label{sec::hybrid_architectures}
Hybrid architectures (\HA) combine multiple computational paradigms within a shared modeling process. Their computation can be abstracted as
\begin{equation}
    \mathbf{s}_m
    =
    \mathcal{F}^{(m)}_{\theta_m}(\mathbf{z}_m),
    \quad m=1,\ldots,M,
    \qquad
    \mathbf{y}
    =
    \mathcal{C}_{\phi}
    (\mathbf{s}_1,\ldots,\mathbf{s}_M),
    \quad M\geq2,
\end{equation}
where $\mathcal{F}^{(m)}_{\theta_m}$ denotes the $m$-th computational mechanism, $\mathbf{z}_m$ is its input or an intermediate state produced by another mechanism, and $\mathbf{s}_m$ is its resulting representation. The composition function $\mathcal{C}_{\phi}$ integrates these representations to produce the final output $\mathbf{y}$. These mechanisms may be composed sequentially, connected through shared representations, executed in parallel, or coupled through iterative feedback. LLM-diffusion architectures, for instance, use autoregressive reasoning to construct semantic conditions before diffusion-based realization~\cite{wang2026hsi,li2025unif}. VLM-guided control architectures combine direct perception or language reasoning with motion generation and physical control~\cite{ye2026visually,wu2025human}. World-action architectures may further couple latent-state prediction with action generation through shared or recurrent representations~\cite{luo2026being}. A model should be assigned to this category only when multiple paradigms are essential to its core computation, rather than merely because it contains conventional encoders, decoders, or task-specific modules.

\subsection{Optimization Strategies}  
\label{sec::optimization_strategies}
Optimization strategies determine how human-centric representation capabilities are acquired and invoked throughout the model lifecycle. In this survey, we mainly focus on training and inference stages.

\begin{tcolorbox}[
    colback=box_midyellow!5,
    colframe=box_midyellow!50,
    colbacktitle=box_midyellow!50,
    coltitle=black,
    title={\textbf{Optimization Strategies}},
    boxrule=5pt,
    arc=5pt,
    drop shadow,
    parbox=false,
    before skip=5pt,
    after skip=10pt,
    left=5pt,   
    right=20pt,
  ]
  \begin{itemize}
    \item \textbf{Training} (Section~\ref{sec::training}): updates all or selected trainable parameters before deployment, determining how human-centric capabilities are learned, adapted, transferred, or aligned.
    \item \textbf{Inference} (Section~\ref{sec::inference}): keeps learned parameters fixed and improves how existing capabilities are elicited through the test-time computation process.
\end{itemize}
\end{tcolorbox}

\subsubsection{Training Mechanisms} 
\label{sec::training}
Training mechanisms differ in model initialization, the parameters being updated, and the optimization signals used to guide capability acquisition. We summarize seven commonly used mechanisms: \textit{scratch-based training}, \textit{full-parameter fine-tuning}, \textit{parameter-efficient tuning}, \textit{instruction tuning}, \textit{reward-based tuning}, \textit{knowledge distillation}, and \textit{task-head tuning}. These mechanisms can be combined within a multi-stage training pipeline.

\textbf{\TrainScratch[0.2ex]~Scratch-based Training.}
It trains a model from randomly initialized parameters using human-centric data~\cite{khirodkarsapiens2,kim2025sapiensid,rempe2026kimodo}. This mechanism enables the architecture to form human-specialized capabilities without relying on pretrained weights, but it usually requires large-scale data and substantial computation.

\textbf{\TrainFull[0.2ex]~Full-Parameter Fine-tuning.}
It updates all parameters of a pretrained model using curated human-centric data~\cite{wang2026thfm,li2024cosmicman,chen2025foundhand}. This mechanism allows the pretrained feature or generation space to shift under human-specific supervision. It is effective when substantial domain adaptation is required, but incurs high computational cost and may weaken previously acquired general capabilities.

\textbf{\TrainParameter[0.2ex]~Parameter-Efficient Tuning.}
It adapts a pretrained model by updating only a selected subset of parameters or introducing lightweight learnable components. Representative mechanisms include low-rank adaptation (LoRA), adapters, prompt tuning, and prefix tuning~\cite{nam2025visual,li2025unipose,wang2026hand2world}. This strategy introduces human-specific capabilities while largely preserving the general knowledge of the original model, making it suitable when compute, memory, or training data are limited.

\textbf{\TrainInstruction[0.2ex]~Instruction Tuning.}
It optimizes a model using instruction-formatted human-centric data so that task descriptions, user intentions, and structured prompts can be mapped to appropriate outputs~\cite{zuo2025man,wang2025hsi,wang2026hsi}. Instruction tuning improves semantic controllability and cross-task usability by exposing multiple capabilities through a shared language-facing interface.

\textbf{\TrainReward[0.2ex]~Reward-based Tuning.}
It optimizes model behavior using reward signals that evaluate generated or executed outputs~\cite{ouyang2025motion,wang2026hsi,shi2026gpc}. These rewards can guide optimization either directly or through reinforcement-learning-style objectives. This strategy is useful when desired human-centric behavior cannot be specified adequately through direct supervised targets.

\textbf{\TrainDistillation[0.2ex]~Knowledge Distillation.}
It transfers capabilities from a teacher to a student by training the student to reproduce teacher-generated outputs, intermediate states, or behaviors~\cite{dai2024motionlcm,song2026interactiveavatar,tessler2024maskedmimic}. Beyond model compression, distillation can transfer capabilities across architectures, modalities, or stages of a perception-generation-control pipeline while reducing inference cost.

\textbf{\TrainTask[0.2ex]~Task-Head Tuning.}
It adapts a frozen pretrained backbone by training a lightweight task-specific prediction head~\cite{tang2023humanbench,ye2025silhouette}. The backbone provides fixed human-centric features, while the task head maps them to a particular output space. In addition to providing an efficient adaptation mechanism, task-head tuning offers a direct way to evaluate whether pretrained features transfer across human-centric tasks.

\subsubsection{Inference Optimization} 
\label{sec::inference}
Inference optimization methods improve how learned capabilities are elicited at test time while keeping model parameters fixed. We categorize them according to their intervention mechanisms, including \textit{semantic augmentation}, \textit{guided sampling}, \textit{iterative refinement}, \textit{preference-based selection}, and \textit{retrieval augmentation}. These mechanisms may also be combined within the same inference process.

\textbf{\InferenceSemantic[0.2ex]~Semantic Augmentation.}
It enriches the semantic inputs provided to a model. Text descriptions, task prompts, or class labels may be rewritten, expanded, or clarified by language models~\cite{song2026interactiveavatar}. This mechanism reduces ambiguity in human behavior descriptions and provides more informative conditions for generation and control.

\textbf{\InferenceGuided[0.2ex]~Guided Sampling.}
It improves inference by steering the sampling process with additional guidance signals. Typical mechanisms include classifier-free guidance, external guidance modules, or constraint-based guidance that biases generated samples toward desired conditions~\cite{chen2025foundhand,wu2026genlca,yao2026hosig}. This strategy is especially useful for improving controllability and consistency in human-centric generation.

\textbf{\InferenceIterative[0.2ex]~Iterative Refinement.}
It repeatedly corrects intermediate or final outputs using gradients, confidence scores, consistency checks, or task-specific constraints~\cite{bravo2026anny,jia2025primhoi}. This mechanism is valuable when a single forward pass cannot satisfy detailed body geometry, temporal coherence, interaction feasibility, or physical validity.

\textbf{\InferencePreference[0.2ex]~Preference-based Selection.}
It generates multiple candidates and selects the preferred result according to a reward model, learned evaluator, or trajectory comparison~\cite{kim2026dexterous,bai2026whole,wang2026lifting}. This mechanism enhances output quality by shifting optimization from single-sample generation to candidate-level selection.

\textbf{\InferenceRetrieve[0.2ex]~Retrieval Augmentation.}
It improves inference by retrieving relevant examples, contexts, or priors from an external memory or database. The retrieved information is then used to guide prediction, generation, or reasoning without changing model parameters~\cite{xu2026vimorag,li2026omni}. This mechanism enriches the model with instance-level human-centric knowledge that may be absent from the input alone.

\section{Human Visual Appearance and Spatial Geometry}
\label{sec::appearance_geometry}
Visual appearance and spatial geometry form the observable-subject perspective of human-centric intelligence. Visual appearance captures image-space cues for perception, identity, and controllable generation, whereas spatial geometry represents the body as an explicit or renderable structure. In the foundation-model era, both are shifting from task-specific models toward reusable human priors and flexible interfaces. Their goals remain distinct: appearance methods generalize across visual tasks and conditions, while geometry methods seek spatially consistent human states for reconstruction, animation, and rendering.

\subsection{Visual Appearance}
\label{sec::visual_appearance}
\begin{figure}[h]
    \centering
    \includegraphics[width=\linewidth]{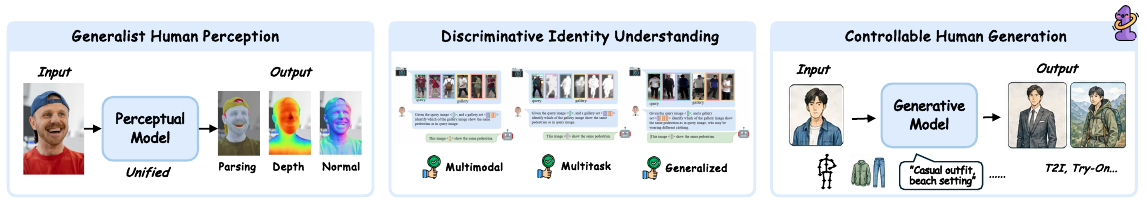}
    \caption{\textbf{Overview of visual appearance in human-centric intelligence.} We organize it into generalist human perception, discriminative identity understanding, and controllable human generation. The illustrations are adapted from~\cite{khirodkarsapiens2,niu2025chatreid}.}
\label{fig::visual_appearance}
\end{figure}

Visual appearance concerns how human-centric systems perceive, distinguish, and synthesize visible human properties. We organize this subsection by capability rather than architecture into generalist human perception, discriminative identity understanding, and controllable human generation. These categories respectively examine whether a model can support diverse human perception tasks, identify particular people under changing observations, and generate human appearance under structured user control, as illustrated in Fig.~\ref{fig::visual_appearance}.

\subsubsection{Generalist Human Perception}
\label{sec::appearance_type_I}
Generalist human perception learns reusable visual priors for multiple dense tasks instead of separate models for each annotation format. HumanBench establishes a common pretraining framework across human-centric tasks~\cite{tang2023humanbench}, while HAP integrates body-part structure into masked image modeling to improve transfer~\cite{yuan2023hap}. Sapiens scales ViTs on 300 million human images for pose, parsing, depth, and surface-normal estimation~\cite{khirodkar2024sapiens}. Sapiens2 extends this paradigm to one billion images and larger models, improving the scope and precision of dense predictions~\cite{khirodkarsapiens2}. These results suggest that scale alone is insufficient; human-specialized data and high-resolution modeling remain important for preserving body regions and surface details.

A complementary direction asks whether broad human perception requires extensive real-world annotations. DAViD shows that carefully generated synthetic humans can train accurate depth and surface-normal predictors with far less data~\cite{saleh2025david}. THFM extends dense perception to video by adapting a pretrained text-to-video diffusion model for temporally consistent human predictions~\cite{wang2026thfm}. Hulk unifies heterogeneous tasks as translation across visual, structural, and linguistic signals~\cite{wang2025hulk}, while L-Man exposes such capabilities through an instruction-following multimodal interface~\cite{zuo2025man}. This progression moves beyond shared encoders toward common interfaces for heterogeneous human information and tasks.

This generalist principle extends to facial perception. FaceXFormer unifies facial-analysis tasks with a shared encoder-decoder architecture~\cite{narayan2025facexformer}, while FSFM learns transferable self-supervised features for facial security~\cite{wang2025fsfm}. Face-MLLM enables instruction-based fine-grained facial understanding~\cite{sun2024face}, and GroundingFace links facial descriptions to pixel-level evidence~\cite{han2025groundingface}. UniF$^2$ace unifies facial understanding and generation in a discrete representation space~\cite{li2025unif}. Together, they specialize general-purpose modeling for the structural and semantic granularity of faces.

\begin{insightbox}[\faLightbulb~Trend: From Task Sharing to Human-Specialized Generality]
Generalist human perception is advancing through two complementary directions: human-specialized pretraining captures fine-grained knowledge underrepresented in generic corpora, while unified interfaces transfer it across perceptual tasks. Human-centric generality may therefore depend not on domain-neutral scaling alone, but on combining broad capabilities with carefully constructed human data. A central challenge is expanding task coverage without sacrificing performance.
\end{insightbox}

\subsubsection{Discriminative Identity Understanding}
\label{sec::appearance_type_II}
Discriminative identity understanding requires preserving person-level evidence across changing observations. In the foundation-model era, the first major shift concerns pretraining. SapiensID learns a unified visual identity model across face and body observations, using variable-resolution human regions to handle pose and scale variation~\cite{kim2025sapiensid}. SapiensID 2.0 extends this foundation by distilling stable semantic traits from an MLLM, disentangling transient appearance cues, and modeling kinematic continuity across frames~\cite{su2026sapiensid}. PLIP instead aligns person images with fine-grained attributes and identity-aware language supervision, producing features transferable across identity-related tasks~\cite{zuo2024plip}. Together, these methods provide complementary foundations: SapiensID unifies face and body recognition, SapiensID 2.0 introduces semantic and temporal identity cues, and PLIP connects identity-bearing appearance with reusable semantic knowledge.

Identity querying is also changing. Instruct-ReID unifies separate re-identification settings as an instruction-guided retrieval problem~\cite{he2024instruct}. Instruct-ReID++ extends this formulation into a more transferable instruction-conditioned model~\cite{he2025instruct}. ChatReID advances interactive retrieval by adapting a vision-language model from attribute understanding to fine-grained matching and multimodal reasoning~\cite{niu2025chatreid}. ReID5o generalizes queries, allowing combinations of RGB, infrared, text, and other visual modalities to retrieve people within one model~\cite{zuo2026reid5o}. Together, these methods shift re-identification from fixed image matching to query-conditioned search that selects relevant evidence.

Tasks answering ``which person'' do not share the same notion of identity. LVFace models persistent facial identity through staged optimization of large vision transformers~\cite{you2025lvface}. GIF enriches face supervision by replacing atomic identity labels with structured supervision~\cite{ebrahimi2025gif}. By contrast, RexSeek grounds people from natural-language descriptions involving attributes, positions, and interactions~\cite{jiang2025referring}, while MMPedestron detects and localizes pedestrians across heterogeneous sensors without establishing persistent identity~\cite{zhang2024pedestrian}. These systems broaden human search and grounding, but their semantic targets differ from biometric identity. This distinction clarifies what foundation-era generalization preserves.

\begin{insightbox}[\faLightbulb~Open Problem: Toward Transferable Yet Persistent Identity]
Foundation-era pretraining is shifting identity modeling from dataset-specific matching toward recognition across heterogeneous observations and queries. However, transferability does not ensure stable identity, as pretrained models may depend on appearance cues that change over time. This motivates separating persistent identity from incidental visual semantics while retaining model adaptability and accounting for privacy, demographic variation, and uncertainty.
\end{insightbox}

\subsubsection{Controllable Human Generation}
\label{sec::appearance_type_III}

It builds foundation generators to synthesize or edit people while preserving identity and anatomical structure. Generic text-to-image models provide broad visual priors but are not supervised for the compositional structure of humans. CosmicMan addresses this gap with large-scale human data, body-aware annotations, and structured objectives that align textual attributes with body regions~\cite{li2024cosmicman}. At a finer scale, FoundHand synthesizes hands through domain-specific pretraining and explicit structural control, compensating for the weak modeling of articulation and occlusion in generic image priors~\cite{chen2025foundhand}. Together, these methods show that generative scale requires human-specific data and structure to support reliable control.

The next development shifts from global prompts to factorized and compositional control. Visual Persona decomposes reference identity across body regions, preserving appearance as pose and scene vary~\cite{nam2025visual}. Voost learns person-garment correspondence in a shared diffusion transformer for virtual try-on and inverse try-off~\cite{lee2025voost}. Tstars-Tryon 1.0 extends this setting to diverse fashion items through larger-scale data construction, semantic conditioning, and preference alignment~\cite{chen2026tstars}. Oxygen-TryOn further develops a fashion-native foundation model through a dedicated data engine and staged pretraining, fine-tuning, and reinforcement learning, supporting multi-reference composition across wearable categories while preserving subject and item consistency~\cite{liu2026oxygen}. Together, these methods move virtual try-on from garment-specific transfer toward compositional control over human appearance and referenced items.

A third direction unifies previously separate generation pipelines. UPGPT combines human generation, appearance editing, and pose transfer in a multimodal diffusion model~\cite{cheong2023upgpt}. DreamActor-M1 extends identity and appearance control to coherent human animation through reference and motion guidance~\cite{luo2025dreamactor}. Archon further tokenizes synchronized multimodal signals in one autoregressive model for holistic digital-human generation~\cite{bao2026archon}. These systems reveal two forms of foundation-era unification: a shared generator for related appearance operations and a shared model across human modalities and outputs.

\begin{insightbox}[\faLightbulb~Trend: Toward Compositional Control in Human Generation]
The foundation-model era has strengthened generative priors for human synthesis, yet controllability remains harder than visual realism. Human-centered generation requires models to edit human properties while preserving unrelated characteristics, creating a tension between flexibility and human-specific invariance. Recent progress points toward compositional control, where multiple intentions can be combined without interference. Achieving this compositionality with reusable foundation architectures remains an important direction.
\end{insightbox}

\begin{table}[!t]
\rowcolors{2}{white}{table_creamyellow}
\centering
\caption{
Representative methods for visual appearance and spatial geometry in human-centric intelligence.
\\[0.3ex]
\textbf{$\bullet$ Modalities:} 
\HumanImage~Exocentric Image,
\HumanVideo~Exocentric Video,
\Infrared~Infrared \& Thermal Image,
\HumanDepth~Depth Map,
\HumanNormal~Normal Map,
\HumanSkeleton~Pose \& Skeleton,
\HumanMesh~Mesh \& Parametric Body Model,
\HumanGS~3D Gaussian Splatting,
\Text~Text
and \Audio~Audio.\\
\textbf{$\bullet$ Training:} 
\TrainScratch~Scratch-based Training,
\TrainFull~Full-Parameter Fine-tuning,
\TrainParameter~Parameter-Efficient Tuning,
\TrainInstruction~Instruction Tuning,
\TrainReward~Reward-based Tuning,
\TrainDistillation~Knowledge Distillation,
and \TrainTask~Task-Head Tuning.\\
\textbf{$\bullet$ Inference:} 
\InferenceSemantic~Semantic Augmentation,
\InferenceGuided~Guided Sampling,
and \InferenceIterative~Iterative Refinement.\\
\textbf{$\bullet$ Architectures:} 
\SMA~Single-Step Mapping,
\SFA~Sequential Factorization,
\IGA~Iterative Generative,
and \HA~Hybrid. \\
\textbf{$\bullet$ Context:} 
\VisualAppearance~Visual Appearance,
and \StructuralGeometry~Spatial Geometry.\\
\textbf{$\bullet$ Markers:} 
\Real~Real-World,
\Synthetic~Synthesized,
\Estimated~Estimated,
and \NA~Not Mentioned.
}
\vspace{-7pt}
\resizebox{\linewidth}{!}{
\begin{tabular}{l|c|c|c|c|c|c|c|c|c|c|c}
    \toprule
    \textbf{Method} & \textbf{Venue} & \textbf{Modality} & \textbf{Data Scale} & \textbf{Model Size} & \textbf{Train.} & \textbf{Inf.}& \textbf{Arc.}  & \textbf{Context} & \textbf{Function} & \textbf{Paper} & \textbf{Web} \\
    \midrule\midrule

    \rowcolor{gray!10} \multicolumn{12}{l}{\textit{\textbf{Sec.~\ref{sec::appearance_type_I}: Generalist Human Perception}}}\\
    \midrule

    Sapiens2~\cite{khirodkarsapiens2} & ICLR'26 & \HumanImage[-0.5ex][0.03\linewidth]~\HumanSkeleton[-0.5ex][0.028\linewidth]~\HumanDepth[-0.5ex][0.028\linewidth]~\HumanNormal[-0.5ex][0.028\linewidth] & 1B Images\textsuperscript{\Real} & 0.4B-5B & \TrainScratch[0.2ex]~\TrainDistillation[0.2ex]~\TrainTask[0.2ex] & \NA & \SMA[0.2ex] & \VisualAppearance[-0.5ex][0.028\linewidth] & Dense Perception & \href{https://openreview.net/forum?id=IVAlYCqdvW}{\faBook} & \href{https://github.com/facebookresearch/sapiens2}{\faGithub} \\

    Sapiens~\cite{khirodkar2024sapiens} & ECCV'24 & \HumanImage[-0.5ex][0.03\linewidth]~\HumanSkeleton[-0.5ex][0.028\linewidth]~\HumanDepth[-0.5ex][0.028\linewidth]~\HumanNormal[-0.5ex][0.028\linewidth] & 300M Images\textsuperscript{\Real} & 0.3B-2B & \TrainScratch[0.2ex]~\TrainTask[0.2ex] & \NA & \SMA[0.2ex] & \VisualAppearance[-0.5ex][0.028\linewidth] & Dense Perception & \href{https://link.springer.com/chapter/10.1007/978-3-031-73235-5_12}{\faBook} & \href{https://github.com/facebookresearch/sapiens}{\faGithub} \\
\bottomrule
\end{tabular}}
\label{tab::summary_appearance_geometry}
\end{table}

\subsection{Spatial Geometry}
\label{sec::structural_geometry}
\begin{wrapfigure}{r}{0.7\textwidth}
    \begin{minipage}{\linewidth}
        \centering
        \vspace{-5pt}
        \includegraphics[width=\linewidth]{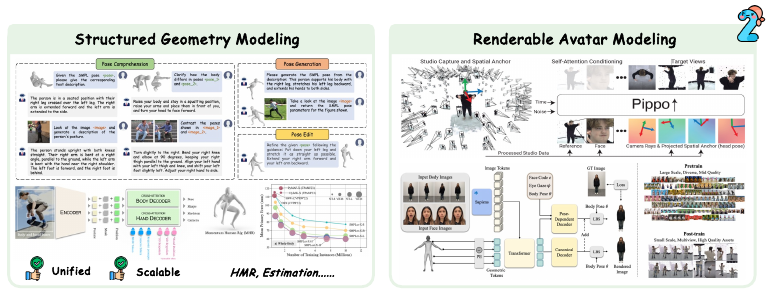}
        \caption{\textbf{Overview of spatial geometry in human-centric intelligence.} We organize spatial geometry into structured geometry modeling and renderable avatar modeling. The images are originally shown in~\cite{feng2024chatpose,yang2026sam,yin2025smplest,kant2025pippo,li2026large}.}
        \label{fig::spatial_geometry}
        \vspace{-5pt}
    \end{minipage}
\end{wrapfigure}

Spatial geometry models the human body structure beyond image-space appearance. We organize it into structured geometry modeling, which estimates explicit body geometry, and renderable avatar modeling, which integrates geometry and appearance into renderable assets. In the foundation-model era, both are shifting from task-specific estimators toward scalable human priors and transferable models with flexible semantic interfaces, as shown in Fig.~\ref{fig::spatial_geometry}.

\subsubsection{Structured Geometry Modeling}
\label{sec::geometry_type_I}

Structured geometry modeling converts visual observations into explicit, view-aware body geometry. One foundation-era trajectory treats expressive mesh recovery as a scaling problem. SMPLer-X studies how training data, model capacity, and dataset composition affect transfer across body estimation settings~\cite{cai2023smpler}. SMPLest-X expands both training corpus and model scale to improve whole-body recovery~\cite{yin2025smplest}. HaMeR applies large-scale transformer training to hand reconstruction, showing the continued value of specialized models for fine articulation~\cite{pavlakos2024reconstructing}. PEAR emphasizes pixel alignment for efficient whole-body recovery~\cite{wu2026pear}, while DanceHMR addresses temporal instability through body-hand coordination in monocular video~\cite{shen2026dancehmr}. Together, these methods show that scaling benefits human geometry when paired with structure at suitable spatial and temporal granularity.

A second trajectory extends recovery from cropped individuals toward scene-level and temporally persistent geometry. Multi-HMR 2 jointly estimates multiple people and camera parameters within a consistent spatial frame~\cite{fiche2026multi}. Building on pretrained visual features, DETRAM combines detection queries with persistent tracking and box-initialized prompt queries to unify full-frame multi-person detection, SMPL recovery, and identity-consistent tracking in a single transformer decoder~\cite{lee2026detram}. PromptHMR makes mesh recovery promptable, using spatial or semantic cues to identify targets and resolve visual ambiguity~\cite{wang2025prompthmr}. SAM 3D Body scales full-body recovery with a large model and VLM-assisted data engine for in-the-wild cases~\cite{yang2026sam}. Together, these methods reflect a foundation-era shift from fixed detection-and-cropping pipelines toward unified geometry models that preserve scene context, temporal identity, and user-specified targets.

A third trajectory connects human geometry with higher-level intelligence. PoseEmbroider aligns images, 3D poses, and language in a shared space for pose retrieval and reasoning~\cite{delmas2024poseembroider}. ChatPose links multimodal language models with SMPL parameters for natural-language interaction with 3D poses~\cite{feng2024chatpose}. UniPose unifies pose understanding and synthesis within a pose-aware language model~\cite{li2025unipose}. This principle also extends to facial geometry: EmoteGPT maps language to controllable expressions~\cite{wang2026emotegpt}, while ContextFace infers expressions compatible with emotional contexts~\cite{kim2025contextface}. These methods expose human geometry to multimodal foundation models, although semantic capability does not guarantee spatial precision.

Foundation models can support geometry without acting as primary predictors. VLM-GPA uses vision-language preferences to guide diffusion-based mesh recovery toward perceptually plausible results~\cite{shen2026vlm}. Anny-Fit combines specialized geometric cues with VLM reasoning to improve body-shape recovery across age groups and multi-person scenes~\cite{bravo2026anny}. Here, foundation models provide knowledge or feedback while geometric components perform spatial reconstruction. Their influence may therefore arise through multiple methodological roles rather than a single unified architecture.

\begin{insightbox}[\faLightbulb~Open Problem: Reconciling Semantic Priors with Metric Geometry]
Foundation models are shifting human mesh recovery from fixed regression toward transferable, interactive geometry modeling. Yet semantic or perceptual plausibility does not ensure metric accuracy, spatial consistency, or physically valid structure. Foundation priors may therefore be most effective when grounded in explicit geometry rather than replacing it. Future work may connect semantic abstraction more closely with verifiable human geometry while representing uncertainty under incomplete observations.
\end{insightbox}

\subsubsection{Renderable Avatar Modeling}
\label{sec::geometry_type_II}
Renderable avatar modeling produces 3D human assets that support rendering and animation. A major foundation-era shift replaces per-subject optimization with amortized feed-forward reconstruction. IDOL
reconstructs a photorealistic 3D human from a single image without prolonged subject-specific fitting~\cite{zhuang2025idol}. LHM scales this approach with a large transformer that directly predicts animatable Gaussian avatars~\cite{qiu2025lhm}. HumanNOVA combines image evidence with a simplified body prior and large human-asset collections for rapid reconstruction across diverse inputs~\cite{hu2026humannova}. These models shift computation from inference-time optimization to pretraining, making avatar creation a reusable capability.

Single-image reconstruction remains underconstrained because only part of a person's appearance and geometry is visible. Portrait 3D Presence handles varying portrait coverage through a shared canonical formulation~\cite{zhang2026bringing}. AniGS uses generated canonical views and reconciles their inconsistencies during Gaussian avatar construction~\cite{qiu2025anigs}. Despite different mechanisms, both use learned priors to infer unobserved content while preserving input evidence. This highlights a central role of foundation-era avatar models: population-level knowledge compensates for incomplete subject observations.

A more direct foundation-model direction expands the human data used to learn avatar priors. Large-scale Codec Avatars first captures broad variation from in-the-wild videos, then improves fidelity through post-training on curated data~\cite{li2026large}. SIGMAN builds a million-scale collection of Gaussian human assets for native 3D generative modeling~\cite{yang2025sigman}. GenLCA uses a pretrained avatar reconstructor as a 3D tokenizer, enabling generative learning from partially observed videos~\cite{wu2026genlca}. DreamCharacter-1 adapts pretrained 3D backbones for animation-ready character creation through targeted post-training and inference optimization~\cite{liu2026dreamcharacter}. Together, these methods bring broad pretraining and capability-oriented adaptation to avatar modeling.

Population-level learning is also advancing in specialized avatar domains. FiCA learns a feed-forward Gaussian prior for detailed head reconstruction from one portrait~\cite{youwang2026fica}. HeadsUp scales high-quality head reconstruction across a larger identity set~\cite{ntavelis2026large}. Avat3r adapts large reconstruction models to create animatable head avatars from few observations~\cite{kirschstein2025avat3r}. Face Anything handles facial reconstruction from unconstrained image sequences with temporal variation~\cite{kocasari2026face}. LUCAS separates head and hair into layered representations that preserve detail during animation~\cite{liu2025lucas}. These specialized tracks show that broad avatar pretraining still benefits from representations tailored to fine geometry and deformation.

Renderable avatar modeling is expanding from reconstruction to controllable generation. Pippo uses large-scale generative learning to synthesize high-resolution, view-consistent human views from a single image~\cite{kant2025pippo}. MV-Performer adapts pretrained video diffusion for synchronized multi-view performer synthesis~\cite{zhi2025mv}. InfiniHuman combines large generative priors with structured human conditioning for controllable 3D creation~\cite{xue2025infinihuman}. These methods connect avatar modeling with broader generative foundation models, but their outputs still require coherent geometry and animation beyond visual consistency.

\begin{insightbox}[\faLightbulb~Trend: From Per-Subject Fitting to Population-Level Priors]
The foundation-model shift in avatar modeling lies less in a 3D representation than in moving subject-specific computation into population-scale pretraining. This amortization enables faster creation and broader generalization, but may conceal uncertainty in unobserved body structure or yield plausible avatars with less reliable geometry and animation. Future progress may therefore depend on balancing population-level learning with coherent, controllable, and animation-ready human models.
\end{insightbox}
\section{Human Kinematic Dynamics and Interaction Modeling} 
\label{sec::dynamics_interaction}
Kinematic dynamics and interaction modeling form the dynamic-actor perspective of human-centric intelligence. Kinematic dynamics focuses on the temporal evolution of the human body, whereas interaction modeling examines how that evolution is shaped by relations with objects, environments, and other people. In the foundation-model era, these areas are moving beyond isolated motion generators and interaction-specific pipelines toward reusable temporal priors, multimodal interfaces, and more general relational models.

\subsection{Kinematic Dynamics}
\label{sec::kinematic_dynamics}
\begin{wrapfigure}{r}{0.7\textwidth}
    \begin{minipage}{\linewidth}
        \centering
        \vspace{-5pt}
        \includegraphics[width=\linewidth]{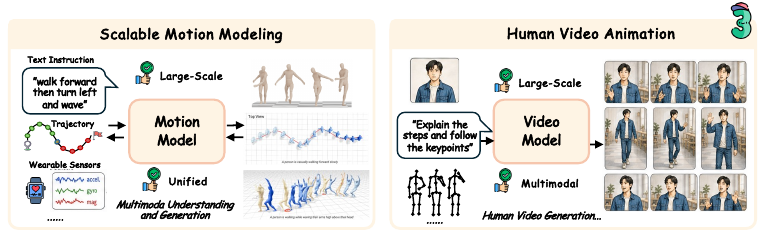}
        \caption{\textbf{Overview of kinematic dynamics in human-centric intelligence.} We organize kinematic dynamics into scalable motion modeling and human video animation. The images are originally shown in~\cite{rempe2026kimodo,wen2025hy}.}
        \label{fig::kinematic_dynamics}
        \vspace{-5pt}
    \end{minipage}
\end{wrapfigure}

Kinematic dynamics concerns how human behavior evolves over time. We organize this subsection into scalable motion modeling, which models body dynamics in structured or learned temporal spaces, and human video animation, which realizes those dynamics in pixel space while preserving human appearance and temporal coherence, as illustrated in Fig.~\ref{fig::kinematic_dynamics}.

\begin{table}[!ht]
\rowcolors{2}{white}{table_creamyellow}
\centering
\caption{
Representative methods for kinematic dynamics in human-centric intelligence.
\\[0.3ex]
\textbf{$\bullet$ Modalities:} 
\HumanImage~Exocentric Image,
\HumanVideo~Exocentric Video,
\EgoVideo~Egocentric Video,
\HumanSkeleton~Pose \& Skeleton,
\HumanMesh~Mesh \& Parametric Body Model,
\Acceleration~Inertial Measurement Signal,
\ActionTrajectory~Action Trajectory,
\Text~Text
and \Audio~Audio.\\
\textbf{$\bullet$ Training:} 
\TrainScratch~Scratch-based Training,
\TrainFull~Full-Parameter Fine-tuning,
\TrainParameter~Parameter-Efficient Tuning,
\TrainInstruction~Instruction Tuning,
\TrainReward~Reward-based Tuning,
\TrainDistillation~Knowledge Distillation,
and \TrainTask~Task-Head Tuning.\\
\textbf{$\bullet$ Inference:} 
\InferenceSemantic~Semantic Augmentation,
\InferenceGuided~Guided Sampling,
\InferenceIterative~Iterative Refinement,
\InferencePreference~Preference-based Selection,
and \InferenceRetrieve~Retrieval Augmentation.\\
\textbf{$\bullet$ Architectures:} 
\SMA~Single-Step Mapping,
\SFA~Sequential Factorization,
\IGA~Iterative Generative,
and \HA~Hybrid. \\
\textbf{$\bullet$ Context:} 
\KinematicDynamics~Kinematic Dynamics.\\
\textbf{$\bullet$ Markers:} 
\Real~Real-World,
\Synthetic~Synthesized,
\Estimated~Estimated,
and \NA~Not Mentioned.
}
\vspace{-7pt}
\resizebox{\linewidth}{!}{
\begin{tabular}{l|c|c|c|c|c|c|c|c|c|c|c}
    \toprule
    \textbf{Method} & \textbf{Venue} & \textbf{Modality} & \textbf{Data Scale} & \textbf{Model Size} & \textbf{Pre.} & \textbf{Inf.}& \textbf{Arc.}  & \textbf{Context} & \textbf{Function} & \textbf{Paper} & \textbf{Web} \\
    \midrule\midrule

    \rowcolor{gray!10} \multicolumn{12}{l}{\textit{\textbf{Sec.~\ref{sec::dynamics_type_I}: Scalable Motion Modeling}}}\\
    \midrule
    Kimodo~\cite{rempe2026kimodo} & arXiv'26 & \HumanMesh[-0.5ex][0.028\linewidth]~\Text[-0.5ex][0.028\linewidth] & 700 Hours\textsuperscript{\Real} & 56M-282M & \TrainScratch[0.2ex]~\TrainFull[0.2ex] & \InferenceGuided[0.2ex]~\InferenceIterative[0.2ex] & \IGA[0.2ex] & \KinematicDynamics[-0.5ex][0.028\linewidth] & Scalable Motion Generation & \href{https://arxiv.org/abs/2603.15546}{\faBook} & \href{https://github.com/nv-tlabs/kimodo}{\faGithub} \\

    ARDY~\cite{zhao2026autoregressive} & TOG'26 & \HumanMesh[-0.5ex][0.03\linewidth]~\Text[-0.5ex][0.028\linewidth]~\ActionTrajectory[-0.5ex][0.028\linewidth] & 700 Hours\textsuperscript{\Real} & 156M & \TrainScratch[0.2ex] & \InferenceGuided[0.2ex]~\InferenceIterative[0.2ex] & \IGA[0.2ex] & \KinematicDynamics[-0.5ex][0.028\linewidth] & Scalable Motion Generation & \href{https://arxiv.org/abs/2607.08741}{\faBook} & \href{https://github.com/nv-tlabs/ardy}{\faGithub} \\
\bottomrule
\end{tabular}}
\label{tab::summary_dynamics}
\end{table}

\subsubsection{Scalable Motion Modeling}
\label{sec::dynamics_type_I}
Scalable motion modeling treats human movement as a temporal signal rather than independent poses. Early foundation-oriented methods built reusable temporal and linguistic interfaces. MotionBERT learns transferable features through masked sequence pretraining for multiple motion-analysis tasks~\cite{zhu2023motionbert}. MotionGPT encodes motion as discrete tokens, unifying motion captioning and generation~\cite{jiang2023motiongpt}. AvatarGPT extends this interface to planning and task decomposition~\cite{zhou2024avatargpt}, while M$^3$GPT adds acoustic and multimodal conditioning~\cite{luo2024m}. Together, they establish motion as a transferable modality rather than a task-specific output.

A subsequent direction examines scaling in motion-native models. Being-M0 studies how T2M generation responds to data and model scaling~\cite{wang2024scaling}. ScaMo analyzes scaling across motion-generator families~\cite{lu2025scamo}, while Go-to-Zero expands pretraining with a much larger motion corpus~\cite{fan2025go}. Kimodo scales a controllable diffusion prior to hundreds of motion-capture hours~\cite{rempe2026kimodo}. HY-Motion further increases training scale and introduces alignment optimization for text-conditioned generation~\cite{wen2025hy}. ARDY adapts large motion priors for online generation through causal denoising and interactive constraints~\cite{zhao2026autoregressive}. Together, these studies suggest that motion scaling depends on the quality and controllability of temporal supervision, not data volume alone.

The foundation-model era also promotes unification across motion capabilities. MG-MotionLLM connects motion understanding and generation through a shared motion-language interface~\cite{wu2025mg}. Being-M0.5 extends this interface with visual observations~\cite{cao2025being}. LLaMo adapts pretrained language models with motion-specific transformers and continuous autoregressive tokens for motion understanding and real-time generation~\cite{li2026llamo}. GENMO instead treats motion estimation as constrained generation, sharing a generative prior between 
estimation and synthesis~\cite{li2025genmo}. Superman aligns skeleton sequences with visual and linguistic inputs for joint motion perception and generation~\cite{wang2026superman}. These models explore whether understanding and generation can reinforce each other through shared temporal knowledge.

Greater generality also reshapes motion control and refinement. KinMo adds explicit kinematic structure to motion-language modeling~\cite{zhang2025kinmo}, while HuMoCon discovers reusable concepts from video and motion data~\cite{fang2025humocon}. MotionAgent uses a language model to coordinate generation tools through interactive instructions~\cite{wu2025motion}, and Motion-R1 introduces reasoning alignment for complex text-to-motion generation~\cite{ouyang2025motion}. VimoRAG grounds parametric generation in retrieved human videos~\cite{xu2026vimorag}. MotionStreamer enables causal, incremental generation~\cite{xiao2025motionstreamer}, while MotionMaster~\cite{jiang2026motionmaster} and MotionLab~\cite{guo2025motionlab} unify generation and editing. Together, these approaches extend pretrained motion priors with external guidance and refinement for greater control.

Motion intelligence is also expanding beyond conventional motion-capture inputs. EgoLM aligns egocentric video, body motion, and language for movement modeling from the actor's viewpoint~\cite{hong2025egolm}. Ego4o combines egocentric video with inertial and body signals for motion capture and understanding~\cite{wang2025ego4o}. AnyLift trains camera-conditioned 2D motion diffusion on specialized synthetic data to recover motion from unconstrained Internet videos~\cite{li2026anylift}. FoundationGait explores transferable gait pretraining~\cite{ye2025silhouette}, while GaitDynamics models biomechanical movement variation generatively~\cite{tan2026gaitdynamics}. Sensor-language models further extend temporal modeling to nonvisual signals. Despite differing observations, these methods target the intrinsic evolution of human movement rather than external relations.

\begin{insightbox}[\faLightbulb~Trend: From Motion Tokenization to Transferable Temporal Intelligence]
Foundation-era motion modeling is moving from language-compatible encodings toward temporal knowledge transferable across motion tasks. The central question is no longer whether motion can be tokenized, but whether shared models can preserve continuous kinematics across objectives and observation domains. Larger corpora may improve generalization, yet current scaling strategies capture physical validity, temporal causality, and subtle human variation only partially.
\end{insightbox}

\subsubsection{Human Video Animation}
\label{sec::dynamics_type_II}
Human video animation realizes body dynamics in photorealistic video. Audio-driven systems directly benefit from pretrained video generators. OmniAvatar adapts a large video model for audio-conditioned portrait and body animation through parameter-efficient tuning~\cite{gan2025omniavatar}. HunyuanVideo-Avatar builds on a text-to-video backbone to improve identity consistency and expressive behavior~\cite{chen2025hunyuanvideo}. InfinityHuman scales data and alignment for longer, more diverse animation~\cite{li2026infinityhuman}. EchoMimicV3 unifies multiple audio-driven settings within one model~\cite{meng2026echomimicv3}. OmniHuman-1.5 adds high-level semantic guidance from a multimodal language model, moving beyond direct audio-motion correspondence~\cite{jiang2025omnihuman}. Together, these methods specialize the visual and temporal priors of video foundation models for human behavior.

Motion-guided animation separates source appearance from driving dynamics. Wan-Animate adapts pretrained video generation to pose-guided character animation and replacement~\cite{cheng2025wan}. HumanDiT uses a human-specific DiT to transfer body motion while preserving appearance~\cite{gan2025humandit}. DreamActor-M1 combines motion and appearance guidance for animation~\cite{luo2025dreamactor}. MTVCraft tokenizes 4D human motion for controllable video generation~\cite{ding2026mtvcrafttokenizing4dmotion}. These methods reveal a challenge: stronger generative priors improve visual quality, but motion transfer still requires precise correspondence between the driving body and the generated person.

Recent models increasingly unify motion with video generation. UniMo jointly models human video and 3D motion autoregressively~\cite{pang2025unimo}. CoMoVi co-generates realistic video and structured body motion~\cite{zhao2026comovi}, while EchoMotion uses motion cues to improve video temporal structure~\cite{yang2025echomotion}. Archon encodes multiple digital-human modalities within one autoregressive vocabulary~\cite{bao2026archon}. HuMo~\cite{chen2026human} and DreamID-Omni~\cite{guo2026dreamid} broaden multimodal conditioning and animation within unified systems. EgoControl extends motion-conditioned generation to first-person video~\cite{pallotta2026egocontrol}. Together, these methods move structured motion and video foundation models toward convergence, although they retain distinct forms of temporal supervision.

\begin{insightbox}[\faLightbulb~Open Problem: Temporal Fidelity Beyond Video-Model Scale]
Large video backbones have improved human animation realism and flexibility, yet visual scale does not ensure reliable dynamics. Small temporal errors can cause identity drift, anatomical inconsistency, or implausible motion despite convincing individual frames. Human video foundation models may therefore benefit from explicit motion knowledge alongside visual generation objectives. Long-horizon coherence and sensitivity to subtle timing remain key indicators of temporal intelligence.
\end{insightbox}

\subsection{Interaction Modeling}
\label{sec::interaction_modeling}
\begin{wrapfigure}{r}{0.7\textwidth}
    \begin{minipage}{\linewidth}
        \centering
        \vspace{-5pt}
        \includegraphics[width=\linewidth]{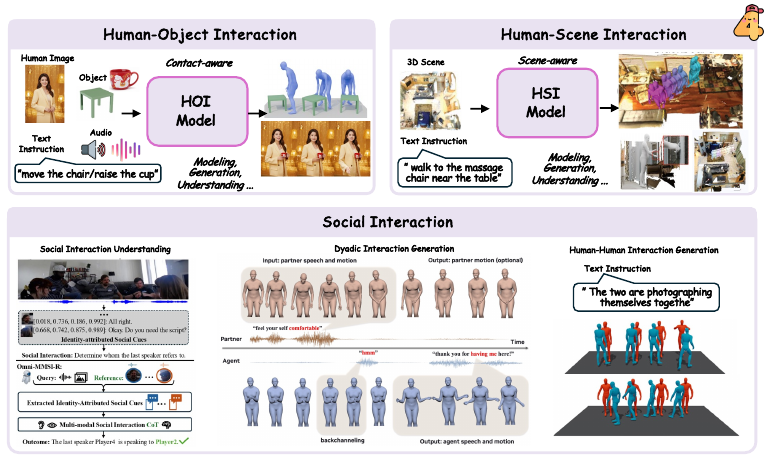}
        \caption{\textbf{Overview of interaction modeling in human-centric intelligence.} We organize interaction modeling into human-object, human-scene, and social interaction. The images are originally shown in~\cite{luo2026cointeract,cai2026vihoi,liu2026open,wang2025hsi,wang2026hsi,bai2026humanomni,nagano2026dyaplex,wang2026social}.}
        \label{fig::interaction_modeling}
        \vspace{-5pt}
    \end{minipage}
\end{wrapfigure}

Interaction modeling treats humans as relational actors whose behavior is jointly shaped by external entities, as shown in Fig.~\ref{fig::interaction_modeling}. We organize this subsection into human-object interaction, human-scene interaction, and social interaction, reflecting an expansion from local physical relations to environmental compatibility and interpersonal coordination. Foundation models broaden these settings through open-ended semantic knowledge, multimodal reasoning, and reusable generative priors.

\subsubsection{Human-Object Interaction}
\label{sec::interaction_type_I}
Human-object interaction models body dynamics with object state, contact, and affordance. Generative methods increasingly learn reusable priors instead of separate models for each object or action. ViHOI transfers knowledge from pretrained image models to human-object motion synthesis, improving generalization with limited data~\cite{cai2026vihoi}. PrimHOI decomposes interactions into recomposable primitives~\cite{jia2025primhoi}. Uni-HOI jointly models language and interaction sequences autoregressively~\cite{zhang2026uni}, while HOIGPT frames long hand-object sequences as a language-modeling problem~\cite{huang2025hoigpt}. Together, these methods shift from closed interaction classes toward reusable, semantically accessible relational knowledge.

\begin{table}[!ht]
\rowcolors{2}{white}{table_creamyellow}
\centering
\caption{
Representative methods for interaction modeling in human-centric intelligence.
\\[0.3ex]
\textbf{$\bullet$ Modalities:} 
\HumanImage~Exocentric Image,
\HumanVideo~Exocentric Video,
\HumanSkeleton~Pose \& Skeleton,
\HumanMesh~Mesh \& Parametric Body Model,
\Gaze~Gaze Signal,
\Acceleration~Inertial Measurement Signal,
\ActionTrajectory~Action Trajectory,
\Text~Text
and \Audio~Audio.\\
\textbf{$\bullet$ Training:} 
\TrainScratch~Scratch-based Training,
\TrainFull~Full-Parameter Fine-tuning,
\TrainParameter~Parameter-Efficient Tuning,
\TrainInstruction~Instruction Tuning,
\TrainReward~Reward-based Tuning,
\TrainDistillation~Knowledge Distillation,
and \TrainTask~Task-Head Tuning.\\
\textbf{$\bullet$ Inference:} 
\InferenceSemantic~Semantic Augmentation,
\InferenceGuided~Guided Sampling,
\InferenceIterative~Iterative Refinement,
\InferencePreference~Preference-based Selection,
and \InferenceRetrieve~Retrieval Augmentation.\\
\textbf{$\bullet$ Architectures:} 
\SMA~Single-Step Mapping,
\SFA~Sequential Factorization,
\IGA~Iterative Generative,
and \HA~Hybrid. \\
\textbf{$\bullet$ Context:} 
\InteractionModeling~Interaction Modeling.\\
\textbf{$\bullet$ Markers:} 
\Real~Real-World,
\Synthetic~Synthesized,
\Estimated~Estimated,
and \NA~Not Mentioned.
}
\vspace{-7pt}
\resizebox{\linewidth}{!}{
\begin{tabular}{l|c|c|c|c|c|c|c|c|c|c|c}
    \toprule
    \textbf{Method} & \textbf{Venue} & \textbf{Modality} & \textbf{Data Scale} & \textbf{Model Size} & \textbf{Pre.} & \textbf{Inf.}& \textbf{Arc.}  & \textbf{Context} & \textbf{Function} & \textbf{Paper} & \textbf{Web} \\
    \midrule\midrule

    \rowcolor{gray!10} \multicolumn{12}{l}{\textit{\textbf{Sec.~\ref{sec::interaction_type_I}: Human-Object Interaction}}}\\
    \midrule

    ViHOI~\cite{cai2026vihoi} & CVPR'26 & \HumanImage[-0.5ex][0.03\linewidth]~\HumanMesh[-0.5ex][0.03\linewidth]~\Text[-0.5ex][0.03\linewidth] & 10 Hours\textsuperscript{\Real} & 3B & \TrainParameter[0.2ex] & \InferenceGuided[0.2ex] & \IGA[0.2ex] & \InteractionModeling[-0.5ex][0.028\linewidth] & HOI Motion Generation & \href{https://arxiv.org/abs/2603.24383}{\faBook} & \href{https://github.com/MPI-Lab/ViHOI}{\faGithub} \\
\bottomrule
\end{tabular}}
\label{tab::summary_interaction}
\end{table}

A second direction unifies previously separate interaction operations. TriDi jointly generates humans, objects, and interactions rather than assuming fixed object geometry~\cite{petrov2025tridi}. OneHOI combines interaction generation and editing within a diffusion framework~\cite{hoe2026onehoi}. OpenHOI uses a multimodal language model to interpret open-world object semantics before synthesizing interactions~\cite{zhang2026openhoi}. ReGenHOI connects reconstruction and generation through shared geometry~\cite{xu2026regenhoi}. Their foundation-era significance lies in transferring interaction knowledge across tasks and object categories, not merely increasing generative diversity.

Foundation video models bring relational learning into pixel space. CoInteract uses spatially structured co-generation to preserve evolving person-object relations~\cite{luo2026cointeract}. ByteLOOM applies progressive training for geometric consistency in large-scale interaction video generation~\cite{liu2025byteloom}. OmniShow unifies multimodal conditions within a shared human-object video generator~\cite{zhou2026omnishow}. HOMIE extends to reference-conditioned personalization, using MLLM guidance to reason over human-object relations and modality-reference embeddings to preserve subject fidelity across inter- and intra-subject references~\cite{cai2026homie}. Despite improved semantic control and visual consistency, these methods still assess contact and object-state correctness only indirectly.

Other methods add explicit relational grounding. HOI-PAGE structures body-object correspondence through part-level affordances before generating 4D interactions~\cite{li2025hoi}. EasyHOI~\cite{liu2025easyhoi} and ArtHOI~\cite{wang2026arthoi} combine large pretrained models with geometric optimization for in-the-wild reconstruction. InterPrior links generative interaction priors to physics-based control~\cite{xu2026interprior}. These hybrids show that semantic priors and physical constraints can be complementary. We classify human-object interaction here because its primary target is relational feasibility, even when it later supports embodied control.

\begin{insightbox}[\faLightbulb~Open Problem: Relational Grounding in General Interaction Priors]
Foundation models are making human-object interaction more open-ended by transferring knowledge across tasks and objects. Yet broad object semantics do not capture how bodies and objects constrain each other during contact, producing semantically and visually plausible yet relationally inconsistent interactions. A promising direction is to ground general interaction priors in affordance, object-state transitions, and physical feasibility while retaining the adaptability of large-scale pretraining.

\end{insightbox}

\subsubsection{Human-Scene Interaction}
\label{sec::interaction_type_II}
Human-scene interaction extends relational context from objects to the surrounding environment. Generation-oriented work examines whether pretrained knowledge can support broader scene-conditioned behavior. InHabit uses image foundation models to infer plausible human placement~\cite{kister2026inhabit}. FunHSI enables open-vocabulary functional interactions, using language to specify scene use beyond fixed actions~\cite{liu2026open}. TRUMANS learns reusable dynamic patterns from large-scale scene-grounded motion~\cite{jiang2024scaling}. Together, these methods move synthesis beyond memorized categories while retaining the scene as an active behavioral constraint.

A complementary direction reconstructs humans and scenes within a shared spatial frame. Human3R adapts generalizable priors to jointly recover people and environments from video~\cite{chen2025human3r}. SHOW performs
feed-forward reconstruction within a common world coordinate system~\cite{shi2026scene}. UniSH combines scene and human reconstruction priors through large-scale video supervision~\cite{li2026unish}. UniCon3R adds contact-aware joint reconstruction~\cite{sur2026unicon3r}, while IMU-to-4D extends recovery beyond vision through wearable sensing~\cite{hsu2026seeing}. Together, these methods shift foundation-era reconstruction from separate estimates toward reusable models of shared human-scene geometry.

Language models offer a third interface for scene-grounded motion. HIS-GPT enables multimodal understanding of humans in 3D scenes~\cite{zhao2025his}. HSI-GPT aligns scene, motion, and language for unified understanding and generation~\cite{wang2025hsi}. HSI-GPT2 adds complementary temporal scales and generative refinement to improve semantic alignment and physical fidelity~\cite{wang2026hsi}. Uni-Inter unifies motion synthesis across interaction contexts~\cite{liu2025uni}. Together, these systems make scene context part of the foundation-model interface rather than an auxiliary condition for scene-free generation.

\begin{insightbox}[\faLightbulb~Trend: Toward Scene-Grounded Human Foundations]
Foundation-era human-scene modeling is connecting reusable scene knowledge with human behavior and geometry. This allows the environment to serve as a persistent behavioral constraint rather than a static visual condition. Yet language-level understanding may miss the geometric and functional details governing feasibility. Progress may therefore depend on aligning broad scene semantics with the spatial organization of human activity across environments.
\end{insightbox}

\subsubsection{Social Interaction}
\label{sec::interaction_type_III}
Social interaction models relations among people, where behavior depends on reciprocal timing and communicative context. Foundation models broaden this area through multimodal social understanding. HumanOmni-Speaker binds visible participants to speech over time for fine-grained speaker attribution~\cite{bai2026humanomni}. Omni-MMSI-R extends this toward identity-grounded social understanding by integrating multimodal evidence within a shared reasoning model~\cite{li2026omni}. These methods show that general multimodal competence does not ensure reliable attribution, which requires tracking participant-specific evidence across concurrent streams.

Generative methods bring relational structure to multi-person motion and video. SocialStructureHHI explicitly models participant organization in motion generation~\cite{wang2026social}. SocialDirector controls social relations in multi-person video without retraining the foundation model~\cite{ouyang2026socialdirector}. VIM unifies interactive-motion understanding, generation, and reasoning through a shared language interface~\cite{park2025unified}. These systems move beyond independently generated people by modeling how participants condition one another.

Interactive digital humans extend this coupling to online response. X-Streamer learns large-scale audiovisual interaction dynamics for streaming human video generation~\cite{xie2025x}. DyaPlex preserves a pretrained speech model's conversational ability while adding synchronized motion for full-duplex interaction~\cite{nagano2026dyaplex}. ViBES~\cite{zhang2026vibes} and SOLAMI~\cite{Jiang_2025_CVPR} link language reasoning with embodied conversational behavior. Mio scales multimodal interactive intelligence for digital humans~\cite{cai2025towards}. OmniResponse models dyadic interaction as online generation from evolving multimodal context~\cite{luo2026omniresponse}. This marks a foundation-era shift from conditioned gestures to systems that perceive and respond within one temporal loop.

\begin{insightbox}[\faLightbulb~Open Problem: Reciprocal Grounding in Social Foundation Models]
Social foundation models integrate perception, reasoning, and generation within shared multimodal systems. Yet processing multiple participants does not ensure an understanding of reciprocity, which requires attribution and adaptation to evolving responses. Future models may therefore benefit from participant-aware temporal memory and explicit mutual-influence modeling. Cultural variation, ambiguous intent, and calibrated uncertainty remain difficult to capture through scale alone.
\end{insightbox}
\section{Human World Simulation and Embodied Agency}
\label{sec::situated_embodied}
World simulation and embodied agency form the situated-agent perspective of human-centric intelligence. World simulation models how human behavior changes an evolving environment, whereas embodied agency converts human-centered knowledge and experience into executable behavior. In the foundation-model era, pretrained video generators, predictive world models, multimodal reasoning systems, and large-scale control policies are bringing these two levels closer together.

\subsection{World Simulation}
\label{sec::world_simulation}
\begin{wrapfigure}{r}{0.7\textwidth}
    \begin{minipage}{\linewidth}
        \centering
        \vspace{-5pt}
        \includegraphics[width=\linewidth]{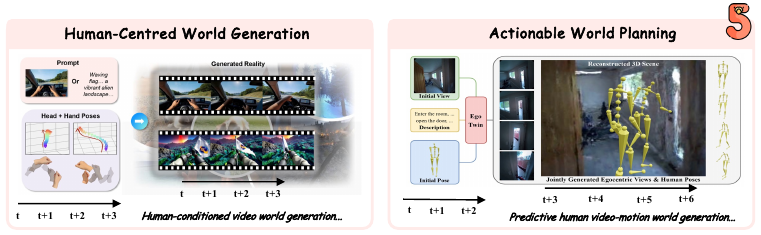}
        \caption{\textbf{Overview of world simulation in human-centric intelligence.} We organize world simulation into human-centered world generation and actionable world planning. The images are originally shown in~\cite{xie2026generated,xiuegotwin}.}
        \label{fig::world_simulation}
        \vspace{-5pt}
    \end{minipage}
\end{wrapfigure}

World simulation connects human behavior with changes in the surrounding environment. We organize this subsection into human-centered world generation, which synthesizes perceptual futures conditioned on human activity, and actionable world planning, which learns action-dependent transitions that can inform decisions and predict futures. This distinction separates visually plausible simulation from predictive models whose internal variables or outputs can support embodied action, as illustrated in Fig.~\ref{fig::world_simulation}.

\subsubsection{Human-Centered World Generation}
\label{sec::world_type_I}

Human-centered world generation adapts large video priors to model environments evolving around human activity. Generated Reality conditions interactive video on tracked head and hand motion, turning a pretrained model into a controllable egocentric simulator~\cite{xie2026generated}. Hand2World uses explicit hand and camera conditioning for autoregressive generation, improving long-horizon stability and separating hand motion from viewpoint change~\cite{wang2026hand2world}. DWM specializes video generation for dexterous hand-object interaction~\cite{kim2026dexterous}, while PlayerOne links exocentric body motion to first-person world evolution~\cite{tu2025playerone}. Ji et al. extend this direction to real-time upper-body interaction by combining a multi-scale implicit human state with discrete contact commands and distilling a pretrained video model for streaming local world generation~\cite{ji2026real}. Together, these methods reorganize general video priors around human motion and interaction as drivers of world change.

A further direction strengthens the persistence and controllability of worlds. AnchorWorld conditions egocentric simulation on reusable scene views and 3D motion, keeping evolution tied to a specific environment~\cite{li2026anchorworld}. EgoForge incorporates task goals and optimizes temporal causality, scene consistency, and goal completion~\cite{shen2026egoforge}. EgoControl generates first-person video from full-body trajectories~\cite{pallotta2026egocontrol}, while EgoTwin models body dynamics and egocentric observations~\cite{xiuegotwin}. EgoX~\cite{kang2026egox} and EgoWorld~\cite{park2025egoworld} synthesize first-person experience from exocentric evidence. These methods shift video generation from unconstrained continuation toward simulation organized around the human agent, though their outputs remain perceptual rather than executable.

\begin{insightbox}[\faLightbulb~Trend: From Video Priors to Human-Conditioned World Evolution]
Foundation video models provide visual and temporal priors, but human-centered simulation requires environments to respond to bodily motion and interaction. The emerging shift is from plausible video generation to worlds that evolve consistently around situated humans. Progress may depend less on visual scale than on persistent world state, causal response, and long-horizon controllability.

\end{insightbox}

\begin{table}[!ht]
\rowcolors{2}{white}{table_creamyellow}
\centering
\caption{
Representative methods for world simulation and embodied agency in human-centric intelligence.
\\[0.3ex]
\textbf{$\bullet$ Modalities:} 
\HumanVideo~Exocentric Video,
\EgoImage~Egocentric Image,
\EgoVideo~Egocentric Video,
\HumanSkeleton~Pose \& Skeleton,
\HumanMesh~Mesh \& Parametric Body Model,
\ActionTrajectory~Action Trajectory,
\ActionControl~Control Signal,
\Text~Text
and \Audio~Audio.\\
\textbf{$\bullet$ Training:} 
\TrainScratch~Scratch-based Training,
\TrainFull~Full-Parameter Fine-tuning,
\TrainParameter~Parameter-Efficient Tuning,
\TrainInstruction~Instruction Tuning,
\TrainReward~Reward-based Tuning,
\TrainDistillation~Knowledge Distillation,
and \TrainTask~Task-Head Tuning.\\
\textbf{$\bullet$ Inference:} 
\InferenceSemantic~Semantic Augmentation,
\InferenceGuided~Guided Sampling,
\InferenceIterative~Iterative Refinement,
\InferencePreference~Preference-based Selection,
and \InferenceRetrieve~Retrieval Augmentation.\\
\textbf{$\bullet$ Architectures:} 
\SMA~Single-Step Mapping,
\SFA~Sequential Factorization,
\IGA~Iterative Generative,
and \HA~Hybrid. \\
\textbf{$\bullet$ Context:} 
\WorldSimulation~World Simulation,
and \EmbodiedAgent~Embodied Agency.  \\
\textbf{$\bullet$ Markers:} 
\Real~Real-World,
\Synthetic~Synthesized,
\Estimated~Estimated,
and \NA~Not Mentioned.
}
\vspace{-7pt}
\resizebox{\linewidth}{!}{
\begin{tabular}{l|c|c|c|c|c|c|c|c|c|c|c}
    \toprule
    \textbf{Method} & \textbf{Venue} & \textbf{Modality} & \textbf{Data Scale} & \textbf{Model Size} & \textbf{Train.} & \textbf{Inf.}& \textbf{Arc.}  & \textbf{Context} & \textbf{Function} & \textbf{Paper} & \textbf{Web} \\
    \midrule\midrule

    \rowcolor{gray!10} \multicolumn{12}{l}{\textit{\textbf{Sec.~\ref{sec::world_type_I}: Human-Centered World Generation}}}\\
    \midrule

    Generated Reality~\cite{xie2026generated}  & CVPRF'26 & \EgoImage[-0.5ex][0.03\linewidth]~\EgoVideo[-0.5ex][0.03\linewidth]~\HumanSkeleton[-0.5ex][0.03\linewidth]~\Text[-0.5ex][0.03\linewidth] & 5.8K Samples\textsuperscript{\Real} & 5B-14B  & \TrainParameter[0.2ex]~\TrainDistillation[0.2ex]~\TrainTask[0.2ex] & \InferenceGuided[0.2ex] & \IGA[0.2ex]  & \WorldSimulation[-0.5ex][0.028\linewidth]  & Hand-Guided Interactive Generation  & \href{https://arxiv.org/abs/2602.18422}{\faBook}  & \href{https://codeysun.github.io/generated-reality/}{\faHome}\\

    Hand2World~\cite{wang2026hand2world} & arXiv'26  & \EgoVideo[-0.5ex][0.03\linewidth]~\Text[-0.5ex][0.03\linewidth] & \NA & 1.3B  & \TrainParameter[0.2ex]~\TrainDistillation[0.2ex]  & \InferenceRetrieve[0.2ex] & \IGA[0.2ex]  & \WorldSimulation[-0.5ex][0.028\linewidth]  & Hand-Guided Interactive Generation  & \href{https://arxiv.org/abs/2602.09600}{\faBook}  & \href{https://github.com/NTUYWANG103/Hand2World}{\faGithub} \\
\bottomrule
\end{tabular}}
\label{tab::summary_world_embodied}
\end{table}

\subsubsection{Actionable World Planning}
\label{sec::world_type_II}

Actionable world modeling connects candidate actions to their consequences rather than only modeling human-conditioned appearance. PEVA uses whole-body pose to predict egocentric futures, linking intended motion 
to visual outcomes~\cite{bai2026whole}. LOME conditions an egocentric world model on manipulation actions for interaction learning~\cite{gao2026lome}. EgoHOI specializes this approach for photorealistic hand-object interaction synthesis~\cite{li2026egocentric}, while EgoMAN predicts 3D hand trajectories from egocentric evidence~\cite{chen2025flowing}. These methods add action structure to pretrained visual models rather than treating the future as passive continuation.

Recent work couples visual prediction with action learning. HandWorld jointly generates hand actions and corresponding videos within one model~\cite{sun2026handworld}. DexWM predicts latent futures conditioned on dexterous actions and transfers interaction knowledge from human video to robotic manipulation~\cite{goswami2026worldmodelslearningdexterous}. These approaches suggest that world-model value depends not only on visual realism but also on capturing control-relevant changes within the predictive space.

A related line develops representations for long-horizon reasoning and planning. EgoAgent jointly predicts future states and actions from egocentric experience~\cite{chen2025egoagent}. EgoSim maintains an updatable 3D world state for closed-loop interaction~\cite{hao2026egosim}. EgoExo-WM transfers interaction knowledge from exocentric human video to an egocentric action-conditioned model, evaluating candidate motions through predicted outcomes~\cite{tran2026egoexo}. LWM simplifies planning by lifting low-level predictions into a waypoint-level action space~\cite{wang2026lifting}. These methods bridge world simulation and embodied agency by making prediction part of action selection.

\begin{insightbox}[\faLightbulb~Open Problem: Actionability Beyond Predictive Fidelity]
Foundation world models produce convincing futures, but perceptual accuracy does not ensure decision utility. Actionability further requires reliable responses to interventions, separation of consequential changes from incidental visual variation, and calibrated uncertainty. The relation between generative fidelity and planning effectiveness remains central to human-centered world modeling.
\end{insightbox}

\subsection{Embodied Agency}
\label{sec::embodied_agency}
\begin{figure}[h]
    \centering
    \includegraphics[width=\linewidth]{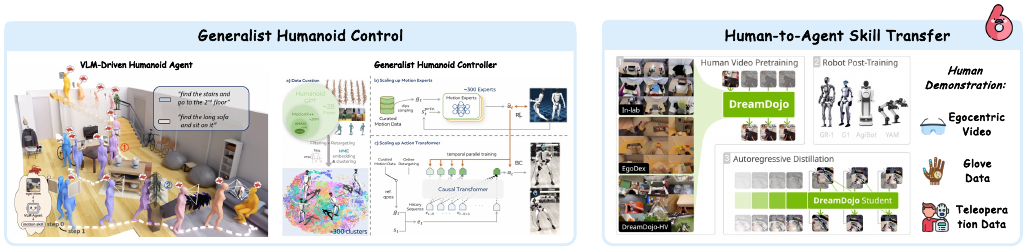}
    \caption{\textbf{Overview of embodied agency in human-centric intelligence.} We organize embodied agency into generalist humanoid control and human-to-agent skill transfer. The images are originally shown in~\cite{li2026humanclaw,qi2026humanoid,gao2026dreamdojo}.}
\label{fig::embodied_agency}
\end{figure}

Embodied agency concerns the acquisition and execution of physically grounded capabilities. We distinguish generalist humanoid control, which develops reusable policies for human-like bodies, from human-to-agent skill transfer, which derives actionable knowledge from human behavior and adapts it to other embodiments. In the foundation-model era, these directions are increasingly connected through shared multimodal interfaces, scalable behavioral pretraining, and large collections of human experience, as shown in Fig.~\ref{fig::embodied_agency}.

\subsubsection{Generalist Humanoid Control}
\label{sec::embodied_type_I}
Humanoid control connects high-level semantics to executable behavior. VGHuman combines visual understanding with motion generation to animate humanoids in reconstructed environments~\cite{ye2026visually}. HOI-HLI translates instructions into structured human-object interaction policies~\cite{wu2025human}. BiBo uses an off-the-shelf vision-language model to interpret tasks before grounding its outputs in humanoid control~\cite{jian2025endowing}. UniHSI converts prompted human-scene interactions into contact sequences executed by a physics-based controller~\cite{xiao2024unified}. VLM-RMD uses vision-language reasoning to construct rewards for interaction-policy learning~\cite{deng2025human}. These methods use foundation models as semantic engines while retaining specialized controllers.

A second line learns reusable motor priors across behaviors. MaskedMimic frames physics-based control as conditional completion of partial motions~\cite{tessler2024maskedmimic}. InterMimic distills noisy human-object interaction data into robust whole-body skills~\cite{xu2025intermimic}, while TokenHSI organizes interaction capabilities through shared task and control tokens~\cite{pan2025tokenhsi}. These approaches provide control abstractions beyond separately trained policies.

Recent systems examine foundation-style scaling and transfer in humanoid control. SCRIPT combines large-scale motion pretraining, language conditioning, and reinforcement post-training within a diffusion policy~\cite{zhang2026script}. SONIC studies joint scaling of motion data, controller capacity, and compute for whole-body tracking~\cite{luo2025sonic}. Humanoid-GPT scales structured motion data for zero-shot tracking across diverse movements~\cite{qi2026humanoid}. GPC tokenizes physical behavior and uses autoregressive pretraining to learn a reusable generative controller~\cite{shi2026gpc}. These works shift the focus from semantic breadth to whether large-scale motor pretraining produces transferable physical competence.

Generalist architectures connect perception, language, and action. GR00T N1 combines vision-language reasoning with action generation in a generalist humanoid policy~\cite{bjorck2025gr00t}. WholeBodyVLA uses a shared latent interface for whole-body locomotion and manipulation~\cite{jiang2025wholebodyvla}. BumbleBee distills expert controllers into a generalist whole-body policy~\cite{wang2026experts}, while CLAIMS expands capabilities through iterative closed-loop motion synthesis~\cite{xu2026iterative}. BFM-Zero learns a behavioral space that supports multiple objectives without policy retraining~\cite{li2025bfm}. Together, these methods pursue reusable control foundations rather than independent skills.

Specialized execution models remain important. UniTracker~\cite{yin2026unitracker} and GMT~\cite{chen2025gmt} learn reusable tracking policies. UniPhys unifies planning and control through a diffusion framework~\cite{wu2025uniphys}. RoboPerform enables audio-conditioned expressive locomotion~\cite{li2026you}, while RoboGhost connects language instructions to motion-latent control~\cite{li2025language}. These systems show that broad interfaces still require stable, responsive physical execution.

\begin{insightbox}[\faLightbulb~Trend: Scalable Motor Priors as a Control Foundation]
Humanoid control is moving from isolated skills toward reusable motor priors adaptable through high-level conditions. This brings foundation-model principles into control, where generality requires both behavioral coverage and physical reliability. Scaling may improve transfer, but how broadly a shared policy can generalize without weakening execution stability remains unclear.
\end{insightbox}

\subsubsection{Human-to-Agent Skill Transfer}
\label{sec::embodied_type_II}

Human-to-agent skill transfer uses human activity as scalable embodied supervision. It addresses a limitation of embodied foundation models: robot trajectories provide precise action labels but are costly and narrow, whereas human video offers broader behavioral and environmental coverage. HumanScale studies this tradeoff and demonstrates the potential of egocentric video for embodied pretraining~\cite{ma2026humanscale}. Being-H0 pretrains a vision-language-action model on large-scale human video before adapting it to robotic manipulation~\cite{luo2025being}. VITRA converts real-world activity videos into action supervision for VLA pretraining~\cite{li2025scalable}. EgoVLA predicts wrist and hand actions from egocentric video and retargets them to robots~\cite{yang2025egovla}. EgoScale extends this paradigm to dexterous manipulation and studies how human-data scale affects downstream performance~\cite{zheng2026egoscale}.

Since human video lacks native robot actions, another line learns intermediate action abstractions. Moto encodes motion changes as latent tokens linking human-video pretraining to robot control~\cite{chen2025moto}. LARA aligns latent-action learning with VLA training so that visual dynamics and executable trajectories refine each other~\cite{liu2026lara}. Being-H0.7 uses future-informed latent variables to provide predictive action structure without generating future frames during inference~\cite{luo2026being}. DreamDojo learns continuous latent actions from large-scale egocentric video and adapts its world model to robots~\cite{gao2026dreamdojo}. ActiveMimic transfers active human viewing behavior to robot perception and manipulation~\cite{lin2026activemimic}. These methods suggest that human experience transfers through structured change and intent rather than direct motion imitation.

A complementary direction addresses embodiment alignment. HumanEgo abstracts hand-object relations from human demonstrations into interaction tokens~\cite{he2026humanego}. Human$_0$ combines in-the-wild activity with task-oriented data for scalable manipulation supervision~\cite{cai2025n}. EgoMimic adapts egocentric human video for robot imitation~\cite{kareer2025egomimic}, while VIPA-VLA aligns human visual evidence with physical action spaces during pretraining~\cite{feng2026spatial}. UniDex provides a transfer pipeline from egocentric video to dexterous hand control~\cite{zhang2026unidex}.

Targeted methods address specific human-to-robot gaps. ZeroMimic distills manipulation skills from web videos~\cite{shi2025zeromimic}. HR-Align aligns visual pretraining across human and robot domains~\cite{zhou2025mitigating}. ManipTrans retargets bimanual human motion for dexterous robot control~\cite{li2025maniptrans}. ActiveUMI adds active perception to learning from robot-free demonstrations~\cite{zeng2025activeumi}. SUGAR transfers human video to generalizable humanoid loco-manipulation~\cite{wu2026sugar}, while HUG derives dexterous grasping from human hand observations~\cite{wu2026human}. Together, these works show that human experience can scale embodied foundation models when transferable knowledge is disentangled from embodiment-specific appearance and kinematics.

\begin{insightbox}[\faLightbulb~Open Problem: Foundation Pretraining Across the Embodiment Gap]
Human behavior provides scalable supervision for embodied learning, but observations do not directly specify actions executable by another body. Current methods bridge this gap through learned action abstractions and embodiment alignment, yet their generality across bodies and environments remains uncertain. A central question is whether human experience can yield reusable action representations that preserve task-relevant structure while adapting across embodiments and control spaces.
\end{insightbox}
\section{Datasets, Benchmarks, and Metrics} 
\label{sec::datasets_benchmarks_metrics}

Progress in human-centric intelligence depends on the resources supporting model development and the criteria used for evaluation. Accordingly, this section first reviews representative datasets and benchmarks, and then organizes commonly used metrics. Together, these two parts clarify what current models learn from, how their capabilities are evaluated, and where existing evaluation practices remain incomplete.

\subsection{Datasets and Benchmarks}
\label{sec::datasets_and_benchmarks}

Datasets and benchmarks play complementary roles in human-centric intelligence: datasets provide the observations and annotations used for model development, whereas benchmarks define standardized tasks and protocols for comparing model capabilities. Following the progression of our human context taxonomy, we organize these resources into four groups covering human subjects, human dynamics, human interactions, and human embodiment. This organization associates heterogeneous resources with the primary human context they support rather than separating them solely by modality or individual task.

\subsubsection{Human Subject Resources}
\label{sec::human_subject_resources}
Human subject resources support the perception, identification, and spatial modeling of humans as observable subjects. We group them into visual human observation, multisensory human sensing, human identity understanding, and renderable human geometry, progressing from visual and multisensory evidence to identity correspondence and explicit 3D assets.

\begin{figure}[h]
    \centering
    \includegraphics[width=\linewidth]{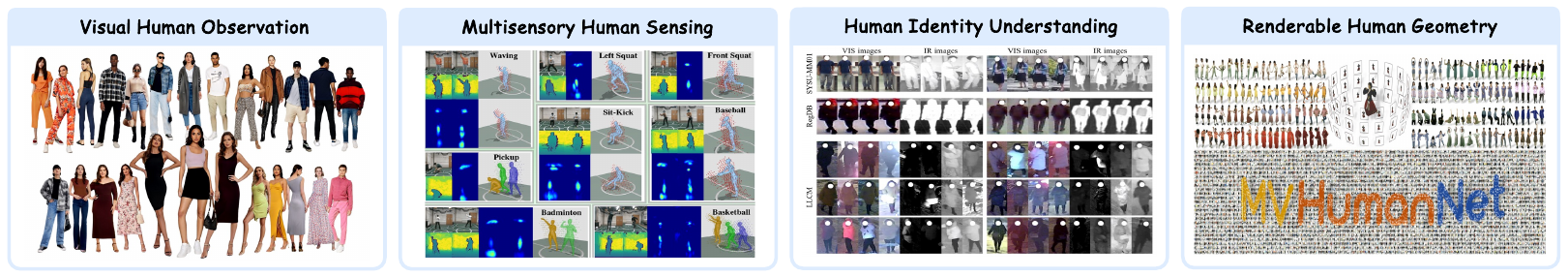}
    \caption{\textbf{Representative human subject resources in human-centric intelligence.} From left to right, the columns show StyleGAN-Human~\cite{fu2022stylegan}, M4Human~\cite{fan2026m4human}, LLCM~\cite{zhang2023diverse}, and MVHumanNet++~\cite{li2026mvhumannet++}.}
\label{fig::human_subject_resources}
\end{figure}

\textbf{Visual Human Observation.}
Visual human observation resources capture how people appear and how such evidence can be queried. Early datasets support appearance modeling and localized perception, including full-body generation in StyleGAN-Human~\cite{fu2022stylegan}, hand gesture recognition in HaGRID~\cite{kapitanov2024hagrid}, and facial video attributes in CelebV-HQ~\cite{zhu2022celebv} and CelebV-Text~\cite{yu2023celebv}. Recent benchmarks introduce language-based evaluation: FaceBench~\cite{wang2025facebench} examines fine-grained facial VQA, Face-Human-Bench~\cite{qin2026face} evaluates face and human understanding in multimodal assistants, and MHPR~\cite{wang2026mhpr} stresses multidimensional perception and reasoning. Together, these resources shift from collecting visual evidence for recognition and generation toward testing whether models ground human attributes and relations in language. They remain dominated by visible appearance, leaving geometric, temporal, and nonvisual context to later categories.

\textbf{Multisensory Human Sensing.}
Multisensory human sensing resources address settings where visible appearance is insufficient for robust human modeling. They align camera observations with complementary signals to recover body geometry and motion under occlusion, poor illumination, or privacy constraints. HuMMan~\cite{cai2022humman} provides large-scale multimodal 4D capture with synchronized geometry and motion annotations. RELI11D~\cite{yan2024reli11d}, mRI~\cite{an2022mri}, and MM-Fi~\cite{yang2023mm} broaden this setting across LiDAR, radar, inertial, event, and wireless sensing. Radar-centered benchmarks such as mmBody~\cite{chen2022mmbody}, HuPR~\cite{lee2023hupr}, RT-Pose~\cite{ho2024rt}, MMVR~\cite{rahman2024mmvr}, and M4Human~\cite{fan2026m4human} further extend pose and mesh recovery beyond cameras. These resources shift the emphasis toward cross-sensor calibration and physically grounded recovery, but remain limited by controlled data collection, sensor-specific protocols, and geometry-centered evaluation.

\textbf{Human Identity Understanding.}
Human identity resources test whether models preserve person-level consistency across changing queries and observation conditions. Unlike generic visual datasets, they evaluate matching and retrieval when direct visual comparison becomes unreliable. MALS~\cite{yang2023towards} extends person retrieval from attribute tags to large-scale language-based search. LLCM~\cite{zhang2023diverse} studies visible-infrared re-identification, while L2RW~\cite{jiang2025laboratory} targets privacy-preserved cross-modality matching. MP-ReID~\cite{ha2025multi} and AG-VPReID~\cite{nguyen2025ag} introduce multi-platform and aerial-ground video scenarios with substantial viewpoint and trajectory variation. These resources reflect a shift from closed-set RGB re-identification toward language-conditioned, cross-modality, and cross-platform matching. They also raise privacy, demographic-bias, and surveillance concerns, making careful governance as important as scale.

\begin{table*}[!ht]
\rowcolors{2}{white}{table_creamyellow}
\centering
\caption{Representative datasets and benchmarks for renderable human geometry in human-centric intelligence.
\\[0.3ex]
\textbf{$\bullet$ Modalities:} 
\HumanImage~Exocentric Image,
\HumanVideo~Exocentric Video,
\EgoImage~Egocentric Image,
\EgoVideo~Egocentric Video,
\HumanDepth~Depth Map,
\HumanNormal~Normal Map,
\HumanSkeleton~Pose \& Skeleton,
\HumanPointCloud~Point Cloud,
\HumanMesh~Mesh \& Parametric Body Model,
and \Text~Text. \\
\textbf{$\bullet$ Markers:} 
\DataSelfCaptured~Self-Captured,
\DataPublicCurated~Public-Curated,
\DataWebCollected~Web-Collected,
and \DataSynthesized~Synthesized.
}
\vspace{-7pt}
\label{tab::summary_db_renderable}
\resizebox{\linewidth}{!}{
\begin{tabular}{l|c|c|c|c|c|c|c|c|c|c|c|c}
    \toprule
    \textbf{Name} & \textbf{Year} & \textbf{Modality} & \textbf{Scale} & \textbf{Subjects} & \textbf{Source} & \textbf{Scene} & \textbf{Human Region} & \textbf{View Type} & \textbf{Multi-Person} & \textbf{Scope} & \textbf{Paper} & \textbf{Web} \\
    \midrule\midrule
    VolHuMe~\cite{martinelli2026volhume} & 2026 & \HumanImage~\HumanDepth~\HumanMesh~\HumanPointCloud & 156K Frames & 104 & \DataSelfCaptured[0.2ex] & Studio & Full-Body & \Exo[0.2ex] & \No & Volumetric Human Reconstruction & \href{https://arxiv.org/abs/2606.23062}{\faBook} & \NA \\
    MVHumanNet++~\cite{li2026mvhumannet++} & 2026 & \HumanVideo~\HumanImage~\HumanMesh~\Text & 645.1M Frames & 4.5K & \DataSelfCaptured[0.2ex] & Studio & Full-Body & \Exo[0.2ex] & \No & 3D Human Digitization & \href{https://arxiv.org/abs/2505.01838}{\faBook} & \href{https://kevinlee09.github.io/research/MVHumanNet++/}{\faHome} \\
    HumanOLAT~\cite{teufel2025humanolat} & 2025 & \HumanVideo[-0.5ex][0.03\linewidth]~\HumanMesh[-0.5ex][0.028\linewidth]~\HumanNormal[-0.5ex][0.028\linewidth] & 850K Frames & 21 & \DataSelfCaptured[0.2ex] & Studio & Full-Body & \Exo[0.2ex] & \No & Human Relighting & \href{https://openaccess.thecvf.com/content/ICCV2025/html/Teufel_HumanOLAT_A_Large-Scale_Dataset_for_Full-Body_Human_Relighting_and_Novel-View_ICCV_2025_paper.html}{\faBook} & \href{https://vcai.mpi-inf.mpg.de/projects/HumanOLAT/}{\faHome} \\
    WildAvatar~\cite{huang2025wildavatar} & 2025 & \HumanVideo~\HumanMesh & 10.6K Clips & 10K & \DataWebCollected[0.2ex] & In-the-Wild & Full-Body & \Exo[0.2ex] & \Yes & In-the-Wild 3D Avatars & \href{https://openaccess.thecvf.com/content/CVPR2025/html/Huang_WildAvatar_Learning_In-the-wild_3D_Avatars_from_the_Web_CVPR_2025_paper.html}{\faBook} & \href{https://wildavatar.github.io/}{\faGithub} \\
    HuGe100K~\cite{zhuang2025idol} & 2025 & \HumanImage~\HumanMesh & 2.4M Images & 100K & \DataPublicCurated[0.2ex]~\DataSynthesized[0.2ex] & Studio & Full-Body & \Exo[0.2ex] & \No & 3D Human Reconstruction & \href{https://openaccess.thecvf.com/content/CVPR2025/html/Zhuang_IDOL_Instant_Photorealistic_3D_Human_Creation_from_a_Single_Image_CVPR_2025_paper.html}{\faBook} & \href{https://github.com/yiyuzhuang/IDOL}{\faGithub} \\
    PKU-DyMVHumans~\cite{zheng2024pku} & 2024 & \HumanVideo~\HumanMesh & 8.2M Frames & 32 & \DataSelfCaptured[0.2ex] & Studio & Full-Body & \Exo[0.2ex] & \Yes & Dynamic Human Modeling & \href{https://arxiv.org/abs/2403.16080}{\faBook} & \href{https://pku-dymvhumans.github.io/}{\faHome}  \\
    Codec Avatar Studio~\cite{martinez2024codec} & 2024 & \HumanVideo~\HumanImage~\EgoImage~\EgoVideo~\HumanMesh~\HumanDepth & 217M Images & 260 & \DataSelfCaptured[0.2ex] & Studio & Full-Body & \Ego[0.2ex]~\Exo[0.2ex] & \No & Driveable Avatar Capture & \href{https://proceedings.neurips.cc/paper_files/paper/2024/hash/9712b78386cebdc3db7f1a48c2d20edb-Abstract-Datasets_and_Benchmarks_Track.html}{\faBook} & \href{https://github.com/facebookresearch/goliath}{\faGithub} \\
    HumanRF~\cite{icsik2023humanrf} & 2023 & \HumanVideo~\HumanMesh & 39.7K Frames & 8 & \DataSelfCaptured[0.2ex] & Studio & Full-Body & \Exo[0.2ex] & \No & Human Neural Rendering & \href{https://doi.org/10.1145/3592415}{\faBook} & \href{https://actors-hq.com/}{\faHome} \\
    NeRSemble~\cite{kirschstein2023nersemble} & 2023 & \HumanVideo~\HumanMesh & 31.7M Frames & 220 & \DataSelfCaptured[0.2ex] & Studio & Head & \Exo[0.2ex] & \No & Head Reconstruction & \href{https://doi.org/10.1145/3592455}{\faBook} & \href{https://tobias-kirschstein.github.io/nersemble/}{\faHome} \\
    SynBody~\cite{yang2023synbody} & 2023 & \HumanImage~\HumanSkeleton~\HumanMesh & 1.2M Images & 10K & \DataSynthesized[0.2ex] & Mixed & Full-Body & \Exo[0.2ex] & \Yes & Synthetic 3D Human Modeling & \href{https://arxiv.org/abs/2303.17368}{\faBook} & \href{https://maoxie.github.io/SynBody/}{\faHome} \\
    DNA-Rendering~\cite{cheng2023dna} & 2023 & \HumanVideo~\HumanImage~\HumanSkeleton~\HumanMesh & 67.5M Frames & 500 & \DataSelfCaptured[0.2ex] & Studio & Full-Body & \Exo[0.2ex] & \No & Human-Centric Rendering & \href{https://arxiv.org/abs/2307.10173}{\faBook} & \href{https://dna-rendering.github.io/}{\faHome} \\
    \bottomrule
\end{tabular}
}
\end{table*}

\textbf{Renderable Human Geometry.}
Renderable human geometry resources provide 3D, 4D, and neural-rendering assets for reconstruction, animation, relighting, and controllable synthesis, as summarized in Table~\ref{tab::summary_db_renderable}. VolHuMe~\cite{martinelli2026volhume}, MVHumanNet++~\cite{li2026mvhumannet++}, and PKU-DyMVHumans~\cite{zheng2024pku} provide volumetric, multi-view, and dynamic captures for geometry-aware digitization and temporal reconstruction. HumanOLAT~\cite{teufel2025humanolat} targets full-body relighting and novel-view synthesis, while WildAvatar~\cite{huang2025wildavatar} and IDOL~\cite{zhuang2025idol} extend avatar creation to in-the-wild videos and single images. Neural-rendering resources such as Codec Avatar Studio~\cite{martinez2024codec}, HumanRF~\cite{icsik2023humanrf}, NeRSemble~\cite{kirschstein2023nersemble}, SynBody~\cite{yang2023synbody}, and DNA-Rendering~\cite{cheng2023dna} support deformable and animation-ready human priors. Together, these resources shift human data from image observations to renderable geometry, but still trade capture scale against diversity and annotation richness.

\subsubsection{Human Dynamics Resources}
\label{sec::human_dynamic_resources}
\begin{figure}[h]
    \centering
    \includegraphics[width=\linewidth]{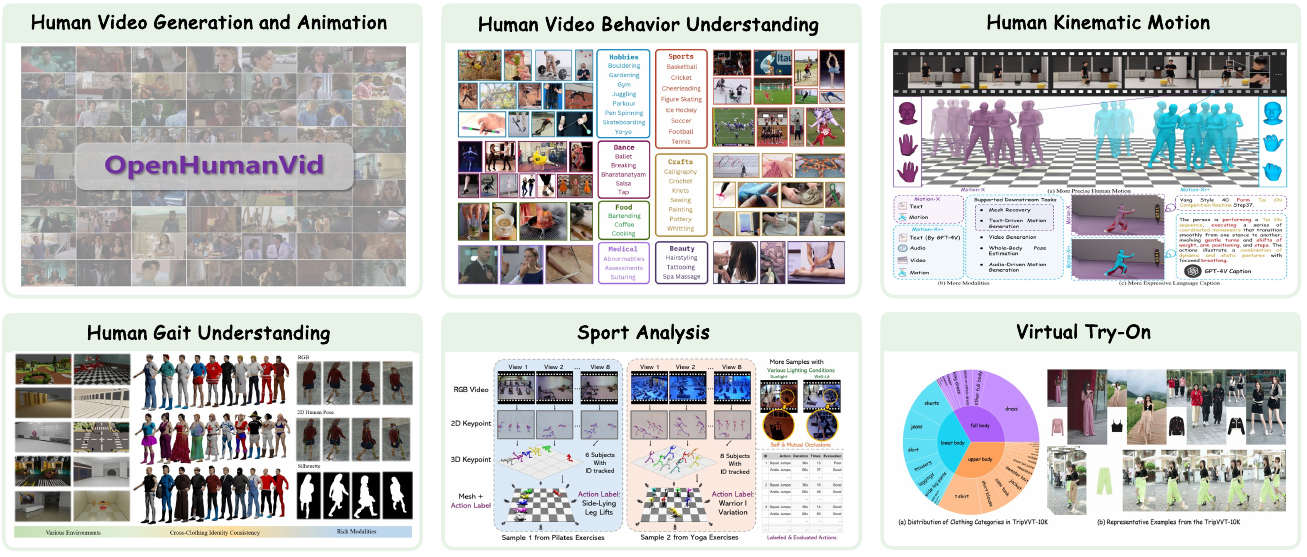}
    \caption{\textbf{Representative human dynamics resources in human-centric intelligence.} From left to right, the top row shows OpenHumanVid~\cite{li2025openhumanvid}, VideoNet~\cite{yadav2026videonet}, and Motion-X++~\cite{zhang2025motionxplusplus}, while the bottom row shows BarbieGait~\cite{cai2026barbiegait}, M3GYM~\cite{xu2025m3gym}, and TripVVT~\cite{shao2026tripvvt}.}
\label{fig::human_dynamics_resources}
\end{figure}

Human dynamics resources capture temporally varying behavior and condition-dependent appearance. In Fig.~\ref{fig::human_dynamics_resources}, we organize them by their primary task settings into human video generation and animation, video behavior understanding, kinematic motion, gait understanding, sport analysis, and virtual try-on, allowing the distinct supervision and evaluation requirements of these domains to be examined systematically.

\textbf{Human Video Generation and Animation.}
Human video generation and animation resources support realistic, temporally coherent, and controllable synthesis. Early datasets target talking-face animation~\cite{zhang2021flow} and responsive listening~\cite{zhou2022responsive}. Newer resources provide larger and more controllable body and face video collections, including HumanVid~\cite{wang2024humanvid}, OpenHumanVid~\cite{li2025openhumanvid}, DH-FaceVid-1K~\cite{di2025dh}, and OmniHuman~\cite{zhu2026omnihuman}. HumanScore~\cite{fang2026humanscore} and AVBench~\cite{yang2026avbench} shift from collection to evaluation by assessing motion plausibility, human alignment, and audio-video consistency. Reliable evaluation of long-horizon and interactive behavior remains limited, particularly for subtle motion errors.

\textbf{Human Video Behavior Understanding.}
Human video behavior resources test recognition, localization, and reasoning from temporal visual evidence. MLLM-oriented benchmarks such as HumanVideo-MME~\cite{cai2025humanvideo}, HumanMoveVQA~\cite{gera2026humanmovevqa}, HumanVBench~\cite{zhou2026humanvbench}, FineBench~\cite{faure2026finebench}, MotionBench~\cite{hong2025motionbench}, and ActionAtlas~\cite{salehi2024actionatlas} move beyond coarse labels toward fine-grained motion understanding and question answering. Larger resources, including HumanNet~\cite{deng2026humannet}, VideoNet~\cite{yadav2026videonet}, OctoNet~\cite{yuan2026octonet}, OMG-Bench~\cite{chang2026omg}, and MMAD~\cite{li2025mmad}, broaden coverage to long-form video, complex activities, and micro-actions. However, many evaluations remain label- or template-driven, limiting assessment of causal, social, and embodied reasoning.

\textbf{Human Kinematic Motion Resources.}
Human kinematic motion resources encode dynamics through structured body trajectories and conditioned annotations, as shown in Table~\ref{tab::summary_db_motion}. BABEL~\cite{punnakkal2021babel}, HumanML3D~\cite{guo2022generating}, Motion-X~\cite{lin2023motion}, and Motion-X++~\cite{zhang2025motionxplusplus} establish large-scale motion-language supervision. Recent resources broaden generation and evaluation through OpenT2M~\cite{cao2026opent2m}, Being-M0.5~\cite{cao2025being}, Go-to-Zero~\cite{fan2025go}, Being-M0~\cite{wang2024scaling}, RoMo~\cite{zhang2026romo}, SnapMoGen~\cite{guo2025snapmogen}, and QUEST~\cite{lin2025quest}. MRBench further supports motion-text retrieval evaluation across heterogeneous motions and description granularities~\cite{liu2026mrbench}. BEDLAM~\cite{black2023bedlam} and BEDLAM2.0~\cite{tesch2026bedlam2} provide synthetic pose and shape data, while AIST++~\cite{li2021ai}, speech-conditioned motion~\cite{yi2023generating}, FineDance~\cite{li2023finedance}, Motion Anything~\cite{zhang2025motion}, and MDD~\cite{gupta2025mdd} introduce specialized conditions. These resources remain limited in physical and scene grounding, long-horizon intent, social interaction, and transfer from mocap or synthetic data to in-the-wild motion.

\begin{table*}[!t]
\rowcolors{2}{white}{table_creamyellow}
\centering
\caption{Representative datasets and benchmarks for human kinematic motion modeling in human-centric intelligence.
\\[0.3ex]
\textbf{$\bullet$ Modalities:} 
\HumanVideo~Exocentric Video,
\HumanDepth~Depth Map,
\HumanSkeleton~Pose \& Skeleton,
\HumanMesh~Mesh \& Parametric Body Model,
\ActionControl~Control Signal,
\Text~Text,
and \Audio~Audio.\\
\textbf{$\bullet$ Markers:} 
\DataSelfCaptured~Self-Captured,
\DataPublicCurated~Public-Curated,
\DataWebCollected~Web-Collected,
and \DataSynthesized~Synthesized.
}
\vspace{-7pt}
\label{tab::summary_db_motion}
\resizebox{\linewidth}{!}{
\begin{tabular}{l|c|c|c|c|c|c|c|c|c|c|c|c}
    \toprule
    \textbf{Name} & \textbf{Year} & \textbf{Modality} & \textbf{Scale} & \textbf{Subjects} & \textbf{Source} & \textbf{Scene} & \textbf{Human Region} & \textbf{View Type} & \textbf{Multi-Person} & \textbf{Scope} & \textbf{Paper} & \textbf{Web} \\
    \midrule\midrule
    OpenT2M~\cite{cao2026opent2m} & 2026 & \HumanSkeleton~\Text & 2.8K Hours & \NA & \DataWebCollected[0.2ex]~\DataPublicCurated[0.2ex] & Mixed & Full-Body & \Exo[0.2ex] & \No & Motion Generation & \href{https://arxiv.org/abs/2603.18623}{\faBook} & \href{https://research.beingbeyond.com/opent2m}{\faHome} \\
    HuMo100M~\cite{cao2025being} & 2025 & \HumanVideo~\HumanSkeleton~\Text & 5M Motions & \NA & \DataWebCollected[0.2ex]~\DataPublicCurated[0.2ex] & Mixed & Full-Body & \Exo[0.2ex] & \No &  Motion Generation & \href{https://arxiv.org/abs/2508.07863}{\faBook} & \href{https://beingbeyond.github.io/Being-M0.5}{\faHome} \\
    MotionMillion~\cite{fan2025go} & 2025 & \HumanSkeleton~\Text & 2K Hours & \NA & \DataWebCollected[0.2ex]~\DataPublicCurated[0.2ex] & Mixed & Full-Body & \Exo[0.2ex] & \No & Motion Generation & \href{https://arxiv.org/abs/2507.07095}{\faBook} & \href{https://github.com/VankouF/MotionMillion-Codes}{\faGithub} \\
    MotionLib~\cite{wang2024scaling} & 2025 & \HumanVideo~\HumanMesh~\HumanDepth~\Text & 1.45K Hours & \NA & \DataWebCollected[0.2ex]~\DataPublicCurated[0.2ex]~\DataSynthesized[0.2ex] & Mixed & Full-Body & \Exo[0.2ex] & \Yes & Motion Generation & \href{https://arxiv.org/abs/2410.03311}{\faBook} & \href{https://beingbeyond.github.io/Being-M0/}{\faHome} \\
    HumanML3D~\cite{guo2022generating} & 2022 & \HumanSkeleton~\Text & 28.6 Hours & \NA & \DataPublicCurated[0.2ex] & \NA & Full-Body & \Exo[0.2ex] & \No & Motion Generation & \href{https://openaccess.thecvf.com/content/CVPR2022/html/Guo_Generating_Diverse_and_Natural_3D_Human_Motions_From_Text_CVPR_2022_paper.html}{\faBook} & \href{https://github.com/EricGuo5513/HumanML3D}{\faGithub} \\
    Motion-X~\cite{lin2023motion} & 2023 & \HumanVideo~\HumanSkeleton~\HumanMesh~\Text & 144.2 Hours & \NA & \DataWebCollected[0.2ex]~\DataPublicCurated[0.2ex] & Mixed & Full-Body & \Exo[0.2ex] & \No & Motion Generation & \href{https://arxiv.org/abs/2307.00818}{\faBook} & \href{https://motion-x-dataset.github.io/}{\faHome}\\
    Motion-X++~\cite{zhang2025motionxplusplus} & 2025 & \HumanVideo~\HumanSkeleton~\HumanMesh~\Text~\Audio & 19.5M Frames & \NA & \DataWebCollected[0.2ex]~\DataPublicCurated[0.2ex]~\DataSelfCaptured[0.2ex] & Mixed & Full-Body & \Exo[0.2ex] & \No & Motion Generation & \href{https://arxiv.org/abs/2501.05098}{\faBook} & \href{https://motion-x-dataset.github.io/}{\faHome} \\
    RoMo~\cite{zhang2026romo} & 2026 & \HumanSkeleton~\Text & 3K Hours & \NA & \DataWebCollected[0.2ex]~\DataPublicCurated[0.2ex] & Mixed & Full-Body & \Exo[0.2ex] & \No & Motion Generation & \href{https://arxiv.org/abs/2605.26241}{\faBook} & \href{https://davidzhang73.github.io/romo-website/}{\faHome}\\
    SnapMoGen~\cite{guo2025snapmogen} & 2025 & \HumanSkeleton~\Text & 44 Hours & 10 & \DataSelfCaptured[0.2ex] & Studio & Full-Body & \Exo[0.2ex] & \No & Motion Generation & \href{https://openreview.net/forum?id=pdE9onSn2h}{\faBook} & \href{https://snap-research.github.io/SnapMoGen/}{\faHome} \\
    ViMoGen/MBench~\cite{lin2025quest} & 2026 & \HumanVideo~\HumanSkeleton~\Text & 369.4 Hours & \NA & \DataSelfCaptured[0.2ex]~\DataWebCollected[0.2ex]~\DataSynthesized[0.2ex] & Mixed & Full-Body & \Exo[0.2ex] & \No & Motion Generation & \href{https://openreview.net/forum?id=KNke6Pkq4o}{\faBook} & \NA \\
    MRBench~\cite{liu2026mrbench} & 2026 & \HumanSkeleton~\Text & 3.39K Motions & \NA & \DataPublicCurated[0.2ex]~\DataWebCollected[0.2ex]~\DataSynthesized[0.2ex] & Mixed & Full-Body & \Exo[0.2ex] & \No & Motion-Text Retrieval & \href{https://arxiv.org/abs/2608.07993}{\faBook} & \NA \\
    BEDLAM~\cite{black2023bedlam} & 2023 & \HumanVideo~\HumanSkeleton~\HumanMesh~\HumanDepth & 380K Frames & 271 & \DataSynthesized[0.2ex] & Mixed & Full-Body & \Exo[0.2ex] & \Yes & Synthetic Human Assets & \href{https://openaccess.thecvf.com/content/CVPR2023/html/Black_BEDLAM_A_Synthetic_Dataset_of_Bodies_Exhibiting_Detailed_Lifelike_Animated_Motion_CVPR_2023_paper.html}{\faBook} & \href{https://bedlam.is.tue.mpg.de/}{\faHome} \\
    BEDLAM2.0~\cite{tesch2026bedlam2} & 2025 & \HumanVideo~\HumanSkeleton~\HumanMesh & 8.0M Images & 1615 & \DataSynthesized[0.2ex] & Mixed & Full-Body & \Exo[0.2ex] & \Yes & Synthetic Human Assets & \href{https://openreview.net/forum?id=ii9kVwf95a}{\faBook} & \href{https://bedlam2.is.tue.mpg.de/}{\faGithub} \\
    BABEL~\cite{punnakkal2021babel} & 2021 & \HumanSkeleton~\Text~\ActionControl & 43.5 Hours & 346 & \DataPublicCurated[0.2ex] & \NA & Full-Body & \Exo[0.2ex] & \No & Dense Motion Labels & \href{https://openaccess.thecvf.com/content/CVPR2021/html/Punnakkal_BABEL_Bodies_Action_and_Behavior_With_English_Labels_CVPR_2021_paper.html}{\faBook} & \href{https://babel.is.tue.mpg.de/}{\faHome}\\
    AIST++~\cite{li2021ai} & 2021 & \HumanSkeleton~\Audio & 5.2 Hours & 30 & \DataPublicCurated[0.2ex] & Studio & Full-Body & \Exo[0.2ex] & \No & Music Dance Motion & \href{https://openaccess.thecvf.com/content/ICCV2021/html/Li_AI_Choreographer_Music_Conditioned_3D_Dance_Generation_With_AIST_ICCV_2021_paper.html}{\faBook} & \href{https://google.github.io/aichoreographer/}{\faHome} \\
    TalkSHOW~\cite{yi2023generating} & 2023 & \HumanSkeleton~\HumanMesh~\Audio~\Text & 26.9 Hours & 4 & \DataWebCollected[0.2ex] & In-the-Wild & Full-Body & \Exo[0.2ex] & \No & Speech Human Motion & \href{https://openaccess.thecvf.com/content/CVPR2023/html/Yi_Generating_Holistic_3D_Human_Motion_From_Speech_CVPR_2023_paper.html}{\faBook} & \href{https://talkshow.is.tue.mpg.de/}{\faHome} \\
    FineDance~\cite{li2023finedance} & 2023 & \HumanSkeleton~\HumanMesh~\Audio & 14.6 Hours & 27 & \DataSelfCaptured[0.2ex] & Studio & Full-Body & \Exo[0.2ex] & \No & Music Dance Motion &  \href{https://openaccess.thecvf.com/content/ICCV2023/html/Li_FineDance_A_Fine-grained_Choreography_Dataset_for_3D_Full_Body_Dance_Generation_ICCV_2023_paper.html}{\faBook} & \href{https://github.com/li-ronghui/FineDance}{\faGithub} \\
    TMD~\cite{zhang2025motion} & 2025 & \HumanSkeleton~\Audio~\Text & 2.1K Samples & \NA & \DataPublicCurated[0.2ex] & \NA & Full-Body & \Exo[0.2ex] & \No & Music Dance Generation & \href{https://arxiv.org/abs/2503.06955}{\faBook} & \href{https://steve-zeyu-zhang.github.io/MotionAnything/}{\faHome} \\
    MDD~\cite{gupta2025mdd} & 2025 & \HumanSkeleton~\Audio~\Text & 10.3 Hours & 30 & \DataSelfCaptured[0.2ex] & Studio & Full-Body & \Exo[0.2ex] & \Yes & Music Dance Generation & \href{https://openaccess.thecvf.com/content/ICCV2025/html/Gupta_MDD_A_Dataset_for_Text-and-Music_Conditioned_Duet_Dance_Generation_ICCV_2025_paper.html}{\faBook} & \href{https://gprerit96.github.io/mdd-page}{\faGithub} \\
    \bottomrule
\end{tabular}
}
\end{table*}

\textbf{Human Gait Understanding.}
Human gait resources identify and characterize people from walking dynamics under changing observation conditions. CASIA-E~\cite{song2022casia} provides a comprehensive gait-recognition foundation, while LidarGait~\cite{shen2023lidargait} and Gait3D~\cite{zheng2022gait} extend analysis beyond 2D silhouettes to point clouds and dense 3D representations. Later resources address clothing changes~\cite{li2023depth}, practical evaluation~\cite{fan2023opengait}, unlabeled walking videos~\cite{fan2023learning}, cross-covariate recognition~\cite{zou2024cross}, multimodal sensing~\cite{wang2026mmgait}, and synthetic identity-consistent data~\cite{cai2026barbiegait}. However, surveillance-oriented identity protocols and controlled walking remain dominant, leaving privacy and fairness, clinical variation, and open-world gait reasoning underexplored.

\textbf{Sport Analysis.}
Sport analysis resources provide fine-grained dynamics for recognizing athletic actions, assessing motion quality, and generating feedback. FineDiving~\cite{xu2022finediving} and FineSports~\cite{xu2024finesports} support procedure-aware and hierarchical action understanding. FSBench~\cite{gao2025fsbench} and SoccerLens~\cite{elsharkawi2026soccerlens} extend evaluation to artistic sports and grounded soccer video understanding. AIFit~\cite{fieraru2021aifit} and M3GYM~\cite{xu2025m3gym} add interpretable feedback and multimodal, multi-view exercise analysis. These benchmarks enable domain-specific evaluation but depend on sport-specific rules, expert annotation, subjective scoring, and limited transfer across activities.

\textbf{Virtual Try-On.}
Virtual try-on resources assess garment or footwear transfer while preserving identity, pose, clothing structure, and realism. VITON-HD~\cite{choi2021viton} and Dress Code~\cite{morelli2022dress} establish controlled high-resolution evaluation, while Street TryOn~\cite{cui2025street} and Tstars-Tryon~\cite{chen2026tstars} extend it to in-the-wild images and diverse fashion items. ViViD~\cite{fang2024vivid}, TripVVT~\cite{shao2026tripvvt}, and ShoeFit~\cite{li2026shoefit} further introduce video consistency, triplet supervision, and footwear-specific try-on. Current resources remain limited in physical fit, occlusion handling, body and garment diversity, and temporal stability under large pose changes.

\subsubsection{Human Interaction Resources}
\label{sec::human_interaction_resources}
Human interaction resources capture human behavior within procedural, physical, environmental, and social contexts. We organize them into egocentric procedural activities, human-object interaction, human-scene interaction, and social interaction according to the setting and relational structure represented by the data.

\begin{figure}[h]
    \centering
    \includegraphics[width=\linewidth]{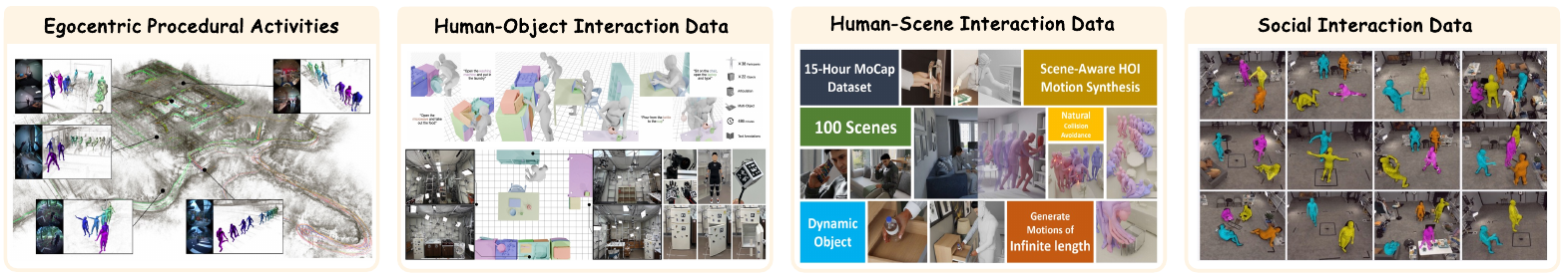}
    \caption{\textbf{Representative human interaction resources in human-centric intelligence.} From left to right, the examples show Nymeria~\cite{ma2024nymeria}, ParaHome~\cite{kim2025parahome}, 
    TRUMANS~\cite{jiang2024scaling},
    and Embody 3D~\cite{mclean2025embody}.}
\label{fig::human_interaction_resources}
\end{figure}

\textbf{Egocentric Procedural Activities.}
Egocentric procedural resources capture tasks from the actor's viewpoint for step recognition, intent anticipation, error detection, and assistance, as shown in Table~\ref{tab::summary_db_egocentric}. Ego4D~\cite{grauman2022ego4d}, EPIC-KITCHENS-100~\cite{damen2022rescaling}, and Assembly101~\cite{sener2022assembly101} establish broad procedural video supervision. AssemblyHands~\cite{ohkawa2023assemblyhands}, CaptainCook4D~\cite{peddi2024captaincook4d}, HD-EPIC~\cite{perrett2025hd}, HoloAssist~\cite{wang2023holoassist}, and Aria Digital Twin~\cite{pan2023aria} enrich it with hand, error, assistance, and dense 3D annotations. EgoExoLearn~\cite{huang2024egoexolearn} and Ego-Exo4D~\cite{grauman2024ego} align first- and third-person views, while Nymeria~\cite{ma2024nymeria} scales multimodal daily activity. EgoLife~\cite{yang2025egolife}, EgoVid-5M~\cite{wang2026egovid}, EgoSAT~\cite{lei2026egosat}, EgoPro-Bench~\cite{ran2026egopro}, ExAct~\cite{yi2026exact}, and EgoTaskQA~\cite{jia2022egotaskqa} support increasingly assistant-oriented and open-ended evaluation. These resources shift from offline recognition toward continual reasoning and proactive assistance, but remain limited by long-tail activities, causal annotation, and real-world task variation.

\textbf{Human-Object Interaction Data.}
Human-object interaction resources record how bodies and objects move and constrain each other during manipulation. First-person and category-level datasets such as H2O~\cite{kwon2021h2o}, HOI4D~\cite{liu2022hoi4d}, DexYCB~\cite{chao2021dexycb}, ARCTIC~\cite{fan2023arctic}, OAKINK2~\cite{zhan2024oakink2}, HOT3D~\cite{banerjee2025hot3d}, TACO~\cite{liu2024taco}, and GigaHands~\cite{fu2025gigahands} emphasize hand pose, object state, bimanual interaction, and compositional manipulation. BEHAVE~\cite{bhatnagar2022behave}, Full-Body Articulated Human-Object Interaction~\cite{jiang2023full}, InterCap~\cite{huang2024intercap}, HUMOTO~\cite{lu2025humoto}, ParaHome~\cite{kim2025parahome}, CORE4D~\cite{liu2025core4d}, and HOIGen-1M~\cite{liu2025hoigen} extend this coverage to full-body and multi-actor interactions. HanDyVQA~\cite{tateno2026handyvqa} and CrossHOI-Bench~\cite{lei2026crosshoi} further evaluate fine-grained interaction understanding. The field is shifting from contact recognition toward physically grounded reconstruction, generation, and reasoning, but object coverage, force and contact fidelity, and scalable 3D annotation remain limited.

\textbf{Human-Scene Interaction Data.}
Human-scene resources capture how bodies occupy and use 3D environments, grounding motion in spatial affordances. HUMANISE~\cite{wang2022humanise} links language-conditioned motion generation to 3D scenes, while dense-contact data~\cite{huang2022capturing} identifies physical body-scene contact. TRUMANS~\cite{jiang2024scaling} scales dynamic modeling to richer scenes and motions, and EgoBody~\cite{zhang2022egobody} adds egocentric captures of interacting people with shape and motion annotations. These resources shift from scene-free motion toward scene-aware synthesis and reconstruction, but remain limited by costly scanning, environment diversity, sparse affordance labels, and incomplete multi-person or object-mediated interactions.

\begin{table*}[!ht]
\rowcolors{2}{white}{table_creamyellow}
\centering
\caption{Representative datasets and benchmarks for egocentric procedural activity modeling in human-centric intelligence.
\\[0.3ex]
\textbf{$\bullet$ Modalities:} 
\HumanImage~Exocentric Image,
\HumanVideo~Exocentric Video,
\EgoVideo~Egocentric Video,
\HumanDepth~Depth Map,
\HumanSkeleton~Pose \& Skeleton,
\Gaze~Gaze Signal,
\Acceleration~Inertial Measurement Signal,
\ActionControl~Control Signal,
\Text~Text,
and \Audio~Audio.\\
\textbf{$\bullet$ Markers:} 
\DataSelfCaptured~Self-Captured,
\DataPublicCurated~Public-Curated,
\DataWebCollected~Web-Collected,
and \DataSynthesized~Synthesized.
}
\vspace{-7pt}
\label{tab::summary_db_egocentric}
\resizebox{\linewidth}{!}{
\begin{tabular}{l|c|c|c|c|c|c|c|c|c|c|c|c}
    \toprule
    \textbf{Name} & \textbf{Year} & \textbf{Modality} & \textbf{Scale} & \textbf{Subjects} & \textbf{Source} & \textbf{Scene} & \textbf{Human Region} & \textbf{View Type} & \textbf{Multi-Person} & \textbf{Scope} & \textbf{Paper} & \textbf{Web} \\
    \midrule\midrule
    AssemblyHands~\cite{ohkawa2023assemblyhands} & 2023 & \EgoVideo~\HumanImage~\HumanSkeleton & 3.03M Images & 34 & \DataPublicCurated[0.2ex] & Indoor & Hands & \Ego[0.2ex]~\Exo[0.2ex] & \No & 3D Hand Pose & \href{https://openaccess.thecvf.com/content/CVPR2023/html/Ohkawa_AssemblyHands_Towards_Egocentric_Activity_Understanding_via_3D_Hand_Pose_Estimation_CVPR_2023_paper.html}{\faBook} & \href{https://assemblyhands.github.io/}{\faHome} \\
    Assembly101~\cite{sener2022assembly101} & 2022 & \EgoVideo~\HumanVideo~\Text & 513 Hours & 53 & \DataSelfCaptured[0.2ex] & Indoor & Hands & \Ego[0.2ex]~\Exo[0.2ex] & \No & Egocentric Activity & \href{https://openaccess.thecvf.com/content/CVPR2022/html/Sener_Assembly101_A_Large-Scale_Multi-View_Video_Dataset_for_Understanding_Procedural_Activities_CVPR_2022_paper.html}{\faBook} & \href{https://assembly-101.github.io/}{\faHome} \\
    EPIC-KITCHENS-100~\cite{damen2022rescaling} & 2022 & \EgoVideo~\Text & 100 Hours & 37 &\DataSelfCaptured[0.2ex] & Kitchen & Hands & \Ego[0.2ex] & \No & Egocentric Action Benchmark & \href{https://link.springer.com/article/10.1007/s11263-021-01531-2}{\faBook} & \href{https://epic-kitchens.github.io/}{\faHome} \\
    CaptainCook4D~\cite{peddi2024captaincook4d} & 2024 & \EgoVideo~\HumanSkeleton~\HumanDepth~\Text & 94.5 Hours & 8 & \DataSelfCaptured[0.2ex] & Kitchen & Hands & \Ego[0.2ex] & \No & Procedural Error Understanding & \href{https://proceedings.neurips.cc/paper_files/paper/2024/hash/7f55e6b053cc5f7d024b111c0a13f604-Abstract-Datasets_and_Benchmarks_Track.html}{\faBook} & \href{https://rohithpeddi.github.io/captaincook}{\faHome} \\
    HD-EPIC~\cite{perrett2025hd} & 2025 & \EgoVideo~\Audio~\Text~\Gaze & 41.3 Hours & 9 & \DataSelfCaptured[0.2ex] & Kitchen & Hands & \Ego[0.2ex] & \No & Detailed Egocentric Video & \href{https://openaccess.thecvf.com/content/CVPR2025/html/Damen_HD-EPIC_A_Highly-Detailed_Egocentric_Video_Dataset_CVPR_2025_paper.html}{\faBook} & \href{http://hd-epic.github.io}{\faHome} \\
    HoloAssist~\cite{wang2023holoassist} & 2023 & \EgoVideo~\HumanSkeleton~\HumanDepth~\Audio~\Text~\Gaze~\Acceleration & 166 Hours & 222 & \DataSelfCaptured[0.2ex] & Indoor & Hands & \Ego[0.2ex] & \Yes & Interactive Assistance & \href{https://openaccess.thecvf.com/content/ICCV2023/html/Wang_HoloAssist_an_Egocentric_Human_Interaction_Dataset_for_Interactive_AI_Assistants_ICCV_2023_paper.html}{\faBook} & \href{https://holoassist.github.io/}{\faHome} \\
    Aria Digital Twin~\cite{pan2023aria} & 2023 & \EgoVideo~\HumanSkeleton~\Gaze~\HumanDepth~\Acceleration & 200 Samples & \NA & \DataSelfCaptured[0.2ex] & Indoor & Full-Body & \Ego[0.2ex] & \Yes & Egocentric 3D Perception & \href{https://openaccess.thecvf.com/content/ICCV2023/html/Pan_Aria_Digital_Twin_A_New_Benchmark_Dataset_for_Egocentric_3D_Machine_ICCV_2023_paper.html}{\faBook} & \href{https://www.projectaria.com/datasets/adt/}{\faHome} \\
    Ego4D~\cite{grauman2022ego4d} & 2022 & \EgoVideo~\Audio~\Text~\Gaze & 3.6K Hours & 931 & \DataSelfCaptured[0.2ex] & In-the-Wild & Full-Body & \Ego[0.2ex] & \Yes & Egocentric Activity & \href{https://openaccess.thecvf.com/content/CVPR2022/html/Grauman_Ego4D_Around_the_World_in_3_000_Hours_of_Egocentric_Video_CVPR_2022_paper.html}{\faBook} & \href{https://ego4d-data.org/}{\faHome} \\
    EgoExoLearn~\cite{huang2024egoexolearn} & 2024 & \EgoVideo~\HumanVideo~\Text~\Gaze & 120 Hours & \NA & \DataSelfCaptured[0.2ex] & Lab & Hands & \Ego[0.2ex]~\Exo[0.2ex] & \No & Skill Learning & \href{https://openaccess.thecvf.com/content/CVPR2024/html/Li_EgoExoLearn_A_Diverse_Large-Scale_Dataset_for_Egocentric_and_Exocentric_Video_CVPR_2024_paper.html}{\faBook} & \href{https://github.com/OpenGVLab/EgoExoLearn}{\faGithub} \\
    Ego-Exo4D~\cite{grauman2024ego} & 2024 & \EgoVideo~\HumanVideo~\HumanSkeleton~\Audio~\Text~\Gaze~\Acceleration & 1.2K Hours & 740 & \DataSelfCaptured[0.2ex] & In-the-Wild & Full-Body & \Ego[0.2ex]~\Exo[0.2ex] & \No & Skilled Activity & \href{https://openaccess.thecvf.com/content/CVPR2024/html/Grauman_Ego-Exo4D_Understanding_Skilled_Human_Activity_from_First-_and_Third-Person_Perspectives_CVPR_2024_paper.html}{\faBook} & \href{https://ego-exo4d-data.org/}{\faHome} \\
    Nymeria~\cite{ma2024nymeria} & 2024 & \EgoVideo~\HumanSkeleton~\Audio~\Text~\Gaze~\Acceleration & 300 Hours & 264 & \DataSelfCaptured[0.2ex] & In-the-Wild & Full-Body & \Ego[0.2ex]~\Exo[0.2ex] & \Yes & Daily Egocentric Motion & \href{https://arxiv.org/abs/2406.09905}{\faBook} & \href{https://www.projectaria.com/datasets/nymeria/}{\faHome} \\
    EgoLife~\cite{yang2025egolife} & 2025 & \EgoVideo~\HumanVideo~\Audio~\Text & 266 Hours & 6 & \DataSelfCaptured[0.2ex] & Indoor & Full-Body & \Ego[0.2ex]~\Exo[0.2ex] & \Yes & Long-Term Egocentric Life & \href{https://openaccess.thecvf.com/content/CVPR2025/html/Lin_EgoLife_Towards_Egocentric_Life_Assistant_CVPR_2025_paper.html}{\faBook} & \href{https://egolife-ai.github.io/}{\faHome} \\
    EgoVid-5M~\cite{wang2026egovid} & 2025 & \EgoVideo~\Text~\ActionControl & 5M Videos & \NA & \DataPublicCurated[0.2ex] & Mixed & Hands & \Ego[0.2ex] & \NA & Egocentric Video Generation & \href{https://proceedings.neurips.cc/paper_files/paper/2025/file/3ef2b740cb22dcce67c20989cb3d3fce-Paper-Datasets_and_Benchmarks_Track.pdf}{\faBook}  & \href{https://egovid.github.io}{\faHome} \\
    InterVLA~\cite{xu2025perceiving} & 2025 & \EgoVideo~\HumanVideo~\ActionControl & 11.4 Hours & 47 & \DataSelfCaptured[0.2ex] & Indoor & Full-Body & \Ego[0.2ex]~\Exo[0.2ex] & \Yes & Egocentric HOH Interaction & \href{https://openaccess.thecvf.com/content/ICCV2025/html/Xu_Perceiving_and_Acting_in_First-Person_A_Dataset_and_Benchmark_for_Egocentric_ICCV_2025_paper.html}{\faBook} & \href{https://liangxuy.github.io/InterVLA/}{\faHome} \\
    EgoProactive/Pro2Bench~\cite{kundu2026plan} & 2026 & \EgoVideo~\Text & 249.6K Clips & \NA & \DataSelfCaptured[0.2ex]~\DataPublicCurated[0.2ex] & Indoor & Hands & \Ego[0.2ex]~\Exo[0.2ex] & \NA & Proactive Egocentric Planning & \href{https://arxiv.org/abs/2606.04970}{\faBook} & \href{https://huggingface.co/datasets/facebook/wearable-ai}{\faHome} \\
    EgoExoBench~\cite{he2026egoexobench} & 2025 & \EgoVideo~\HumanVideo~\Text & 7.3K Samples & \NA & \DataPublicCurated[0.2ex] & Mixed & Full-Body & \Ego[0.2ex]~\Exo[0.2ex] & \Yes & Ego-Exo Video Understanding & \href{https://proceedings.neurips.cc/paper_files/paper/2025/hash/a974bceb4d1af7d46673d1876482bd4f-Abstract-Datasets_and_Benchmarks_Track.html}{\faBook} & \href{https://github.com/ayiyayi/EgoExoBench}{\faHome}  \\
    EgoSAT~\cite{lei2026egosat} & 2026 & \EgoVideo~\Text & 165 Hours & \NA & \DataPublicCurated[0.2ex] & In-the-Wild & Hands & \Ego[0.2ex] & \NA & Spatial Affordance & \href{https://arxiv.org/abs/2606.24422}{\faBook} & \href{https://leiyj23.github.io/EgoSAT/}{\faHome} \\
    Minerva-Ego~\cite{nagrani2026minerva} & 2026 & \EgoVideo~\Text & 1.16K Questions & 9 & \DataPublicCurated[0.2ex] & Kitchen & Hands & \Ego[0.2ex] & \No & Egocentric Spatiotemporal QA & \href{https://openaccess.thecvf.com/content/CVPR2026/html/Nagrani_Minerva-Ego_Spatiotemporal_Hints_for_Egocentric_Video_Understanding_CVPR_2026_paper.html}{\faBook} & \href{https://github.com/google-deepmind/neptune}{\faGithub} \\
    EgoPro-Bench~\cite{ran2026egopro} & 2026 & \EgoVideo~\Text & 14.3K Videos & \NA & \DataPublicCurated[0.2ex] & In-the-Wild & Hands & \Ego[0.2ex] & \NA & Proactive Egocentric Understanding & \href{https://arxiv.org/abs/2605.07299}{\faBook} & \NA \\
    ExAct~\cite{yi2026exact} & 2025 & \HumanVideo~\Text & 3.5K Pairs & \NA & \DataPublicCurated[0.2ex] & Mixed & Full-Body & \Exo[0.2ex] & \NA & Expert Action Analysis & \href{https://arxiv.org/abs/2506.06277}{\faBook} & \href{https://github.com/Texaser/Exact}{\faGithub} \\
    EgoTaskQA~\cite{jia2022egotaskqa} & 2022 & \EgoVideo~\Text & 40k Questions & \NA & \DataPublicCurated[0.2ex] & Indoor & Hands & \Ego[0.2ex] & \Yes & Egocentric Task QA & \href{https://openreview.net/forum?id=ttxAvIQA4i_}{\faBook} & \href{https://sites.google.com/view/egotaskqa}{\faHome} \\
    \bottomrule
\end{tabular}
}
\end{table*}

\textbf{Social Interaction Data.}
Social interaction resources cover communicative behavior, coupled human motion, and the evaluation of generated interactions. BEAT~\cite{liu2022beat} and CANDOR~\cite{reece2023candor} provide semantic, emotional, and naturalistic multimodal signals, while Embody 3D~\cite{mclean2025embody} and Seamless Interaction~\cite{agrawal2025seamless} scale body behavior and dyadic audiovisual motion. Hi4D~\cite{yin2023hi4d}, Harmony4D~\cite{khirodkar2024harmony4d}, Inter-X~\cite{xu2024inter}, SocialGesture~\cite{cao2025socialgesture}, EgoHumans~\cite{khirodkar2023ego}, and InterAct~\cite{ho2025interact} extend this foundation to geometric, egocentric, and expressive interactions. Generation-oriented resources further support embodied conversation and responsive digital humans through audio-to-photoreal embodiment~\cite{ng2024audio}, INFP~\cite{zhu2025infp}, SpeakerVid-5M~\cite{zhang2025speakervid}, and SentiAvatar~\cite{jin2026sentiavatar}. As generation expands from isolated individuals to interacting groups, MPIE-Bench evaluates multi-person interaction editing through 2,500 video-mined triplets organized by interaction category and contact density~\cite{lin2026mpie}. It complements semantic and perceptual evaluation by examining whether edited interactions preserve human anatomy and plausible contact geometry. These resources reflect a broader shift toward temporally coupled, physically situated, and generatively evaluated social behavior, although privacy, cultural coverage, long-horizon context, and intent annotation remain limited.

\subsubsection{Human Embodiment Resources}
\label{sec::human_embodiment_resources}
Human embodiment resources connect observed human behavior and experience with physically executable
\begin{wrapfigure}{r}{0.7\textwidth}
    \begin{minipage}{\linewidth}
        \centering
        \vspace{-5pt}
        \includegraphics[width=\linewidth]{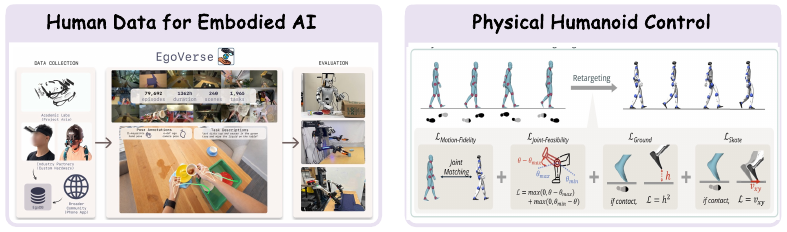}
        \caption{\textbf{Representative human embodiment resources in human-centric intelligence.} From left to right, the examples show EgoVerse~\cite{punamiya2026egoverse} and PHUMA~\cite{lee2025phuma}.}
        \label{fig::human_embodiment_resources}
        \vspace{-5pt}
    \end{minipage}
\end{wrapfigure}
action. We distinguish human data for embodied AI, which provides demonstrations and interaction experience for learning transferable agent capabilities, from physical humanoid control resources, which support the development and evaluation of physically feasible and human-like behavior.

\textbf{Human Data for Embodied AI.}
Human data for embodied AI provides experiential supervision linking human activity to robot learning, as shown in Table~\ref{tab::summary_db_embodied}. EgoVerse~\cite{punamiya2026egoverse} broadens egocentric demonstrations across everyday settings, while OpenEgo~\cite{jawaid2025openego}, EgoDex~\cite{hoque2025egodex}, In-N-On~\cite{cai2025n}, and EgoLive~\cite{li2026egolive} scale first-person dexterous manipulation and real-world task execution. ACE-Data-0 extends this coverage through synchronized ambient capture of long-horizon household activities~\cite{cao2026ace}. Resources closer to robot deployment address different transfer requirements: HRDexDB~\cite{lim2026hrdexdb} supports cross-embodiment grasping, DexH2R~\cite{wang2025dexh2r} targets dynamic handover, HandEdit~\cite{yang2026handedit} enables URDF-conditioned human-to-robot visual transfer, Hoi!~\cite{engelbracht2026hoi} provides force-grounded manipulation data, and WatchAct~\cite{li2026watchact} supports behavior-grounded evaluation. Together, these resources progress from collecting scalable human experience toward preserving the physical and embodiment-specific information required by robot policies and world models. Transfer nevertheless remains constrained by embodiment mismatch, uneven physical sensing, sparse failure cases, and limited environment diversity.

\begin{table*}[!ht]
\rowcolors{2}{white}{table_creamyellow}
\centering
\caption{Representative datasets and benchmarks for transferring human experience to embodied agents.
\\[0.3ex]
\textbf{$\bullet$ Modalities:} 
\HumanVideo~Exocentric Video,
\EgoVideo~Egocentric Video,
\HumanDepth~Depth Map,
\HumanSkeleton~Pose \& Skeleton,
\ActionControl~Control Signal,
\HumanTactile~Tactile Signal,
\Text~Text,
and \Audio~Audio.\\
\textbf{$\bullet$ Markers:} 
\DataSelfCaptured~Self-Captured,
\DataPublicCurated~Public-Curated,
\DataWebCollected~Web-Collected,
and \DataSynthesized~Synthesized.
}
\vspace{-7pt}
\label{tab::summary_db_embodied}
\resizebox{\linewidth}{!}{
\begin{tabular}{l|c|c|c|c|c|c|c|c|c|c|c|c}
    \toprule
    \textbf{Name} & \textbf{Year} & \textbf{Modality} & \textbf{Scale} & \textbf{Subjects} & \textbf{Source} & \textbf{Scene} & \textbf{Human Region} & \textbf{View Type} & \textbf{Multi-Person} & \textbf{Scope} & \textbf{Paper} & \textbf{Web} \\
    \midrule\midrule
    ACE-Data-0~\cite{cao2026ace} & 2026
    & \EgoVideo~\HumanVideo~\HumanSkeleton~\HumanTactile~\Audio~\Text
    & 150 Hours & 50
    & \DataSelfCaptured[0.2ex]
    & Indoor & Full-Body
    & \Ego[0.2ex]~\Exo[0.2ex]
    & \No
    & Embodied Interaction
    & \href{https://arxiv.org/abs/2607.28625}{\faBook}
    & \href{https://ace-data-engine.github.io/ACE-Data-0/}{\faHome} \\
    
    EgoVerse~\cite{punamiya2026egoverse} & 2026 & \EgoVideo~\HumanSkeleton~\Text~\ActionControl & 1.36K Hours & 2K & \DataSelfCaptured[0.2ex] & Indoor & Hands & \Ego[0.2ex] & \No & Egocentric Robot Learning & \href{https://arxiv.org/abs/2604.07607}{\faBook} & \href{https://github.com/GaTech-RL2/EgoVerse}{\faGithub} \\
    OpenEgo~\cite{jawaid2025openego} & 2025 & \EgoVideo~\HumanSkeleton~\Text & 1.1K Hours & 258 & \DataPublicCurated[0.2ex] & Indoor & Hands & \Ego[0.2ex] & \No & Dexterous Manipulation & \href{https://arxiv.org/abs/2509.05513}{\faBook} & \href{https://www.openegocentric.com}{\faHome} \\
    EgoDex~\cite{hoque2025egodex} & 2026 & \EgoVideo~\HumanSkeleton~\Text & 829 Hours & \NA & \DataSelfCaptured[0.2ex] & Indoor & Hands & \Ego[0.2ex] & \No & Dexterous Manipulation & \href{https://arxiv.org/abs/2505.11709}{\faBook} & \href{https://github.com/apple/ml-egodex}{\faGithub} \\
    Human0/PHSD~\cite{cai2025n} & 2025 & \EgoVideo~\HumanSkeleton~\Text~\ActionControl & 1K Hours  & \NA & \DataPublicCurated[0.2ex] & Indoor & Hands & \Ego[0.2ex] & \No & Egocentric Manipulation & \href{https://arxiv.org/abs/2511.15704}{\faBook} & \href{https://xiongyicai.github.io/In-N-On/}{\faHome} \\
    EgoLive~\cite{li2026egolive} & 2026 & \EgoVideo~\HumanSkeleton~\HumanDepth~\Text & 1.68K Hours & \NA & \DataSelfCaptured[0.2ex] & Indoor & Hands & \Ego[0.2ex] & \No & Egocentric Robot Learning & \href{https://arxiv.org/abs/2604.23570}{\faBook} & \href{https://robotdata-market.jdcloud.com/console/market}{\faHome} \\
    HRDexDB~\cite{lim2026hrdexdb} & 2026 & \EgoVideo~\HumanVideo~\HumanSkeleton~\ActionControl~\HumanTactile & 24M Frames & \NA & \DataSelfCaptured[0.2ex] & Indoor & Hands & \Ego[0.2ex]~\Exo[0.2ex] & \No & Cross-Embodiment Grasping & \href{https://arxiv.org/abs/2604.14944}{\faBook} & \href{https://github.com/snuvclab/HRDexDB}{\faGithub} \\
    HandEdit~\cite{yang2026handedit} & 2026 & \EgoVideo & 200M Instances & \NA & \DataPublicCurated[0.2ex]~\DataSynthesized[0.2ex] & Mixed & Hands & \Ego[0.2ex] & \No & Human-to-Robot Editing & \href{https://arxiv.org/abs/2608.12122}{\faBook} & \NA \\
    WatchAct~\cite{li2026watchact} & 2026 & \HumanVideo~\Text~\ActionControl & 20.1 Hours & \NA & \DataSelfCaptured[0.2ex]~\DataSynthesized[0.2ex] & Indoor & Hands & \Exo[0.2ex] & \No & Behavior-Grounded Manipulation & \href{https://arxiv.org/abs/2606.26443}{\faBook} & \href{https://github.com/Baiqi-Li/WatchAct}{\faGithub} \\
    DexH2R~\cite{wang2025dexh2r} & 2025 & \HumanVideo~\HumanSkeleton~\HumanDepth~\ActionControl & 456K Frames & 39 & \DataSelfCaptured[0.2ex] & Indoor & Hands & \Ego[0.2ex]~\Exo[0.2ex] & \No & Dexterous Grasping & \href{https://arxiv.org/abs/2506.23152}{\faBook} & \href{https://github.com/4DVLab/DexH2R}{\faGithub} \\
    Hoi!~\cite{engelbracht2026hoi} & 2026 & \EgoVideo~\HumanVideo~\HumanDepth~\ActionControl~\HumanTactile & 48 Hours & 7 & \DataSelfCaptured[0.2ex] & Indoor & Hands & \Ego[0.2ex]~\Exo[0.2ex] & \No & Articulated Manipulation & \href{https://arxiv.org/abs/2512.04884}{\faBook} & \href{https://github.com/timengelbracht/hoi-dataset-tools}{\faGithub} \\
    \bottomrule
\end{tabular}
}
\end{table*}

\textbf{Physical Humanoid Control.}
Physical humanoid control resources evaluate whether motion is dynamically feasible and human-like, not merely visually or kinematically plausible. PHUMA~\cite{lee2025phuma} provides physically grounded locomotion data. HumanTracker broadens evaluation to contact-rich, long-horizon whole-body tracking and includes pairwise human preferences for failures overlooked by frame-wise kinematic metrics~\cite{liu2026humantracker}. The Motion Turing Test~\cite{li2026towardsmotion} complements these resources by assessing whether robot motion appears human-like through behavior-oriented evaluation. Together, they shift evaluation from motion reproduction toward physical reliability and human-aligned behavior. Coverage remains sparse, with limited transfer across robot morphologies and sim-to-real settings; manipulation, perturbation recovery, and safety-aware evaluation remain underdeveloped.

\subsection{Metrics}
\label{sec::metrics}

Human-centric methods produce heterogeneous outputs, ranging from geometric structures and semantic predictions to generated content, physical interactions, and executable behavior, making their evaluation difficult to organize by task alone. We therefore classify representative metrics according to their primary evaluation targets into five families: reconstruction-centric, semantic-centric, generative-centric, interaction-centric, and efficiency-centric metrics. Table~\ref{tab::summary_metric_notation} defines the shared notation and optimization directions, while Table~\ref{tab::summary_metric_formulas_all} summarizes their formulations and interpretations.


\begin{table*}[!h]
\rowcolors{2}{white}{table_creamyellow}
\centering
\caption{Unified notation for the evaluation metrics summarized in Table~\ref{tab::summary_metric_formulas_all}.}
\begingroup
\renewcommand{\arraystretch}{0.90}
\resizebox{\linewidth}{!}{
\begin{tabular}{l|l}
    \toprule
    \textbf{Symbol} & \textbf{Meaning} \\
    \midrule\midrule
    $\hat{\cdot}$ & Model prediction; the corresponding unhatted variable denotes the reference. \\
    $N,T,J,V$ & Numbers of evaluation units, frames, joints, and vertices. \\
    $C$ & Number of classes or conditioning groups, depending on the metric. \\
    $K_r,K_n,K_c$ & Retrieval cutoff, maximum n-gram order, and samples per condition. \\
    $L_i,L$ & Numbers of reference captions for sample $i$ and mel-cepstral coefficients. \\
    $\mathcal{R}_i,R_i$ & Relevant-item set for query $i$ and its cardinality. \\
    $G_i$ & Set of relevant gallery items for query $i$. \\
    $\mathcal{U}$ & Union of generated and reference sample sets. \\
    $\mathrm{NN}_{-}(\mathbf{z})$ & Nearest neighbor of $\mathbf{z}$ excluding $\mathbf{z}$ itself. \\
    $D_{\mathrm{real}}$ & Diversity measured on the reference data distribution. \\
    $\mathcal{B}_a,\mathcal{B}_m$ & Sets of detected audio and motion beat times. \\
    $\sigma$ & Temporal bandwidth used by beat-alignment metrics. \\
    $\mathcal{D},d_{\delta}$ & Candidate temporal offsets and feature distance at offset $\delta$. \\
    $\pi$ & Dynamic-time-warping alignment path. \\
    $P_i$ & Number of evaluated points in sample $i$. \\
    $\mathbf{q}_t$ & Trajectory position at frame $t$. \\
    $\mathrm{FAR},\mathrm{FRR}$ & False-acceptance and false-rejection rates. \\
    $w_n$ & Weight assigned to the $n$-gram order. \\
    $\mathbf{p}_{t,j}$ & 3D coordinate of joint $j$ at frame $t$. \\
    $\mathbf{v}_{i}$ & Coordinate of mesh vertex $i$. \\
    $\mathbf{x},\mathbf{y}$ & Generic dense signal, point, image, or feature vector. \\
    $X,Y$ & Predicted and reference point sets. \\
    $\hat{M},M$ & Predicted and reference masks or regions. \\
    $\hat{\mathcal{Z}},\mathcal{Z}$ & Generated and reference sample sets. \\
    $\phi(\cdot)$ & Evaluator feature extractor; subscripts specify the feature space. \\
    $\mathbf{A}\in\mathrm{Sim}(3)$ & Similarity transform containing scale, rotation, and translation. \\
    $\Delta,\Delta^2$ & First- and second-order temporal differences. \\
    $\tau$ & Evaluation threshold or tolerance. \\
    $\mathrm{TP},\mathrm{FP},\mathrm{FN}$ & True positives, false positives, and false negatives. \\
    $P,R,F_1$ & Precision, recall, and F1 score. \\
    $C_{\mathrm{tp}},C_{\mathrm{pred}},C_{\mathrm{gt}}$ & True-positive, predicted, and ground-truth contacts. \\
    $\mathrm{SDF}(\cdot)$ & Signed distance to an object or scene surface. \\
    $\mu,\Sigma$ & Mean and covariance of evaluator features. \\
    $\mathbf{a},\mathbf{v}$ & Audio and visual streams. \\
    $\mathbf{g},r_t,\gamma$ & Target state, reward, and discount factor. \\
    $|\theta|,\mathrm{FLOPs},\mathrm{Mem}$ & Parameter count, operation count, and memory footprint. \\
    \midrule
    \textcolor{crReal}{$\uparrow$} & Larger values indicate better performance. \\
    \textcolor{crEstimated}{$\downarrow$} & Smaller values indicate better performance. \\
    $\rightarrow$ & Values closer to the target are preferred. \\
    \bottomrule
\end{tabular}}
\endgroup
\label{tab::summary_metric_notation}
\end{table*}

\subsubsection{Reconstruction-Centric Metrics}
\label{sec::reconstruction_centric_metrics}
Reconstruction-centric metrics assess the agreement between predicted outputs and corresponding reference observations. We organize them into articulated human state, surface geometry, and image or rendering fidelity, progressing from sparse body structures to dense three-dimensional surfaces and photometric appearance.

\textbf{Articulated Human State.}
Articulated human state metrics quantify geometric accuracy in joint and mesh recovery. MPJPE averages Euclidean errors between predicted and ground-truth 3D joints and is standard for pose and motion reconstruction~\cite{yang2026sam,cai2023smpler,wang2025prompthmr}. PA-MPJPE applies Procrustes alignment first, isolating pose-configuration error from global scale, rotation, and translation~\cite{cai2023smpler,pavlakos2024reconstructing,zhu2023motionbert}. PVE/MPVPE extends this measure to dense mesh vertices, better capturing shape and surface-deformation errors~\cite{yang2026sam,cai2023smpler,pavlakos2024reconstructing,li2025genmo}. PCK reports the percentage of keypoints within a distance threshold~\cite{yang2026sam,wu2026pear,pavlakos2024reconstructing}, while AUC summarizes PCK across tolerances~\cite{pavlakos2024reconstructing}. F-score combines precision and recall under a distance threshold for tolerance-based surface or keypoint evaluation~\cite{yang2026sam,pavlakos2024reconstructing,chen2025foundhand}.

\textbf{Surface Geometry.}
Surface geometry metrics assess dense 3D structure beyond sparse articulated states. Chamfer Distance uses bidirectional nearest-neighbor distances between predicted and ground-truth point sets and is common in avatar, object, and surface reconstruction~\cite{hu2026humannova,liu2025easyhoi}. Surface F-score reports precision and recall within a geometric tolerance, making thresholded reconstruction quality easier to interpret~\cite{hu2026humannova,liu2025easyhoi,martinelli2026volhume}. Point-to-surface error measures the distance from predicted points to the target surface for point-based or volumetric geometry~\cite{martinelli2026volhume,li2026mvhumannet++}. Normal angular error evaluates local surface orientation, complementing positional errors~\cite{khirodkarsapiens2,saleh2025david,wang2026thfm}. Depth RMSE and AbsRel quantify absolute and relative depth errors for scene reconstruction and egocentric prediction~\cite{saleh2025david,chen2025human3r,li2026unish}. Object/camera pose error measures translation and rotation consistency, linking human geometry to scene and world-coordinate reconstruction~\cite{chen2025human3r,li2026unish,hao2026egosim}.

\textbf{Image and Rendering Fidelity.}
Image and rendering metrics compare synthesized frames, avatar renderings, or world-model rollouts with references. PSNR derives from pixel MSE and measures low-level fidelity but poorly reflects perceptual realism~\cite{hu2026humannova,zhuang2025idol,youwang2026fica}. SSIM evaluates luminance, contrast, and structure, capturing structural preservation in rendering and video reconstruction~\cite{hu2026humannova,gan2025humandit,zuo2025dreamvvt}. LPIPS measures learned perceptual-feature distance beyond pixel-wise error~\cite{zhuang2025idol,youwang2026fica,hao2026egosim}. L1/L2/MSE/MAE/RMSE directly quantify dense signal errors when targets have well-defined numeric scales~\cite{zhuang2025idol,gan2025humandit,kocasari2026face}. DreamSim and related distances compare perceptual embeddings for world prediction and egocentric synthesis~\cite{kim2026dexterous,bai2026whole,pallotta2026egocontrol}. Matting and warping metrics evaluate alpha boundaries and temporal correspondence for foreground extraction, editing, and video reconstruction~\cite{saleh2025david,wang2026thfm,kocasari2026face}.

\subsubsection{Semantic-Centric Metrics}
\label{sec::semantic_centric_metrics}
Semantic-centric metrics evaluate whether a model preserves task-relevant, identity-related, and cross-modal meaning. We organize them into discriminative recognition, retrieval and verification, and cross-modal semantic alignment, respectively covering direct prediction, relational matching, and semantic consistency across modalities.

\textbf{Discriminative Recognition.}
Discriminative metrics assess human recognition, localization, parsing, and understanding. Accuracy and Top-k accuracy measure closed-set correctness across classification and VQA tasks~\cite{wang2025hulk,tang2023humanbench,kim2025sapiensid,narayan2025facexformer}. Precision, recall, and F1 balance false positives and false negatives while exposing class imbalance~\cite{zuo2024plip,narayan2025facexformer,wang2025hulk}. AP and mAP integrate precision-recall behavior over confidence thresholds for detection, ReID, person search, and segmentation~\cite{khirodkar2024sapiens,khirodkarsapiens2,wang2025hulk,tang2023humanbench,kim2025sapiensid}. IoU and mIoU measure region overlap for parsing, segmentation, and grounding~\cite{khirodkar2024sapiens,khirodkarsapiens2,wang2025hulk,tang2023humanbench}. AUC and HTER summarize threshold-dependent verification, especially in security-oriented evaluation~\cite{wang2025fsfm}.

\textbf{Retrieval Verification.}
Retrieval and verification metrics assess ranking and matching across identities, modalities, or language conditions. Rank-k, CMC, and mINP measure top-ranked identity retrieval and coverage of positive matches~\cite{niu2025chatreid,he2025instruct,zuo2026reid5o}. R-Precision and Recall@K apply the same logic to cross-modal retrieval by testing whether matched samples appear near the top~\cite{jiang2023motiongpt,wu2025mg,wang2026motiongpt}, with MRBench extending this evaluation across heterogeneous motions and description granularities~\cite{liu2026mrbench}. MM-Dist or Matching Distance quantifies paired-embedding distance, complementing ranking metrics with the compactness of cross-modal alignment~\cite{jiang2023motiongpt,wu2025mg,wang2026motiongpt}.

\textbf{Cross-Modal Semantic Alignment.}
Cross-modal semantic metrics test whether human outputs follow multimodal conditions. CLIP-family scores measure similarity in pretrained vision-language spaces for text-conditioned image, video, avatar, and world generation~\cite{li2024cosmicman,wu2026genlca,yang2025sigman,tu2025playerone,shen2026egoforge}. DINO-based similarities compare visual semantics when pixel-level comparison is inadequate~\cite{wang2026hand2world,tu2025playerone,shen2026egoforge}. BLEU, ROUGE, METEOR, CIDEr, and BERTScore assess generated language, including captions, action descriptions, and reasoning responses~\cite{wang2025hsi,wang2026hsi,wu2025mg}.

\subsubsection{Generative-Centric Metrics}
\label{sec::generative_centric_metrics}
Generative outputs require evaluation along multiple complementary dimensions because visual or distributional realism alone cannot capture their overall quality. We therefore consider fidelity quality, diversity and coverage, temporal dynamics, audio-visual synchronization, and identity and appearance consistency.

\textbf{Fidelity Quality.}
Distributional fidelity metrics assess whether generated human samples match the real data distribution. FID compares Gaussian statistics of real and generated features and is widely used for human image, avatar, motion, and try-on generation~\cite{lin2025omnihuman,gan2025omniavatar,lee2025voost,zhang2024motiondiffuse}. FVD extends this comparison to video features, capturing appearance and temporal realism~\cite{lin2025omnihuman,chen2025hunyuanvideo,jiang2025omnihuman,gan2025humandit}. KID estimates maximum mean discrepancy with a polynomial kernel and is useful when sample size or estimator bias matters~\cite{lee2025voost,morelli2022dress}. FGD applies Frechet-style comparison to motion or gesture features~\cite{chen2025language,mughal2026miburi,liu2022beat}. Inception Score measures classifier confidence and diversity but is less diagnostic when the classifier domain does not reflect human realism~\cite{morelli2022dress}.

\textbf{Diversity Coverage.}
Diversity and coverage metrics detect mode collapse when multiple outputs can satisfy the same condition. Diversity averages pairwise distances among generated samples or features to measure global spread~\cite{zhang2024motiondiffuse,zhang2023generating,zhang2024motiongpt,wang2026motionbricks}. MultiModality measures variation among samples generated from the same condition, capturing conditional diversity~\cite{zhang2024motiondiffuse,zhang2023generating,wang2026motiongpt,wang2026hsi}. Coverage, minimum matching distance, and 1-NNA use nearest-neighbor relationships to assess reference coverage and distributional separability~\cite{petrov2025tridi}. Maximum mean discrepancy compares generated and real distributions in a kernel feature space~\cite{wang2026motionbricks}.

\textbf{Temporal Dynamics.}
Temporal dynamics metrics assess motion coherence beyond frame-wise fidelity. Velocity, acceleration, and temporal joint errors compare trajectory derivatives between predictions and ground truth for pose, mesh, and motion reconstruction~\cite{wang2025prompthmr,zhu2023motionbert,li2025genmo}. Jerk and smoothness metrics penalize high-frequency artifacts and unstable transitions in generated motion and video~\cite{wang2026motionbricks,fan2025go,liu2024programmable,shi2024interactive}. Foot-skating and floating diagnostics target contact failures by testing foot stability and adherence to support surfaces~\cite{wang2025prompthmr,rempe2026kimodo,wang2026motionbricks,patel2025uniegomotion}.

\textbf{Audio-Visual Synchronization.}
Audio-visual synchronization metrics assess temporal alignment in speech-, music-, and interaction-driven human animation. Sync-C, Sync-D, and SyncNet-style metrics compare audio and mouth-motion features to measure lip synchronization and confidence~\cite{bao2026archon,gan2025omniavatar,chen2025hunyuanvideo,jiang2025omnihuman,zhang2026soul}. MCD-DTW measures spectral distortion between aligned generated and reference speech when audio is synthesized with motion~\cite{bao2026archon}. Beat, audio-motion, and audio/music alignment scores evaluate whether body movement follows rhythm, prosody, or partner cues~\cite{mughal2026miburi,liu2022beat,zhang2026soul}.

\textbf{Identity and Appearance Consistency.}
Identity and appearance metrics assess whether generation preserves the intended person and clothing across viewpoints, poses, and time. Identity embedding similarity uses cosine similarity or distance from ArcFace-like recognizers for avatar generation and human animation~\cite{qiu2025lhm,youwang2026fica,kant2025pippo,liang2026avatar}. Face/body consistency compares regional or temporal features, detecting identity drift and shape changes missed by global image metrics~\cite{kant2025pippo,zhou2025realisdance,li2025openhumanvid}. Garment and try-on consistency measures clothing appearance and subject-garment compatibility after synthesis or transfer~\cite{lee2025voost,song2026fashionchameleon,shao2026tripvvt}.

\subsubsection{Interaction-Centric Metrics}
\label{sec::interaction_centric_metrics}
Interaction-centric metrics assess whether predicted or generated behavior remains valid within relational, physical, and task-oriented contexts. We organize them into contact and collision, physical plausibility, embodied task performance, and human preference and subjective quality, progressing from local geometric validity to physical execution, task completion, and perceived naturalness.

\textbf{Contact and Collision.}
Contact and collision metrics assess the physical validity of human-object-scene relations beyond visual realism. Contact precision, recall, and F1 compare predicted and ground-truth contact events or regions for interaction and contact-aware 4D reconstruction~\cite{cai2026vihoi,ym2026graft,sur2026unicon3r,wang2025hsi,wang2026hsi}. Contact distance measures geometric localization error between predicted and annotated contact regions~\cite{li2025hoi,wang2026arthoi,wu2026human,lu2025humoto}. Penetration, collision, and intersection metrics quantify invalid overlap through measures such as penetration depth or intersection volume~\cite{li2025hoi,sur2026unicon3r,wang2025hsi,wang2026hsi}.

\textbf{Physical Plausibility.}
Physical plausibility metrics complement contact and collision scores by testing whether motion and manipulation obey task dynamics or simulator constraints. Force- and physics-based scores assess dynamic stability, force consistency, and constraint satisfaction in motion generation, human-object interaction synthesis, and embodied control~\cite{fan2025go,zhang2026openhoi,lim2026hrdexdb,engelbracht2026hoi}.

\textbf{Embodied Task Performance.}
Task and control metrics assess downstream planning, manipulation, and execution. Success and completion rates report whether trials, instructions, or subgoals are completed, providing direct measures for embodied agents~\cite{chen2025egoagent,gao2026dreamdojo,shi2026gpc,ma2026humanscale,chen2025moto,bjorck2025gr00t}. Return, reward, and task score aggregate progress under graded objectives~\cite{shi2026gpc,wang2026hsi,lin2026activemimic,zheng2026egoscale}. Goal and planning errors quantify deviations from target states or action sequences~\cite{shi2026gpc,lionar2026teamhoi,goswami2026worldmodelslearningdexterous,lin2026activemimic}. Robot-suite metrics aggregate performance across standardized tasks for comparing generalist policies and embodied foundation models~\cite{chen2025moto,luo2026being,liu2026lara}.

\textbf{Human Preference and Subjective Quality.}
Human preference metrics assess qualities that automatic scores may miss. Preference and win rates compare methods through pairwise or multiway judgments, including humanoid tracking rollouts whose physical failures may be overlooked by frame-wise errors~\cite{lin2025omnihuman,wang2026motionbricks,cai2026vihoi,mughal2026miburi,liu2026humantracker}. Likert, MOS, and rating protocols quantify perceived quality, alignment, or controllability on ordinal scales, providing more diagnostic feedback than binary preferences~\cite{liang2026avatar,xue2025infinihuman,li2025hoi}. Naturalness, realism, and Motion Turing-style tests assess whether generated motion, avatars, or humanoid behavior appear human-like to observers~\cite{cai2026vihoi,mughal2026miburi,li2026towardsmotion}.

\subsubsection{Efficiency-Centric Metrics}
\label{sec::efficiency_centric_metrics}
Efficiency-centric metrics assess whether human-centric models can operate under practical computational and deployment constraints. We focus on runtime and resource cost, covering inference speed, latency, throughput, model size, memory consumption, and computational complexity.

\textbf{Runtime and Resource Cost.}
Runtime and resource metrics assess whether systems can scale beyond offline evaluation. FPS, latency, and runtime measure processing speed for real-time reconstruction, motion generation, and streaming avatars~\cite{wang2026motionbricks,song2026interactiveavatar,qiu2025lhm,sur2026unicon3r}. Memory, parameter count, and FLOPs characterize model footprint and computation, relating performance to hardware-limited deployability~\cite{qiu2025lhm,kocasari2026face,lu2025scamo,wang2026superman}. Throughput and frequency measure processed samples, frames, or control steps per unit time for scalable pipelines and embodied control~\cite{zhang2026soul,zeng2025activeumi}.

\begingroup
\setlength{\tabcolsep}{1.5pt}
\renewcommand{\arraystretch}{1.18}
\rowcolors{2}{white}{table_creamyellow}
{\fontsize{8}{9.6}\selectfont
\begin{longtable}{>{\raggedright\arraybackslash}m{2.1cm}|>{\centering\arraybackslash}m{0.8cm}|>{\centering\arraybackslash}m{8.0cm}|>{\raggedright\arraybackslash}m{5.2cm}}
\caption{Summary of representative evaluation metrics for human-centric intelligence.}
\label{tab::summary_metric_formulas_all}\\
\toprule
\textbf{Metric} & \textbf{Better} & \textbf{Formula} & \textbf{Description} \\
\midrule\midrule
\endfirsthead

\multicolumn{4}{l}{\tablename~\thetable\ (continued)}\\
\toprule
\textbf{Metric} & \textbf{Better} & \textbf{Formula} & \textbf{Description} \\
\midrule\midrule
\endhead

\midrule
\multicolumn{4}{r}{\textit{Continued on next page}}\\
\endfoot

\bottomrule
\endlastfoot

\rowcolor{gray!10} \multicolumn{4}{l}{\textit{\textbf{Articulated Human State}}}\\
    \midrule
    MPJPE & \textcolor{crEstimated}{$\downarrow$} & $\frac{1}{J}\sum_{j=1}^{J}\|\hat{\mathbf{p}}_j-\mathbf{p}_j\|_2$ & Mean 3D joint-position error. \\
    PA-MPJPE & \textcolor{crEstimated}{$\downarrow$} & $\min_{\mathbf{A}\in\mathrm{Sim}(3)}\frac{1}{J}\sum_{j=1}^{J}\|\mathbf{A}\hat{\mathbf{p}}_j-\mathbf{p}_j\|_2$ & Joint error after similarity alignment. \\
    MPVPE & \textcolor{crEstimated}{$\downarrow$} & $\frac{1}{V}\sum_{i=1}^{V}\|\hat{\mathbf{v}}_i-\mathbf{v}_i\|_2$ & Mean dense vertex-position error. \\
    PCK & \textcolor{crReal}{$\uparrow$} & $\frac{1}{J}\sum_{j=1}^{J}\mathbf{1}[\|\hat{\mathbf{p}}_j-\mathbf{p}_j\|_2<\tau]$ & Fraction of keypoints within threshold $\tau$. \\
    AUC-PCK & \textcolor{crReal}{$\uparrow$} & $\frac{1}{\tau_{\max}-\tau_{\min}}\int_{\tau_{\min}}^{\tau_{\max}}\mathrm{PCK}(\tau)d\tau$ & Area under the PCK-threshold curve. \\
    F-score@$\tau$ & \textcolor{crReal}{$\uparrow$} & $\frac{2P_{\tau}R_{\tau}}{P_{\tau}+R_{\tau}}$ & Harmonic mean of thresholded precision and recall. \\
    Velocity Error & \textcolor{crEstimated}{$\downarrow$} & $\frac{1}{(T-1)J}\sum_{t=2}^{T}\sum_{j=1}^{J}\|\Delta\hat{\mathbf{p}}_{t,j}-\Delta\mathbf{p}_{t,j}\|_2$ & Error in first-order motion dynamics. \\
    Acceleration Error & \textcolor{crEstimated}{$\downarrow$} & $\frac{1}{(T-2)J}\sum_{t=3}^{T}\sum_{j=1}^{J}\|\Delta^2\hat{\mathbf{p}}_{t,j}-\Delta^2\mathbf{p}_{t,j}\|_2$ & Error in second-order motion dynamics. \\

    \midrule
    \rowcolor{gray!10} \multicolumn{4}{l}{\textit{\textbf{Surface Geometry}}}\\
    \midrule
    Chamfer Distance & \textcolor{crEstimated}{$\downarrow$} & $\frac{1}{|X|}\sum_{\mathbf{x}\in X}\min_{\mathbf{y}\in Y}\|\mathbf{x}-\mathbf{y}\|_2^2+\frac{1}{|Y|}\sum_{\mathbf{y}\in Y}\min_{\mathbf{x}\in X}\|\mathbf{y}-\mathbf{x}\|_2^2$ & Bidirectional distance between point sets. \\
    Point-to-Surface & \textcolor{crEstimated}{$\downarrow$} & $\frac{1}{N}\sum_{i=1}^{N}d(\hat{\mathbf{x}}_i,S)$ & Distance from predicted points to a target surface. \\
    Normal Error & \textcolor{crEstimated}{$\downarrow$} & $\frac{1}{N}\sum_{i=1}^{N}\arccos(\hat{\mathbf{n}}_i^\top\mathbf{n}_i)$ & Mean angular surface-orientation error. \\
    Depth RMSE & \textcolor{crEstimated}{$\downarrow$} & $\sqrt{\frac{1}{N}\sum_{i=1}^{N}(\hat{d}_i-d_i)^2}$ & Root mean squared depth error. \\
    AbsRel & \textcolor{crEstimated}{$\downarrow$} & $\frac{1}{N}\sum_{i=1}^{N}\frac{|\hat{d}_i-d_i|}{d_i}$ & Relative absolute depth error. \\
    Translation Error & \textcolor{crEstimated}{$\downarrow$} & $\|\hat{\mathbf{t}}-\mathbf{t}\|_2$ & Camera or object translation error. \\
    Rotation Error & \textcolor{crEstimated}{$\downarrow$} & $\arccos\frac{\mathrm{tr}(\hat{\mathbf{R}}^\top\mathbf{R})-1}{2}$ & Camera or object rotation error. \\

    \midrule
    \rowcolor{gray!10} \multicolumn{4}{l}{\textit{\textbf{Image and Rendering Fidelity}}}\\
    \midrule
    PSNR & \textcolor{crReal}{$\uparrow$} & $10\log_{10}\frac{\mathrm{MAX}^2}{\mathrm{MSE}}$ & Pixel-level reconstruction fidelity. \\
    SSIM & \textcolor{crReal}{$\uparrow$} & $\frac{(2\mu_{\hat{x}}\mu_x+C_1)(2\sigma_{\hat{x}x}+C_2)}{(\mu_{\hat{x}}^2+\mu_x^2+C_1)(\sigma_{\hat{x}}^2+\sigma_x^2+C_2)}$ & Structural similarity of local image statistics. \\
    LPIPS & \textcolor{crEstimated}{$\downarrow$} & $\sum_{\ell}w_{\ell}\|\phi_{\ell}(\hat{\mathbf{x}})-\phi_{\ell}(\mathbf{x})\|_2^2$ & Learned perceptual feature distance. \\
    DreamSim & \textcolor{crEstimated}{$\downarrow$} & $D_{\mathrm{DreamSim}}(\hat{\mathbf{x}},\mathbf{x})$ & Human-aligned perceptual image distance. \\
    L1 Error & \textcolor{crEstimated}{$\downarrow$} & $\frac{1}{N}\sum_{i=1}^{N}|\hat{x}_i-x_i|$ & Mean absolute dense signal error. \\
    L2 Error & \textcolor{crEstimated}{$\downarrow$} & $\sqrt{\sum_{i=1}^{N}(\hat{x}_i-x_i)^2}$ & Euclidean dense signal error. \\
    MSE & \textcolor{crEstimated}{$\downarrow$} & $\frac{1}{N}\sum_{i=1}^{N}(\hat{x}_i-x_i)^2$ & Mean squared dense signal error. \\
    RMSE & \textcolor{crEstimated}{$\downarrow$} & $\sqrt{\frac{1}{N}\sum_{i=1}^{N}(\hat{x}_i-x_i)^2}$ & Root mean squared dense signal error. \\
    Alpha Error & \textcolor{crEstimated}{$\downarrow$} & $\frac{1}{N}\sum_{i=1}^{N}|\hat{\alpha}_i-\alpha_i|$ & Foreground opacity estimation error. \\
    Warping Error & \textcolor{crEstimated}{$\downarrow$} & $\frac{1}{N}\sum_{i=1}^{N}\|\hat{\mathbf{x}}_{i\rightarrow t}-\mathbf{x}_{i,t}\|_1$ & Temporal reprojection or warp consistency error. \\

    \midrule
    \rowcolor{gray!10} \multicolumn{4}{l}{\textit{\textbf{Discriminative Recognition and Retrieval Verification}}}\\
    \midrule
    Accuracy & \textcolor{crReal}{$\uparrow$} & $\frac{1}{N}\sum_{i=1}^{N}\mathbf{1}[\hat{y}_i=y_i]$ & Fraction of correct predictions. \\
    Top-$k$ Accuracy & \textcolor{crReal}{$\uparrow$} & $\frac{1}{N}\sum_{i=1}^{N}\mathbf{1}[y_i\in\mathrm{Top}k(\hat{\mathbf{s}}_i)]$ & Correctness within top-ranked candidates. \\
    Precision & \textcolor{crReal}{$\uparrow$} & $\frac{\mathrm{TP}}{\mathrm{TP}+\mathrm{FP}}$ & Reliability of positive predictions. \\
    Recall & \textcolor{crReal}{$\uparrow$} & $\frac{\mathrm{TP}}{\mathrm{TP}+\mathrm{FN}}$ & Coverage of true positive instances. \\
    F1 Score & \textcolor{crReal}{$\uparrow$} & $\frac{2PR}{P+R}$ & Harmonic mean of precision and recall. \\
    AP & \textcolor{crReal}{$\uparrow$} & $\int_0^1P(r)dr$ & Area under the precision-recall curve. \\
    mAP & \textcolor{crReal}{$\uparrow$} & $\frac{1}{C}\sum_{c=1}^{C}\mathrm{AP}_c$ & Mean AP over classes, identities, or queries. \\
    IoU & \textcolor{crReal}{$\uparrow$} & $\frac{|\hat{M}\cap M|}{|\hat{M}\cup M|}$ & Overlap between predicted and reference regions. \\
    mIoU & \textcolor{crReal}{$\uparrow$} & $\frac{1}{C}\sum_{c=1}^{C}\frac{|\hat{M}_c\cap M_c|}{|\hat{M}_c\cup M_c|}$ & Mean region overlap across classes. \\
    Rank-$k$ & \textcolor{crReal}{$\uparrow$} & $\frac{1}{N}\sum_{i=1}^{N}\mathbf{1}[y_i\in\mathrm{Top}k(\hat{\mathbf{s}}_i)]$ & Retrieval success within top-$k$ results. \\
    CMC & \textcolor{crReal}{$\uparrow$} & $\Pr[\mathrm{rank}(y_i)\leq k]$ & Cumulative match probability at rank $k$. \\
    mINP & \textcolor{crReal}{$\uparrow$} & $\frac{1}{N}\sum_{i=1}^{N}\frac{|G_i|}{\max_{g\in G_i}\mathrm{rank}_i(g)}$ & Retrieval quality for hard positive matches. \\
    ROC-AUC & \textcolor{crReal}{$\uparrow$} & $\int\mathrm{TPR}(\mathrm{FPR})d\mathrm{FPR}$ & Threshold-aggregated verification quality. \\
    HTER & \textcolor{crEstimated}{$\downarrow$} & $\frac{\mathrm{FAR}+\mathrm{FRR}}{2}$ & Average false-acceptance and false-rejection rate. \\

    \midrule
    \rowcolor{gray!10} \multicolumn{4}{l}{\textit{\textbf{Cross-Modal Semantic Alignment}}}\\
    \midrule
    R-Precision & \textcolor{crReal}{$\uparrow$} & $\frac{1}{N}\sum_{i=1}^{N}\frac{|\mathrm{Top}_{R_i}(i)\cap\mathcal{R}_i|}{R_i},\quad R_i=|\mathcal{R}_i|$ & Precision at the number of relevant items for each query. \\
    Recall@$K_r$ & \textcolor{crReal}{$\uparrow$} & $\frac{1}{N}\sum_{i=1}^{N}\frac{|\mathrm{Top}_{K_r}(i)\cap\mathcal{R}_i|}{|\mathcal{R}_i|}$ & Fraction of relevant items retrieved within the top-$K_r$ results. \\
    MM-Dist & \textcolor{crEstimated}{$\downarrow$} & $\frac{1}{N}\sum_{i=1}^{N}\|\phi_m(\hat{\mathbf{m}}_i)-\phi_t(\mathbf{t}_i)\|_2$ & Distance between generated motion and text. \\
    CLIP Similarity & \textcolor{crReal}{$\uparrow$} & $\frac{\phi_{\mathrm{CLIP}}(a)^\top\phi_{\mathrm{CLIP}}(b)}{\|\phi_{\mathrm{CLIP}}(a)\|_2\|\phi_{\mathrm{CLIP}}(b)\|_2}$ & Semantic agreement in CLIP space. \\
    DINO Similarity & \textcolor{crReal}{$\uparrow$} & $\frac{\phi_{\mathrm{DINO}}(a)^\top\phi_{\mathrm{DINO}}(b)}{\|\phi_{\mathrm{DINO}}(a)\|_2\|\phi_{\mathrm{DINO}}(b)\|_2}$ & Visual agreement in self-supervised feature space. \\
    BLEU & \textcolor{crReal}{$\uparrow$} & $\mathrm{BP}\exp\left(\sum_{n=1}^{K_n}w_n\log p_n\right)$ & N-gram precision for generated text. \\
    ROUGE-L & \textcolor{crReal}{$\uparrow$} & $\frac{(1+\beta^2)R_{\mathrm{LCS}}P_{\mathrm{LCS}}}{R_{\mathrm{LCS}}+\beta^2P_{\mathrm{LCS}}}$ & Longest-common-subsequence text overlap. \\
    CIDEr & \textcolor{crReal}{$\uparrow$} & $\sum_{n=1}^{K_n}w_n\frac{1}{L_i}\sum_{j=1}^{L_i}\frac{\mathbf{g}_n(\hat{\mathbf{y}}_i)^\top\mathbf{g}_n(\mathbf{y}_{i,j})}{\|\mathbf{g}_n(\hat{\mathbf{y}}_i)\|_2\|\mathbf{g}_n(\mathbf{y}_{i,j})\|_2}$ & Average TF-IDF weighted cosine similarity to reference captions. \\

    \midrule
    \rowcolor{gray!10} \multicolumn{4}{l}{\textit{\textbf{Fidelity Quality and Diversity Coverage}}}\\
    \midrule
    FID & \textcolor{crEstimated}{$\downarrow$} & $\|\mu_r-\mu_g\|_2^2+\mathrm{Tr}(\Sigma_r+\Sigma_g-2(\Sigma_r\Sigma_g)^{1/2})$ & Feature-distribution gap for generated samples. \\
    KID & \textcolor{crEstimated}{$\downarrow$} & $\mathrm{MMD}_k^2(\mathcal{F}_r,\mathcal{F}_g)$ & Kernel distance between real and generated features. \\
    FVD & \textcolor{crEstimated}{$\downarrow$} & $\mathrm{FID}(\phi_v(\mathcal{V}_r),\phi_v(\mathcal{V}_g))$ & Frechet distance over video features. \\
    FGD & \textcolor{crEstimated}{$\downarrow$} & $\mathrm{FID}(\phi_m(\mathcal{M}_r),\phi_m(\mathcal{M}_g))$ & Frechet distance over motion features. \\
    Diversity & $\rightarrow D_{\mathrm{real}}$ & $\frac{2}{N(N-1)}\sum_{i<j}\|\phi(\hat{\mathbf{z}}_i)-\phi(\hat{\mathbf{z}}_j)\|_2$ & Global spread compared with the diversity of real samples. \\
    MultiModality & \textcolor{crReal}{$\uparrow$} & $\frac{1}{C}\sum_{c=1}^{C}\frac{2}{K_c(K_c-1)}\sum_{i<j}\|\phi(\hat{\mathbf{z}}_{c,i})-\phi(\hat{\mathbf{z}}_{c,j})\|_2$ & Conditional one-to-many variation. \\
    Coverage & \textcolor{crReal}{$\uparrow$} & $\frac{\left|\left\{\operatorname*{arg\,min}_{\mathbf{z}\in\mathcal{Z}}d(\hat{\mathbf{z}},\mathbf{z})\,\middle|\,\hat{\mathbf{z}}\in\hat{\mathcal{Z}}\right\}\right|}{|\mathcal{Z}|}$ & Fraction of distinct reference samples selected as nearest neighbors of generations. \\
    MinMatchDist & \textcolor{crEstimated}{$\downarrow$} & $\frac{1}{|\mathcal{Z}|}\sum_{\mathbf{z}\in\mathcal{Z}}\min_{\hat{\mathbf{z}}\in\hat{\mathcal{Z}}}d(\mathbf{z},\hat{\mathbf{z}})$ & Minimum matching distance from reference to generated samples. \\
    1-NNA & $\rightarrow 0.5$ & $\frac{1}{|\mathcal{U}|}\sum_{\mathbf{z}\in\mathcal{U}}\mathbf{1}[\mathrm{dom}(\mathbf{z})=\mathrm{dom}(\mathrm{NN}_{-}(\mathbf{z}))],\quad\mathcal{U}=\mathcal{Z}\cup\hat{\mathcal{Z}}$ & Leave-one-out domain separability between real and generated sets. \\

    \midrule
    \rowcolor{gray!10} \multicolumn{4}{l}{\textit{\textbf{Temporal Dynamics and Audio-Visual Synchronization}}}\\
    \midrule
    ADE & \textcolor{crEstimated}{$\downarrow$} & $\frac{1}{T}\sum_{t=1}^{T}\|\hat{\mathbf{q}}_t-\mathbf{q}_t\|_2$ & Average trajectory error over time. \\
    FDE & \textcolor{crEstimated}{$\downarrow$} & $\|\hat{\mathbf{q}}_T-\mathbf{q}_T\|_2$ & Final trajectory endpoint error. \\
    Beat Alignment & \textcolor{crReal}{$\uparrow$} & $\frac{1}{|\mathcal{B}_a|}\sum_{t_a\in\mathcal{B}_a}\exp\left(-\frac{\min_{t_m\in\mathcal{B}_m}\|t_m-t_a\|_2^2}{2\sigma^2}\right)$ & Gaussian temporal alignment between audio and motion beats. \\
    Foot Skating & \textcolor{crEstimated}{$\downarrow$} & $\frac{1}{|\mathcal{C}|}\sum_{(t,j)\in\mathcal{C}}\|\hat{\mathbf{p}}_{t,j}-\hat{\mathbf{p}}_{t-1,j}\|_2$ & Foot velocity during contact frames. \\
    Sync-C & \textcolor{crReal}{$\uparrow$} & $\operatorname{median}_{\delta\in\mathcal{D}}d_{\delta}-\min_{\delta\in\mathcal{D}}d_{\delta}$ & Confidence margin of the best audio-visual temporal offset. \\
    Sync-D & \textcolor{crEstimated}{$\downarrow$} & $\min_{\delta\in\mathcal{D}}d_{\delta}$ & Minimum audio-visual feature distance across temporal offsets. \\
    MCD-DTW & \textcolor{crEstimated}{$\downarrow$} & $\frac{1}{|\pi|}\sum_{(t,s)\in\pi}\frac{10}{\ln 10}\sqrt{2\sum_{\ell=1}^{L}(\hat{c}_{t,\ell}-c_{s,\ell})^2}$ & Mel-cepstral distortion over a DTW alignment path. \\

    \midrule
    \rowcolor{gray!10} \multicolumn{4}{l}{\textit{\textbf{Contact, Collision, and Physical Plausibility}}}\\
    \midrule
    Contact Precision & \textcolor{crReal}{$\uparrow$} & $\frac{C_{\mathrm{tp}}}{C_{\mathrm{pred}}}$ & Reliability of predicted contact regions. \\
    Contact Recall & \textcolor{crReal}{$\uparrow$} & $\frac{C_{\mathrm{tp}}}{C_{\mathrm{gt}}}$ & Coverage of true contact regions. \\
    Contact F1 & \textcolor{crReal}{$\uparrow$} & $\frac{2P_{\mathrm{con}}R_{\mathrm{con}}}{P_{\mathrm{con}}+R_{\mathrm{con}}}$ & Balanced contact precision and recall. \\
    Contact Distance & \textcolor{crEstimated}{$\downarrow$} & $\frac{1}{|\hat{\mathcal{C}}|}\sum_{\mathbf{x}\in\hat{\mathcal{C}}}d(\mathbf{x},\mathcal{C})$ & Geometric contact localization error. \\
    Penetration Depth & \textcolor{crEstimated}{$\downarrow$} & $\frac{1}{N}\sum_{i=1}^{N}\max(0,-\mathrm{SDF}(\hat{\mathbf{x}}_i))$ & Depth of invalid object or scene interpenetration. \\
    Collision Rate & \textcolor{crEstimated}{$\downarrow$} & $\frac{1}{N}\sum_{i=1}^{N}\mathbf{1}\left[\min_{1\leq n\leq P_i}\mathrm{SDF}(\hat{\mathbf{x}}_{i,n})<0\right]$ & Fraction of generated samples containing at least one collision. \\
    Physical Energy & \textcolor{crEstimated}{$\downarrow$} & $\sum_{k}\lambda_k g_k(\hat{\mathbf{s}}_{1:T})$ & Penalty for task-specific physical violations. \\

    \midrule
    \rowcolor{gray!10} \multicolumn{4}{l}{\textit{\textbf{Embodied Task Performance and Human Judgment}}}\\
    \midrule
    Success Rate & \textcolor{crReal}{$\uparrow$} & $\frac{N_{\mathrm{succ}}}{N_{\mathrm{trial}}}$ & Fraction of completed tasks or goals. \\
    Return & \textcolor{crReal}{$\uparrow$} & $\frac{1}{N}\sum_{i=1}^{N}\sum_{t=1}^{T}\gamma^t r_{i,t}$ & Discounted cumulative task reward. \\
    Goal Error & \textcolor{crEstimated}{$\downarrow$} & $\|\hat{\mathbf{g}}-\mathbf{g}\|_2$ & Distance to target state or waypoint. \\
    Preference Rate & \textcolor{crReal}{$\uparrow$} & $\frac{N_{\mathrm{win}}}{N_{\mathrm{cmp}}}$ & Pairwise human preference win rate. \\
    MOS & \textcolor{crReal}{$\uparrow$} & $\frac{1}{N}\sum_{i=1}^{N}r_i$ & Mean human opinion score. \\
    Identity Similarity & \textcolor{crReal}{$\uparrow$} & $\frac{\phi_{\mathrm{id}}(\hat{\mathbf{x}})^\top\phi_{\mathrm{id}}(\mathbf{x})}{\|\phi_{\mathrm{id}}(\hat{\mathbf{x}})\|_2\|\phi_{\mathrm{id}}(\mathbf{x})\|_2}$ & Identity preservation in generated humans. \\
    Garment Consistency & \textcolor{crReal}{$\uparrow$} & $\frac{\phi_g(\hat{\mathbf{x}})^\top\phi_g(\mathbf{x})}{\|\phi_g(\hat{\mathbf{x}})\|_2\|\phi_g(\mathbf{x})\|_2}$ & Clothing appearance and structure preservation. \\

    \midrule
    \rowcolor{gray!10} \multicolumn{4}{l}{\textit{\textbf{Efficiency and Scalability}}}\\
    \midrule
    FPS & \textcolor{crReal}{$\uparrow$} & $\frac{N_{\mathrm{frame}}}{t}$ & Rendered or processed frames per second. \\
    Throughput & \textcolor{crReal}{$\uparrow$} & $\frac{N_{\mathrm{unit}}}{t}$ & Processed samples or control steps per time. \\
    Runtime & \textcolor{crEstimated}{$\downarrow$} & $t_{\mathrm{total}}$ & Total wall-clock execution time. \\
    Latency & \textcolor{crEstimated}{$\downarrow$} & $\frac{t_{\mathrm{total}}}{N_{\mathrm{unit}}}$ & Wall-clock time per processed unit. \\
    Parameters & \textcolor{crEstimated}{$\downarrow$} & $|\theta|$ & Number of trainable or total model parameters. \\
    FLOPs & \textcolor{crEstimated}{$\downarrow$} & $\sum_{\ell}\mathrm{Ops}_{\ell}$ & Floating-point operation count. \\
    Memory & \textcolor{crEstimated}{$\downarrow$} & $\mathrm{Mem}_{\mathrm{peak}}$ & Peak memory footprint during evaluation. \\
\end{longtable}
}
\endgroup

\section{Open Challenges and Future Directions} 
\label{sec::future}
The preceding sections show that human-centric intelligence is progressing from specialized models toward more general systems that connect human perception, generation, interaction, and action. This transition raises several challenges that cannot be resolved by scaling individual tasks alone. We discuss six directions that may shape the next stage of the field.

\subsection{Scalable and Trustworthy Human Data}
\label{sec::scalable_and_trustworthy_human_data}
The limited availability of real human data remains a major obstacle to scaling human-centric models. Collecting such data often requires specialized capture systems, extensive annotation, and careful treatment of personal information. Synthetic data therefore offers an important path toward scalable pretraining. It can be generated with automatic annotations, expanded at relatively low marginal cost, and targeted toward cases that are difficult to capture in the real world. Recent studies show that synthetic human data can improve recognition and generation when it is appropriately combined with real data~\cite{kim2025vigface,lin2025ai,fei2026exploring}. However, the same mechanism that makes synthetic data scalable can also scale artifacts and biases inherited from the generator. Future research should therefore establish data scaling laws that distinguish the effect of data quantity from that of data quality. Controlled studies of real-synthetic mixtures, together with evaluation on real and underrepresented populations, could clarify when synthetic data provides genuine transfer. Provenance and consent should also remain traceable as data move through pretraining and adaptation~\cite{xiang2025fair,baltaretu2026identifying}.

\subsection{Unified Human-Centric Foundation Models}
Current unified models usually cover a restricted group of tasks or operate within a particular modality, such as human perception, motion-language modeling, or digital human generation~\cite{khirodkarsapiens2,wang2026thfm,li2025genmo,wang2026motiongpt,bao2026archon}. The next step is a general human-centric foundation model that supports unified representation learning across the full human context spectrum. Rather than learning separate representations for isolated tasks, such a foundation model should establish a shared representation space in which knowledge acquired from observable humans can support the modeling of human behavior, interaction, and embodied action. This space does not need to encode every signal in the same form. Modality-specific encoders and decoders can preserve the precision required by different outputs, while shared representations capture transferable human knowledge. Joint pretraining and cross-context alignment may provide practical routes toward this goal. The central evidence would be whether one pretrained model transfers across contexts with limited adaptation, while avoiding negative transfer and catastrophic forgetting.

\subsection{Physical Grounding for Next-Generation Human Representation Learning}
Humans act through physical bodies, yet current human representation learning still relies mainly on visual and linguistic regularities. As a result, learned representations may support the recognition or generation of plausible human behavior without capturing the physical laws that make it possible~\cite{gozlan2025opencapbench,huang2026phymotion,cavada2026newtphys}. Physical information should therefore become a central component of next-generation human representation learning rather than a constraint applied only after prediction or generation. In human understanding, physical constraints can reduce ambiguity when recovering motion from incomplete observations. In human generation, they can guide models away from unstable movement and invalid contact. They also provide the connection between observed motion and its effects during interaction or control. Progress may come from combining biomechanical measurements with physics-informed learning and differentiable simulation~\cite{werling2024addbiomechanics,xing2026interphys,le2026gravity,han2026piavatar}. Evaluation should correspondingly examine whether predicted or generated behavior remains physically valid, rather than relying only on perceptual and kinematic similarity.

\subsection{Human-Centric World Models and Embodied Action Intelligence}
A human-centric world model should jointly predict how future human or agent actions unfold and how the environment evolves in response. Current video-based world models can generate visually plausible futures, but perceptual fidelity alone does not establish that they have learned the causal relationship between actions and their consequences~\cite{xie2026generated,wang2026hand2world,shen2026egoforge}. Human videos provide scalable experience for learning such relationships, although the embodiment gap prevents observed behavior from being transferred directly into robot control~\cite{gao2026dreamdojo,ma2026humanscale,luo2025being,luo2026being}. Beyond prediction, embodied agents require action intelligence, namely the ability to select and sequence appropriate actions within a closed interaction loop. HumanCLAW separates this decision-making capability from low-level motor execution and reveals that current vision-language models still struggle to track their own bodies and the physical consequences of their actions~\cite{li2026humanclaw}. ComBodied Agents extends this closed-loop perspective from physical task execution to sustained human support. It places the person's evolving condition and agency at the center of modeling and uses Personal World Models to compare future human trajectories under alternative decisions and interventions~\cite{ding2026combodied}. Together, these directions motivate human-centric agents that connect world prediction with action selection while accounting for both physical outcomes and human responses. Evaluation should distinguish decision quality from motor execution and, for long-term support, examine whether agent interventions preserve human agency rather than measuring task completion alone.

\subsection{Evaluation for Generalization and Integrated Capabilities}
The main limitation of current evaluation is not the absence of metrics, but the lack of protocols that connect them. As reviewed in Sec.~\ref{sec::metrics}, existing benchmarks provide detailed measurements for individual forms of human perception, generation, and interaction~\cite{wang2025facebench,wang2026mhpr,cai2025humanvideo,zhou2026humanvbench}. However, strong performance on separate benchmarks does not establish that a model can transfer knowledge across human contexts or combine several capabilities within the same process. Future evaluation should be better organized as a benchmark suite. Task-specific metrics should remain responsible for diagnosing local accuracy, while a shared protocol evaluates generalization. When a system produces related outputs, their mutual consistency should be evaluated. For world models and embodied agents, prediction quality should additionally be connected to planning and closed-loop performance. The final result would be a structured evaluation profile rather than a single aggregate score~\cite{gozlan2025opencapbench,fang2026humanscore,yang2026avbench}.

\subsection{Efficient, Modular, and Deployable Human-Centric Systems}
The practical value of human-centric intelligence depends on whether increasingly capable models can operate efficiently and reliably in real systems. Continued scaling introduces substantial training cost, inference latency, privacy exposure, and maintenance complexity, making a single monolithic model unsuitable for many applications. Agentic and tool-augmented systems offer an alternative by coordinating foundation models, specialized human models, and external tools for perception~\cite{lin2025chathuman}, garment modeling~\cite{bian2025chatgarment}, motion generation~\cite{li2025chatmotion}, video understanding~\cite{yuan2025video}, and other human-centered tasks~\cite{zhu2026fusionagent}. Future systems may therefore become increasingly modular, invoking specialized capabilities when needed while retaining a shared semantic interface. This direction also raises questions concerning routing reliability, latency, privacy, error propagation, and whether independently developed components preserve a coherent account of the human subject.

\section{Conclusion}
\label{sec::conclusion}
This survey presents a full-spectrum account of human-centric intelligence in the foundation-model era. We introduce a human context taxonomy that connects six levels, namely visual appearance, spatial geometry, kinematic dynamics, interaction modeling, world simulation, and embodied agency, through three perspectives on humans as observable subjects, dynamic actors, and situated agents. Guided by this taxonomy, we establish the methodological foundations of the field, review advances across the six levels, and organize the datasets, benchmarks, and metrics used to develop and evaluate them. Our analysis suggests that the foundation-model era is characterized not simply by larger models, but by a transition toward scalable human data, reusable priors, multimodal interfaces, transferable capabilities, and general connections between perception, generation, reasoning, and action. Nevertheless, these connections remain incomplete, particularly when models are required to preserve human consistency across contexts, obey physical constraints, or transfer knowledge into executable behavior. By consolidating these developments and identifying their shared challenges, we hope this survey and its continuously updated project resources provide a coherent reference for advancing scalable, trustworthy, physically grounded, and deployable human-centric intelligence.

\clearpage
\bibliographystyle{plainnat}
\bibliography{reference}

\end{document}